\documentclass[twoside,letterpaper]{article}

\usepackage[preprint]{aistats2027}
\usepackage[round,authoryear]{natbib}

\usepackage[utf8]{inputenc}
\usepackage[T1]{fontenc}
\usepackage{microtype}
\usepackage{amsmath,amssymb,amsthm,mathtools}
\usepackage{bm}
\usepackage{booktabs}
\usepackage{multirow}
\usepackage{graphicx}
\usepackage{xcolor}
\usepackage{listings}
\usepackage{float}
\usepackage{placeins}
\usepackage{needspace}
\usepackage[colorlinks=true,linkcolor=blue!60!black,citecolor=blue!60!black,urlcolor=blue!60!black]{hyperref}
\usepackage{url}
\hypersetup{
  pdftitle={Neural Dimensional Analysis: Foundations of Dimensionless and Scale-Consistent Computation},
  pdfauthor={Bum Jun Kim, Makoto Kawano, Yusuke Iwasawa, Yutaka Matsuo}
}
\usepackage{titletoc}

\definecolor{ndacodebg}{HTML}{F7F9FC}
\definecolor{ndacodeframe}{HTML}{CBD5E1}
\definecolor{ndacodetext}{HTML}{1F2937}
\definecolor{ndacodekeyword}{HTML}{1D4ED8}
\definecolor{ndacodebuiltin}{HTML}{7E22CE}
\definecolor{ndacodefunction}{HTML}{047857}
\definecolor{ndacodestring}{HTML}{B45309}
\definecolor{ndacodecomment}{HTML}{64748B}
\definecolor{ndacodenumber}{HTML}{9F1239}
\definecolor{ndacodelineno}{HTML}{64748B}

\lstdefinestyle{ndapython}{
  language=Python,
  basicstyle=\ttfamily\scriptsize\color{ndacodetext},
  backgroundcolor=\color{ndacodebg},
  keywordstyle=\color{ndacodekeyword}\bfseries,
  morekeywords=[2]{self,super,True,False,None,torch,nn,F},
  keywordstyle=[2]\color{ndacodebuiltin},
  commentstyle=\color{ndacodecomment}\itshape,
  stringstyle=\color{ndacodestring},
  emph={GCTBlock,canon,apply_rope,_attention,_mlp,_update,forward,__init__},
  emphstyle=\color{ndacodefunction}\bfseries,
  numbers=left,
  numberstyle=\tiny\color{ndacodelineno},
  numbersep=8pt,
  stepnumber=1,
  frame=single,
  rulecolor=\color{ndacodeframe},
  framerule=0.5pt,
  framesep=5pt,
  xleftmargin=2em,
  framexleftmargin=1.4em,
  framexrightmargin=0.4em,
  columns=fullflexible,
  keepspaces=true,
  showstringspaces=false,
  tabsize=4,
  breaklines=true,
  breakatwhitespace=false,
  captionpos=b,
  abovecaptionskip=6pt,
  belowcaptionskip=0pt,
  literate=*
    {0}{{{\color{ndacodenumber}0}}}{1}
    {1}{{{\color{ndacodenumber}1}}}{1}
    {2}{{{\color{ndacodenumber}2}}}{1}
    {3}{{{\color{ndacodenumber}3}}}{1}
    {4}{{{\color{ndacodenumber}4}}}{1}
    {5}{{{\color{ndacodenumber}5}}}{1}
    {6}{{{\color{ndacodenumber}6}}}{1}
    {7}{{{\color{ndacodenumber}7}}}{1}
    {8}{{{\color{ndacodenumber}8}}}{1}
    {9}{{{\color{ndacodenumber}9}}}{1}
}

\newcommand{\secref}[1]{Sec.~\ref{#1}}

\makeatletter
\def\p@equation#1{Eq.~#1}
\makeatother
\renewcommand{\eqref}[1]{\ref{#1}}

\newtheorem{theorem}{Theorem}
\newtheorem{proposition}{Proposition}
\newtheorem{lemma}{Lemma}
\newtheorem{corollary}{Corollary}

\newcommand{\uof}[1]{u(#1)}
\newcommand{\RMS}{\operatorname{RMS}}
\newcommand{\Var}{\operatorname{Var}}
\newcommand{\canon}{\mathcal{C}}
\newcommand{\Lcal}{\mathcal{L}}
\newcommand{\UVI}{\mathrm{UVI}}
\newcommand{\gct}{GCT}
\newcommand{\dl}{\mathrm{d}}
\newcommand{\E}{\mathbb{E}}

\newcommand{\quantgcthfour}{0.828}
\newcommand{\quantgcthfourtok}{0.107}

\newcommand{\quantngpthfourtok}{0.52}
\newcommand{\quantworsthfour}{7.71}

\newcommand{\quantworsthfourtok}{1.044}

\newcommand{\sqwsmoothworst}{+0.061}
\newcommand{\sqwsmoothworstname}{QK-Norm}
\newcommand{\sqhchanworst}{+0.003}
\newcommand{\sqhchanworstname}{LLaMA-style}
\newcommand{\sqgctwsmooth}{+0.008}
\newcommand{\sqgcthchan}{+0.000}
\newcommand{\sqwstatworst}{+0.270}
\newcommand{\sqhchanworstoneb}{+0.007}
\newcommand{\sqgcthchanoneb}{+0.000}

\newcommand{\wdfairnessclaim}{setting weight decay to zero changes the other 51M baselines by at most 0.004 nats, with $-0.004$ for Pre-LN and $+0.001$ for LLaMA-style, while \gct{} changes by $+0.003$, leaving the clean-loss ordering unchanged}

\newcommand{\perturbworst}{up to 19.7 nats}
\newcommand{\perturbfullworst}{43.1 nats}
\newcommand{\perturbuctembdown}{+1.406}
\newcommand{\perturbuctmiddown}{+0.211}
\newcommand{\perturbucteqdown}{0.164}
\newcommand{\kvworst}{0.0004}
\newcommand{\kvworsttok}{0.0001}
\newcommand{\pinnunitworst}{0.015}
\newcommand{\pinnbaseworst}{2.973}
\newcommand{\pinncomboworst}{0.005}

\newcommand{\vitclaim}{has the highest clean top-1 accuracy among the five variants at $74.2\%$, compared with $70.7\%$ for QK-Norm}

\newcommand{\noprojdepthclaim}{removing $P^\perp$ increases loss by $0.010$ nats at $L=24$ and $0.035$ nats at $L=48$}
\newcommand{\depthgctln}{3.903}
\newcommand{\depthngptln}{4.243}

\newcommand{\quantworsthstat}{1.10}

\newcommand{\quantgcthstatoneb}{0.009}
\newcommand{\quantworsthstatoneb}{5.51}
\newcommand{\quantworsthstatonebname}{LLaMA-style}

\newcommand{\quantscaletrend}{The largest baseline loss increase under static per-tensor stream int8 rises from $+1.10$ nats at 124M to $+2.45$ at 350M and $+5.51$ at 1B, while \gct{} increases by $+0.001$, $+0.003$, and $+0.009$ nats, respectively}
\newcommand{\otskappamin}{48}
\newcommand{\otskappamax}{596}

\newcommand{\otsowngctdyn}{0.001}

\newcommand{\otspplladder}{27.7, 20.3, 17.5, and 16.0}

\newcommand{\intervtau}{2.79}

\newcommand{\ieightnarch}{8}
\newcommand{\ieightgapmax}{0.002}
\newcommand{\ieightdeltamax}{0.233}
\newcommand{\widthplateauclaim}{Raising the rate from $6\times10^{-4}$ to $4.8\times10^{-3}$ improves \gct{} at every width, with changes of $-0.592$, $-0.369$, and $-0.245$ nats for $d\in\{256,512,1024\}$. At the higher rate, Pre-LN is $+0.141$, $+0.131$, and $+2.039$ nats above Pre-LN's best sampled value at the respective width}
\newcommand{\mupwidthclaim}{The rule changes \gct{} loss by at most $0.006$ nats at either width, namely \mupgctdelta{}, while the Pre-LN changes are \mupprelndelta{}}
\newcommand{\mupgctdelta}{$-0.005$ at $d=256$ and $+0.006$ at $d=1024$}
\newcommand{\mupprelndelta}{$+0.020$ at $d=256$ and $-0.068$ at $d=1024$}

\newcommand{\ngptwddelta}{+0.012}
\newcommand{\lrtransferclaim}{brings \gct{} to $3.045$, within $0.002$ nats of nGPT at $3.043$}

\newcommand{\muongctdelta}{-0.309}
\newcommand{\muonprelndelta}{-0.329}
\newcommand{\muonqknormdelta}{-0.242}
\newcommand{\muonclaim}{All three architectures improve at matched tokens, with similar gains for the typed and untyped blocks}

\newcommand{\kvworstoneb}{0.0002}

\newcommand{\kvonebsentence}{The control holds unchanged at 1B, where the worst increase is $+$\kvworstoneb{} nats.}

\newcommand{\pgmatch}{54 of 54}

\newcommand{\pgtypedmax}{$2.4\times10^{-7}$}
\newcommand{\pgworst}{$2.6\times10^{-2}$}

\newcommand{\pgminnonnull}{$3.5\times10^{-3}$}
\newcommand{\pguctfd}{$1.5\times10^{-7}$}
\newcommand{\pguctpair}{$1.1\times10^{-3}$}
\newcommand{\pgmatchoneb}{34 of 34}
\newcommand{\pgtypedmaxoneb}{$3.3\times10^{-9}$}

\newcommand{\ptcostratio}{1.02$\times$}
\newcommand{\ptcostmetakb}{48}
\newcommand{\ptcostfrac}{72\%}
\newcommand{\servingbfgib}{5.25}
\newcommand{\servingeightgib}{2.63}
\newcommand{\servingfourgib}{1.32}

\newcommand{\servinggctcomp}{4}

\newcommand{\servinggctdelta}{+0.739}

\newcommand{\servingbenchbatch}{32}
\newcommand{\servingbenchctx}{2048}

\newcommand{\quantdstasks}{LAMBADA, HellaSwag, PIQA, and ARC-Easy}

\newcommand{\vittinclaim}{has the highest clean top-1 accuracy, $59.2\%$, compared with $56.4\%$ for LayerScale}
\newcommand{\vitnoisegct}{34.4\%}
\newcommand{\vitnoisebest}{37.5\%}

\newcommand{\ngptbespokeclaim}{Removing both weight decay and warmup, as in nGPT's original recipe, while retaining our other settings increases nGPT's best loss at the two tested high learning rates by $0.093$ nats, from $3.500$ to $3.593$. These controls do not indicate a disadvantage from the shared recipe in this setting.}

\newcommand{\ngptpperptransplant}{Adding $P^\perp$ changes nGPT's loss by approximately $+0.010$ nats at $L=24$ and reduces nGPT's loss by $0.275$ nats at $L=48$, from $4.243$ to $3.968$}

\newcommand{\sqtwoworsthstat}{5.67}
\newcommand{\sqtwoworstname}{LLaMA-style}
\newcommand{\sqtwogcthstat}{0.008}
\newcommand{\sqtwogcthfourtok}{0.47}
\newcommand{\sqtwobasehfourtokmin}{5.13}
\newcommand{\pgmatchonebstwo}{34 of 34}

\begin{document}
\runningtitle{Neural Dimensional Analysis}
\runningauthor{Bum Jun Kim, Makoto Kawano, Yusuke Iwasawa, Yutaka Matsuo}

\renewcommand{\thefootnote}{\fnsymbol{footnote}}
\twocolumn[
\aistatstitle{Neural Dimensional Analysis:\\ Foundations of Dimensionless and Scale-Consistent Computation}
\aistatsauthor{Bum Jun Kim\footnotemark[1] \And Makoto Kawano \And Yusuke Iwasawa \And Yutaka Matsuo}
\aistatsaddress{The University of Tokyo}
]
\footnotetext[1]{\hangindent=1.8em\hangafter=1 Corresponding author:\newline\href{mailto:bumjun.kim@weblab.t.u-tokyo.ac.jp}{\texttt{bumjun.kim@weblab.t.u-tokyo.ac.jp}}}
\renewcommand{\thefootnote}{\arabic{footnote}}

\begin{abstract}
Modern Transformers rely on many scale-control mechanisms, including normalization, residual scaling, attention temperatures, loss weighting, and adaptive updates. To explain how these mechanisms maintain scale consistency, we introduce Neural Dimensional Analysis (NDA), a unified framework based on dimensional consistency. We define neural units that specify how tensors rescale and which quantities can be added, just as physical units do in physics. Our rules match residual units, nondimensionalize attention and gate inputs, combine losses across units, and convert gradient units into parameter units. We implement NDA through dimensionless Unit Violation Indices (UVI) for diagnosing scale mismatches within Transformers, a unit-consistent multi-loss objective, and the Gauge-Canonical Transformer (GCT) with normalized coordinates and orthogonalized residual additions. In language-model experiments spanning 51M to 1B parameters, GCT remains stable during training across a broad sixteenfold learning-rate range, outperforming Pre-LN, LLaMA-style, and QK-Norm in 124M validation loss at tuned learning rates. GCT's normalized residual stream also shows improved robustness to quantization. At 1B parameters, calibrated per-tensor int8 quantization increases GCT's validation loss by only 0.009 nats, compared with up to 5.51 nats for the baselines. Vision and physics-informed neural network experiments further confirm NDA's benefits beyond language modeling. These results establish dimensional consistency as a neural network design principle.
\end{abstract}

\section{Introduction}
\label{sec:intro}

\begin{figure*}[t!]
\centering
\includegraphics[width=\textwidth]{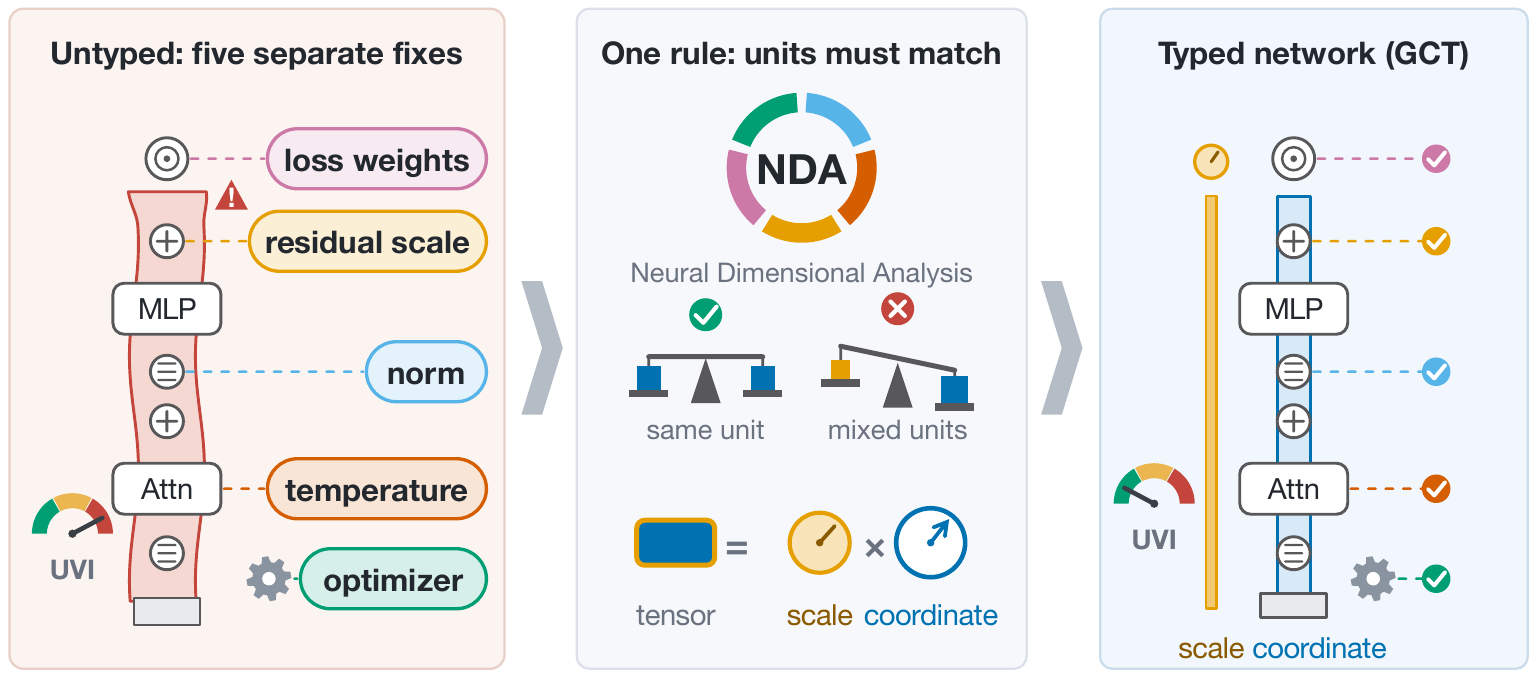}
\caption{NDA at a glance. (Left) Scale-control mechanisms in deep learning, including normalization, residual scaling, attention temperature, loss weighting, and adaptive updates, manage scale at different interfaces of a Transformer. Unit Violation Indices (UVI) flag deviations from the intended scale regime, locating the interface where a scale mismatch arises. (Middle) NDA expresses these mechanisms through unit-consistent operations and a factorization into scales and canonical coordinates. (Right) The Gauge-Canonical Transformer (\gct{}) builds the whole network on this factorization. Every block computes on dimensionless coordinates (blue), ensuring matching units at every interface, including attention and multilayer perceptron (MLP) branches. The scale (gold) is reattached only when an output needs units.}
\label{fig:overview}
\end{figure*}

Deep learning employs a broad family of mechanisms for controlling internal scale. Batch normalization, layer normalization (LayerNorm), and root-mean-square normalization (RMSNorm) normalize activations \citep{ioffe2015batchnorm,ba2016layer,zhang2019rmsnorm}. DeepNet, ReZero, LayerScale, and Fixup control the scale of residual additions \citep{wang2022deepnet,bachlechner2020rezero,touvron2021cait,zhang2019fixup}. Attention scaling and query--key normalization (QK-Norm) regulate softmax inputs \citep{vaswani2017attention,henry2020qknorm,dehghani2023vit22b,riviere2024gemma}. Loss-balancing methods and adaptive optimizers manage scale imbalances in objectives and parameter updates \citep{kendall2018uncertainty,chen2018gradnorm,wang2021understanding,kingma2015adam,you2020lamb}. These scale-control mechanisms play a critical role in improving training stability and performance \citep{xiong2020layer,zhai2023stabilizing,wortsman2024proxies}. All of these mechanisms manage how quantities of different scales interact within a neural network (Figure~\ref{fig:overview}).

We interpret these scale-control mechanisms through dimensional consistency. In physics, terms added together must have compatible units, and the arguments of $\exp$ and $\log$ must be dimensionless. A physical unit, such as a centimeter or a meter, provides a reference scale for expressing a measured quantity. Inspired by physical units and dimensions, we define a neural unit as the rule governing how an internal tensor scales when the computation graph is rescaled. Neural units describe internal scaling behavior even when the data have no physical units. In terms of neural units, each scale-control mechanism in deep learning becomes a unit operation. Normalization removes units, residual branches restore stream units, attention temperatures convert logits to dimensionless quantities, loss weights convert loss units, and optimizers map gradient units back to parameter units. Building on this view, we propose Neural Dimensional Analysis (NDA), a framework that unifies these unit operations under one set of local typing rules.

\paragraph{Positioning.} Prior work on scale control has addressed individual aspects of scale for specific purposes. For example, units-equivariant learning handles the physical units of input and output data \citep{villar2023dimensionless}, maximal-update parameterization ($\mu$P) prescribes parameterization and update scales for hyperparameter transfer \citep{yang2021tensor,yang2021tp5}, unit scaling chooses scales for low-precision training \citep{blake2023unit}, and the normalized Transformer (nGPT) constrains activations and weights to the unit hypersphere \citep{loshchilov2024ngpt}. In contrast, our NDA framework offers a unified view of scale control over the whole computation graph, covering activations, objectives, gradients, and optimizers, and additionally provides diagnostics and local corrections. App.~\ref{app:positioning} develops these connections.

\paragraph{Contributions.} We make three contributions:
\begin{enumerate}
\clubpenalty=10000
\item A graph-level unit calculus covering activations, objectives, gradients, and updates, together with a theoretical analysis. This calculus links forward computation and backpropagation through a single set of unit rules (Prop.~\ref{prop:dualsound}). To guide unit-consistent network design and training, we prove theorems on loss weighting, optimizer updates, canonical factorization, and quantization error (App.~\ref{app:supptheory}).
\item Diagnostics and designs derived from the calculus. To make the unit rules testable in trained networks, we introduce dimensionless UVI and a weight-space audit with per-block rescaling predictions (Prop.~\ref{prop:pgauge}). \gct{} computes every block on dimensionless coordinates and uses an orthogonal projection to control scale growth during residual updates. The calculus also provides local scale corrections for existing architectures, along with unit-consistent training objectives and parameter updates.
\item Language-model experiments from 51M to 1B parameters under matched recipes. \gct{} training remains stable across a $16\times$ learning-rate range, and the \gct{} hyperparameters transfer unchanged to other \gct{} models up to $6\times$ as deep, with clean validation loss comparable to strong baselines (Figure~\ref{fig:lrsweep}; Tables~\ref{tab:depth} and~\ref{tab:clean124m}). The orthogonal projection also improves nGPT in the depth comparison, reducing nGPT's loss by $0.275$ nats at $L=48$. \gct{}'s normalized residual stream also improves robustness to low-precision storage by keeping token scales uniform. At 1B, calibrated per-tensor 8-bit integer (int8) quantization of this stream increases \gct{}'s validation loss by only \quantgcthstatoneb{} nats, compared with \quantworsthstatoneb{} nats for the worst baseline (Table~\ref{tab:quant1b}). Vision and physics-informed neural network (PINN) studies validate the same principles beyond language modeling (App.~\ref{app:full}).
\end{enumerate}

\section{Neural Unit Consistency}
\label{sec:units}

This section formulates local rules for managing scale on a neural network's computation graph. The rules specify which quantities can be added and which inputs must be dimensionless. The rules also determine how loss weights must compensate for changes in loss units and how gradients must rescale when activations or parameters are rescaled. These rules provide the basis for the diagnostics in \secref{sec:uvi} and the architecture in \secref{sec:gct}.

\subsection{Units as transformation laws}

Let $\mathcal{G}$ be a computation graph whose node $v$ carries a tensor $x_v$ in a finite-dimensional real inner-product space $V_v$. We declare a rescaling group $G \coloneqq (\mathbb{R}_+)^{K}$, with one factor for each independent internal scale degree of freedom. A neural unit of node $v$ is a representation
\begin{equation}
\rho_v : G \to GL(V_v), \qquad x_v \mapsto \rho_v(g)x_v .
\end{equation}
A tensor is dimensionless if and only if $\rho_v \equiv I$. The scalar units considered here are the continuous characters $\chi_u(g) \coloneqq \prod_{k=1}^{K} g_k^{u_k}$, $u \in \mathbb{R}^{K}$, with $\rho_v=\chi_{u_v}I$. We then write $\uof{x}\coloneqq\chi_{u_x}$ and use the multiplicative notation $\uof{xy}=\uof{x}\uof{y}$.

The unit type $\uof{x}$ specifies how $x$ transforms under $G$, whereas the empirical scale $s_x$ measures magnitude and shares that type, making $x/s_x$ dimensionless. The root-mean-square (RMS) value is one such scale on a shared-unit block; mixed-unit tensors require separate scales or a unit-compatible metric (App.~\ref{app:typingexamples}).

For an operation $y\coloneqq f(x_1,\dots,x_n)$, the declared transformation laws are $\rho_1,\dots,\rho_n$ for the input tensors and $\rho_y$ for the output tensor. We call the operation unit-consistent when the following equivariance identity holds for every $g\in G$:
\begin{equation}
\label{eq:unitconsistent}
f(\rho_1(g)x_1,\dots,\rho_n(g)x_n) = \rho_y(g) f(x_1,\dots,x_n).
\end{equation}

In words, rescaling the inputs before the operation must give the same result as rescaling the output afterward.

Parameters follow the same rule. In $y\coloneqq f_\phi(x)$, we treat $\phi$ as another input and declare the parameter transformation $\phi\mapsto\rho_\phi(g)\phi$. App.~\ref{app:typingexamples} details linear maps and explicit scale converters.

We also need gradient diagnostics that can be compared across blocks with different units. The next proposition shows that the forward unit declarations fix how every backpropagated gradient transforms.

\begin{proposition}[Reverse-mode dual typing]
\label{prop:dualsound}
Let $\mathcal{G}$ be a directed acyclic graph (DAG) whose nodes carry declared units $\rho_v$ and whose every primitive operation is unit-consistent with respect to those declarations. If the loss node is typed dimensionless, then, at every point where the relevant derivatives exist, every reverse-mode gradient node transforms by the dual law $\rho_v^{-\top}$. The proof is given in App.~\ref{app:dualsound}.
\end{proposition}

Under the declared scalar rescaling, $x\mapsto cx$ gives $\nabla_x\Lcal\mapsto c^{-1}\nabla_x\Lcal$, meaning that multiplying the activation by $c$ divides the activation gradient by $c$. These factors cancel in the primal--dual RMS product in \ref{eq:G}, making the product comparable across rescalings of a shared scalar-unit block. To keep a scalar diagnostic unchanged under a change of units, construct the diagnostic's dimensionless output from unit-consistent operations.

\subsection{Local typing rules}
\label{sec:rules}

\paragraph{Addition and residuals.} $z \coloneqq x + y$ requires $\uof{x} = \uof{y} = \uof{z}$. In $h_{\ell+1} = h_\ell + r_\ell$, the branch must return the residual-stream unit. In pre-layer-normalization (Pre-LN) blocks, $r_\ell \coloneqq F_\ell(N(h_\ell))$ with $N : U_{\mathrm{res}} \to 1$. The branch $F_\ell : 1 \to U_{\mathrm{res}}$ restores the stream unit, a role played implicitly by output projections and gains in vanilla blocks.

\paragraph{Normalization as nondimensionalization.} With $s_x\coloneqq\RMS(x)$ on a nonzero shared-unit block, $N(x)\coloneqq x/s_x$ satisfies $N(cx)=N(x)$ for $c>0$. This division removes the unit and the overall magnitude. For outputs that need units, \secref{sec:gct} retains explicit scales alongside the dimensionless coordinates. App.~\ref{app:typingexamples} gives the gauge interpretation and affine-normalization examples.

\paragraph{Softmax and attention.} Softmax depends on logit differences and is unchanged by a common shift within each row. A logit difference must be dimensionless because exponentiating the logit difference gives a probability ratio, which must not change merely because the unit system changes. If raw scores $z_{ij}\coloneqq q_i^{\top}k_j$ carry unit $\uof{q}\uof{k}$, the temperature is a unit converter with $\uof{\tau}=\uof{q}\uof{k}$. With per-head coordinate scales $s_Q$ and $s_K$, the calibrated temperature is
\begin{equation}
\label{eq:tau}
\tau_{\ell,h} \coloneqq \sqrt{d_h} s_{Q,\ell,h} s_{K,\ell,h},
\end{equation}
of which the standard $\sqrt{d_h}$ is the $s_Q = s_K = 1$ special case. Value aggregation preserves $\uof{V}$. The output projection must map $U_V$ to $U_{\mathrm{res}}$. App.~\ref{app:attentionderiv} gives the variance assumptions and empirical scale estimates.

\paragraph{Activations and gates.} Standard non-homogeneous activations, including sigmoid, tanh, the Gaussian error linear unit (GELU), and the sigmoid linear unit (SiLU), require dimensionless arguments. In gated MLPs $z\coloneqq v\odot g$, units multiply. To retain the value-branch unit $\uof{v}$ at the output, the gate must be dimensionless, $\uof{g}=1$. The raw gate $a\sigma(a)$ of a Swish-gated linear unit (SwiGLU) silently assumes that $a$ is dimensionless, whereas the normalized gate $\tilde g\coloneqq(a/s_a)\sigma(a/s_a)$ is typed. App.~\ref{app:typingexamples} gives the homogeneity and transition-scale examples.

\paragraph{Loss aggregation.} To keep the objective value unchanged under a change of unit system, we construct a dimensionless objective using unit-consistent operations. A dimensionless objective has transformation factor one. Cross-entropy is dimensionless when the input logits are dimensionless, whereas squared error has $\uof{\|y-\hat y\|^2}=U_y^2$. In $\Lcal\coloneqq\sum_i\lambda_i\Lcal_i$, consistency forces $\uof{\lambda_i}=\uof{\Lcal_i}^{-1}$. Each loss weight acts as an inverse-unit converter. We factor each loss weight into a dimensionless preference $\alpha_i$ and an inverse characteristic scale $1/S_i$:
\begin{equation}
\label{eq:lossfact}
\lambda_i = \alpha_i / S_i, \qquad \uof{S_i} = \uof{\Lcal_i}, \quad \uof{\alpha_i} = 1,
\end{equation}
where $1/S_i$ converts each loss to a dimensionless value and $\alpha_i$ sets the corresponding task preference. App.~\ref{app:lossweights} gives characteristic-scale examples and characterizes invariant linear weighting rules.

\paragraph{Gradients and dual units.} If $\Lcal$ is dimensionless and $\uof{x} = U$, then $\dl\Lcal = \langle \nabla_x \Lcal, \dl x\rangle$ forces $\uof{\nabla_x \Lcal} = U^{-1}$. A raw gradient norm changes with the chosen unit. Comparing blocks with different units requires unit cancellation. Dimensionless sensitivity scales include $x \odot \nabla_x\Lcal$ elementwise, or on shared-unit blocks
\begin{equation}
\label{eq:G}
\Pi_x \coloneqq \RMS(x)\RMS(\nabla_x \Lcal),
\end{equation}
the primal--dual RMS product.

\paragraph{Optimizer updates and primal units.} The update $\theta_{t+1}=\theta_t+\Delta\theta_t$ needs $\uof{\Delta\theta}=U_\theta$, while $\uof{\nabla_\theta\Lcal}=U_\theta^{-1}$. With a dimensionless learning rate, a typed preconditioner supplies the missing factor through $\uof{P_\theta}=U_\theta^2$. App.~\ref{app:optimizergeometry} gives relative-update examples and the covariance and quotient-geometry results.

\subsection{Residual scale dynamics}
\label{sec:dynamics}

Residual additions can amplify scale even when both terms carry the same unit. Here $r_\ell$ is the branch contribution added to the hidden state during the forward pass. We track this amplification using the branch-to-stream RMS ratio and the branch's alignment with the stream.

Let $s_\ell\coloneqq\|h_\ell\|_2/\sqrt d$ and assume $h_\ell\ne0$. For $r_\ell\ne0$, define the dimensionless residual step size $\alpha_\ell\coloneqq\RMS(r_\ell)/\RMS(h_\ell)>0$. Expanding $\|h_\ell+r_\ell\|_2^2$ gives
\begin{equation}
\label{eq:drift}
s_{\ell+1}^2 = s_\ell^2(1 + \alpha_\ell^2 + 2\alpha_\ell c_\ell), \quad c_\ell \coloneqq \frac{\langle h_\ell, r_\ell\rangle}{\|h_\ell\|\|r_\ell\|}.
\end{equation}
For persistent positive alignment $c_\ell\ge c_0>0$, bounded scale requires $\sum_\ell\alpha_\ell=O(1)$, giving $O(L^{-1})$ per layer for uniformly comparable steps. Orthogonalization removes the linear term and permits $\sum_\ell\alpha_\ell^2=O(1)$ with uniform $\Theta(L^{-1/2})$ steps, as in depth-wise $\mu$P \citep{yang2024tp6}. \secref{sec:gct} implements this projection. App.~\ref{app:residualderiv} discusses zero branches, nonuniform schedules, and near-orthogonal contributions.

App.~\ref{app:supptheory} collects the full loss-weighting, optimizer, and residual-stability results, while App.~\ref{app:deriv} derives the attention temperature and residual recursion.

\section{UVI}
\label{sec:uvi}

To locate scale mismatches within a neural network during training, we propose UVI. The indices measure dimensionless log-deviations from a target or log-dispersion across depth. Large values point to the corresponding interface for inspection. Each index is defined relative to a declared unit assignment and reference regime. With a small dimensionless floor $\epsilon$, the residual and attention indices are
\begin{align}
\UVI_{\mathrm{res}}(\ell) &\coloneqq \Big|\log \tfrac{\alpha_\ell + \epsilon}{\alpha^{\star} + \epsilon}\Big|, \qquad \alpha^{\star} \coloneqq \tfrac{1}{\sqrt{2L}},
\label{eq:uvires}\\
\UVI_{\mathrm{attn}}(\ell,h) &\coloneqq \Big|\log \tfrac{\bar v_{\ell,h} + \epsilon}{v^{\star} + \epsilon}\Big|, \qquad v^{\star} \coloneqq 1,
\label{eq:uviattn}
\end{align}
where $\alpha_\ell$ is the measured residual step size, and $\bar v_{\ell,h}$ is the batch-averaged row-centered logit variance after all model-specific scaling.

App.~\ref{app:uvicalibration} defines the activation, gradient, and optimizer indices and the auxiliary drift diagnostics. All five indices and the two additional diagnostics can be computed inexpensively on small batches at intervals during training.

\section{GCT Architecture}
\label{sec:gct}

The goal is to make the block's dimensionless computation independent of arbitrary changes in internal scale. We apply the NDA rules in two designs, the Unit-Consistent Transformer (UCT) and \gct{}. UCT adds scale-correction modules to a conventional Transformer, while \gct{} normalizes the residual stream after every residual addition. App.~\ref{app:uctmodules} specifies UCT's three corrective modules.

\gct{} computes every block on normalized, dimensionless coordinates and keeps any required magnitude and units separate. The name Gauge-Canonical reflects our choice to fix the scale gauge through RMS normalization. In the exact construction, all positive rescalings of a nonzero vector share one RMS-one representative, which is the canonical coordinate. Each unitful quantity is factorized as
\begin{equation}
\text{unitful} = \text{unit scale} \times \text{canonical coordinate}.
\end{equation}

\subsection{Canonical state and residual block}

\paragraph{State factorization.} We separate a token state's magnitude and direction by writing $h_\ell = s_\ell\bar h_\ell$, with $s_\ell=\RMS(h_\ell)$ and $\uof{s_\ell} = U_{\mathrm{res}}$. For $h_\ell\ne0$, dividing by $s_\ell$ gives $\bar h_\ell=h_\ell/s_\ell$ with $\RMS(\bar h_\ell)=1$. We call $s_\ell$ the scale state, which carries the magnitude and units separately from the vector used for computation. The dimensionless vector $\bar h_\ell$ is the canonical coordinate, which preserves the direction of $h_\ell$ and has RMS one. For numerical stability, we use the canonicalizer
\begin{equation}
\canon(x) \coloneqq \frac{x}{\sqrt{\RMS(x)^2 + \epsilon_x}}.
\end{equation}
For $x\ne0$, setting $\epsilon_x=0$ gives exact RMS normalization. The finite-floor implementation and its deviations from exact unit consistency are described in App.~\ref{app:floordefect}.

\paragraph{Orthogonalized residual addition.} An attention or MLP branch produces an output vector $\bar a_\ell$, which we call the branch proposal to denote the branch's suggested contribution to the current state. \gct{} projects, normalizes, and scales the branch proposal before adding the resulting contribution to the residual stream. To remove the first-order scale growth identified in \secref{sec:dynamics}, we subtract the component of the branch output that points along the current state. The analysis uses the exact orthogonal projector
\begin{align*}
P^{\perp}_{\bar h}a \coloneqq a-\frac{\langle a,\bar h\rangle}{\|\bar h\|^2}\bar h, \qquad \bar h\ne0,
\end{align*}
and learnable dimensionless scalars $\alpha_\ell$, initialized at $1/\sqrt{2L}$. Each block applies the attention sub-block followed by the MLP sub-block, using the following update for each:
\begin{equation}
\label{eq:gctblock}
\begin{aligned}
\bar a_\ell &\coloneqq \Phi_\ell(\bar h_\ell),\qquad \hat a_\ell \coloneqq \canon(P^{\perp}_{\bar h_\ell}\bar a_\ell),\\
\bar h_{\ell^+} &\coloneqq \canon(\bar h_\ell + \alpha_\ell\hat a_\ell).
\end{aligned}
\end{equation}
Here $\Phi_\ell$ is the attention or MLP branch, and $\ell^+$ denotes the state after that sub-block. With a nonzero projected proposal and zero canonicalization floors, $\hat a_\ell\perp\bar h_\ell$ and both vectors have RMS one. Before the final normalization (retraction), the contribution $\alpha_\ell\hat a_\ell$ is orthogonal to the stream and has relative RMS $|\alpha_\ell|$; retraction returns the state to RMS one. Orthogonality applies before retraction, and the final displacement need not be orthogonal. App.~\ref{app:residualderiv} gives the exact rotation and limiting cases.

\subsection{Canonical interfaces and guarantees}

\paragraph{Attention in canonical coordinates.} Queries and keys are canonicalized per head vector, $\bar q\coloneqq\canon(q)$ and $\bar k\coloneqq\canon(k)$. A per-head dimensionless gain $g_h$, initialized at $1$, tunes the attention entropy within these canonical coordinates. App.~\ref{app:attentionderiv} gives the variance assumptions.

\paragraph{Gates in canonical coordinates.} The gated MLP canonicalizes preactivations per token to ensure that the sigmoid receives a dimensionless input. The MLP uses the normalized gate $\tilde g \coloneqq \canon(a)\sigma(\canon(a))$ to multiply the value branch.

\paragraph{Heads, scales, and parameters.} The classification head produces dimensionless logits from $\canon(\bar h_L)$ with a learnable dimensionless logit gain. Outputs with units use $\hat y\coloneqq s_y\hat{\bar y}$, where the explicit scale $s_y$ restores the magnitude and units. For tasks with dimensionless outputs, such as language modeling, the computation does not depend on the scale state, and we implement the residual stream using only canonical coordinates. App.~\ref{app:controls} gives the scale parameterization.

\paragraph{Scope and guarantees.} On nonzero inputs with zero canonicalization floors, $\canon(ch)=\canon(h)$ for $c>0$ makes the dimensionless network output invariant at every block-input boundary, including $h_0$, and at the final block-output boundary (App.~\ref{app:exactgauge}). Canonical coordinates preserve continuous invariant expressivity under independent positive rescalings; targets depending on absolute magnitude or dimensionless scale ratios require retaining that information (Apps.~\ref{app:factorization} and~\ref{app:coupledunits}).

\section{Experiments}
\label{sec:exp}

We test whether \gct{}'s explicit scale control makes learning-rate and depth transfer more reliable, keeps predictions unchanged under internal rescaling, and preserves performance when residual states are stored at low precision. The practical goal is to obtain these benefits while retaining competitive clean validation loss.

\subsection{Setup}
\label{sec:exp-setup}

We evaluate decoder-only Transformers with 51M, 124M, 350M, and 1B parameters to test \gct{} across a broad range of model sizes. The baselines include blocks modeled on Large Language Model Meta AI (LLaMA). Within each comparison, we train the architectures on FineWeb-Edu using the same data order, token budget, and optimization recipe. We test higher learning rates in the 51M models and transfer the 51M hyperparameters to deeper networks. We compare clean loss and internal-rescaling invariance at 124M, and repeat quantization tests at 350M and 1B. At 124M, \gct{} ranks third of eight architectures by unrounded three-seed mean loss. \gct{}'s mean loss is within $0.005$ nats of LayerScale, nGPT, and Pre-LN and $0.023$ nats above QK-Norm, which makes these models comparable in clean validation performance before the robustness tests (Table~\ref{tab:clean124m}). To test transfer across model widths and compatibility with another optimizer, we also evaluate width transfer and the momentum orthogonalized by Newton--Schulz (Muon) optimizer. To test the principles beyond language modeling, we include two vision benchmarks and PINNs. App.~\ref{app:details} specifies the architectures, recipes, seeds, and evaluation protocols.

\subsection{Learning-rate robustness}
\label{sec:exp-lr}

\begin{figure}[t!]
\centering
\includegraphics[width=\linewidth]{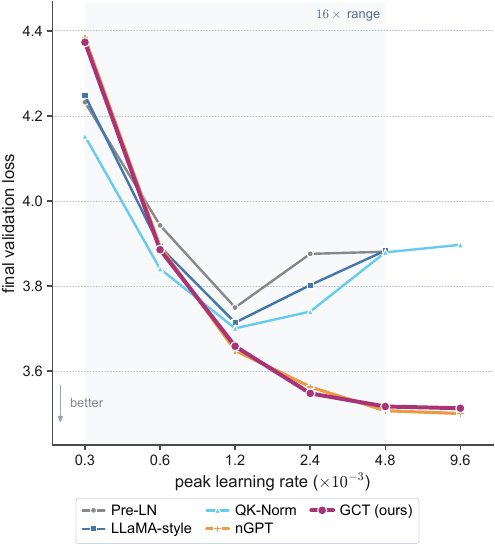}
\caption{Final validation loss as a function of peak learning rate for primary-seed 51M models trained on 600M tokens. \gct{} and nGPT retain low loss at high rates that worsen the other three architectures, showing a wider useful learning-rate range.}
\label{fig:lrsweep}
\end{figure}

Internal scale drift can make the same learning rate behave differently during training. \gct{} limits this variation by keeping the residual state normalized before each attention and MLP computation. We test whether \gct{} can train stably at higher learning rates while continuing to reduce validation loss.

\gct{} and nGPT remain stable across the full $16\times$ learning-rate range and retain low validation loss at the highest rates (Table~\ref{tab:sweep} and Figure~\ref{fig:lrsweep}). Pre-LN and LLaMA-style deteriorate beyond the respective best sampled rates. QK-Norm finishes training at every sampled rate without numerical divergence, but QK-Norm's upper-end loss is $0.18$ nats above the architecture's best sampled loss. At the additional rate $9.6\times10^{-3}$, \gct{} and nGPT improve further, while QK-Norm's loss is $0.197$ nats above the architecture's best sampled loss. At this rate, we stop the Pre-LN and LLaMA-style runs at step 100, about $8\%$ of the training budget, when the runs' early UVI scores exceed the preset stopping threshold (App.~\ref{app:details}). Both normalized models continue improving through this highest tested rate, placing both models' best measured results at the upper boundary of the sweep. \gct{} maintains low validation loss across a wide range of high learning rates, including rates that worsen validation performance in the other three architectures.

\gct{} improves by $0.125$ nats and nGPT by $0.127$ nats when the learning rate selected at 51M is applied unchanged at 124M, whereas the other baselines improve by $0.052$ to $0.072$ nats (Table~\ref{tab:transfer}). Selection uses the original five-point $16\times$ grid. The additional $9.6\times10^{-3}$ point is a separate stress test. Every architecture receives a separate sweep and transfer. QK-Norm reaches $3.091$ at the QK-Norm-selected rate of $1.2\times10^{-3}$, while nGPT and \gct{} reach $3.043$ and $3.045$, respectively, at $4.8\times10^{-3}$. The normalized architectures achieve the lowest losses under this transfer protocol. For \gct{}, this transfer result shows that a useful rate can be selected on the smaller model and reused at the larger size without retuning.

\begin{table}[t!]
\centering\small
\caption{Validation loss in fixed-budget depth transfer with the primary seed, 400M tokens, and hyperparameters transferred unchanged from $L=8$. \gct{} achieves the lowest loss at both depths, and adding the \gct{} projection also improves nGPT at $L=48$.}
\label{tab:depth}
\begin{tabular}{lcc}
\toprule
architecture & $L=24$ & $L=48$\\
\midrule
Pre-LN & 4.397 & 4.598 \\
LLaMA-style & 4.322 & 4.513 \\
QK-Norm & 4.206 & 4.322 \\
\midrule
nGPT & 3.996 & 4.243 \\
\gct{} (ours) & 3.973 & 3.903 \\
\addlinespace
\gct{} without $P^\perp$ & 3.983 & 3.938 \\
nGPT with $P^\perp$ & 4.005 & 3.968 \\
\bottomrule
\end{tabular}
\end{table}

\subsection{Depth transfer and architectural ablations}
\label{sec:exp-ngpt}
nGPT is the closest architectural baseline because both models compute with normalized states and are scale-invariant in the zero-floor formulations of \gct{} and nGPT. We test the direction of the added residual contribution because the analysis in \secref{sec:dynamics} shows that subtracting the component along the state removes first-order magnitude growth. This orthogonalization permits per-layer steps of order $L^{-1/2}$ while keeping the cumulative squared step size bounded. To test depth transfer without retuning, we apply the $L=8$ hyperparameters unchanged at $L=24$ and $L=48$. \gct{} remains stable in this primary-seed probe and reaches a validation loss of \depthgctln{} at $L=48$ with a fixed 400M-token budget, compared with \depthngptln{} for nGPT (Table~\ref{tab:depth}).

\begin{table*}[t!]
\centering\small
\caption{PTQ at 1B with $d=1792$, $L=24$, 4B tokens, and the primary seed. Base is the unquantized 16-bit brain floating-point (bf16) loss on a fixed 32-batch subset, while other entries are loss increases in nats. Per-channel 4-bit weight quantization (W4) changes only weights. In 8-bit weight-and-activation quantization (W8A8), linear-input activations use static scales. Stream columns use static per-layer tensor scales or dynamic per-token scales at every block boundary. The logit head remains in full precision. \gct{} has the smallest stream loss increase with per-tensor int8 and per-token int4 quantization, while per-tensor int4 causes large loss increases for every architecture.}
\label{tab:quant1b}
\begin{tabular}{lccccccc}
\toprule
 & & \multicolumn{2}{c}{weights and activations} & \multicolumn{2}{c}{stream $h\to\mathrm{int8}$} & \multicolumn{2}{c}{stream $h\to\mathrm{int4}$}\\
\cmidrule(lr){3-4}\cmidrule(lr){5-6}\cmidrule(lr){7-8}
architecture & base & W4 & W8A8 & tensor & token & tensor & token\\
\midrule
Pre-LN & 2.759 & +0.045 & +0.028 & +3.018 & +0.049 & +6.390 & +7.101 \\
LLaMA-style & 2.744 & +0.032 & +0.348 & +5.513 & +0.030 & +7.117 & +5.474 \\
QK-Norm & 2.737 & +0.029 & +0.833 & +5.426 & +0.021 & +8.074 & +5.230 \\
\midrule
nGPT & 2.751 & +0.038 & +0.075 & +0.032 & +0.008 & +19.083 & +4.661 \\
\gct{} (ours) & 2.756 & +0.028 & +0.079 & +0.009 & +0.001 & +9.517 & +0.739 \\
\bottomrule
\end{tabular}
\end{table*}

We remove proposal canonicalization or the orthogonal projection $P^\perp$ from \gct{} to measure how each component affects validation loss. At $L=8$, removing proposal canonicalization costs $+0.6$ nats, whereas removing $P^\perp$ has negligible effect (Table~\ref{tab:seeds-depth-abl}). In the primary-seed depth cells, \noprojdepthclaim{} (Table~\ref{tab:depth}). Proposal normalization is important already in the shallow model, while the projection's benefit grows with depth in these probes. Figure~\ref{fig:qual-depth-schedule} connects this depth comparison to the measured residual step sizes and alignments.

To test whether the projection also helps an existing normalized architecture, we add $P^\perp$ to nGPT's update direction before nGPT's per-dimension step, removing the direction's component along the current state. \ngptpperptransplant{} (Table~\ref{tab:depth}). The gain at $L=48$ shows that this NDA-derived component can also improve depth transfer beyond \gct{} under the same fixed-budget protocol.

\subsection{Invariance under internal rescaling}
\label{sec:exp-perturb}

We test sensitivity to internal rescaling by multiplying a hidden state by a positive factor, $h\to ch$, at the embedding output or at mid-depth. This rescaling changes the state's magnitude while preserving the state's direction and leaves the trained weights fixed. For \gct{} and nGPT, validation loss is unchanged to the reported precision and 32-bit floating-point (fp32) block invariance error remains near the numerical floor over $c\in[10^{-2},10^2]$ (App.~\ref{app:rescalingaudits}, Figure~\ref{fig:perturb}, and Table~\ref{tab:perturb}). The trained \gct{} model retains validation performance across these internal scale changes without retraining.

\paragraph{Weight-space fingerprint.}
\label{sec:exp-pgauge}
Prop.~\ref{prop:pgauge} predicts zero radial sensitivity $\langle W,\nabla_W\Lcal\rangle_F$ when downstream computation cancels rescaling of a matrix's output. All \pgmatch{} architecture-by-role predictions match across the eight trained 124M models, and the 1B audit likewise matches \pgmatchoneb{} cells. Predicted-null cells lie at the fp32 floor. App.~\ref{app:rescalingaudits} gives the measured sensitivities and rolewise interpretation.

\subsection{Post-training quantization (PTQ) robustness}
\label{sec:exp-quant}

Scale invariance removes pure changes in magnitude, but quantization can also change a hidden state's direction through rounding (Thm.~\ref{thm:tangent} and Cor.~\ref{cor:quantfunctional}). We test whether \gct{}'s normalized residual stream (\secref{sec:gct}) limits this additional error and preserves validation performance when states are stored at low precision. NDA predicts that reducing token-to-token scale variation lowers the error associated with a shared quantization scale. To test this prediction, we apply PTQ to the residual stream at every block boundary, as in hidden-state caching, offloading, or transmission. In a pre-norm network, the skip path carries the stored tensor, including the rounding errors, into subsequent residual additions.

Prop.~\ref{prop:quant} and Thm.~\ref{thm:quantminimax} show that, under the stated assumptions and high-resolution regime, the worst-case downstream output error grows as $\Theta(\kappa_{\mathrm{scale}}/q)$, where $\kappa_{\mathrm{scale}}$ is the ratio of the largest token RMS to the smallest and $q=2^{b-1}-1$. This dependence remains even when subsequent normalization cancels pure changes in magnitude exactly. In effective-resolution terms, cross-token scale variation consumes $\log_2\kappa_{\mathrm{scale}}$ bits. The zero-floor \gct{} stream fixes every token's RMS at one, setting $\kappa_{\mathrm{scale}}=1$ and removing this source of resolution loss. We compare one scale for the entire tensor with one scale per token to test whether adapting the quantization spacing to each token's magnitude reduces the baselines' error and narrows the error gap to \gct{}.

We use weight-only, linear-input activation, and key--value (KV) cache quantization as controls to identify where scale causes the largest errors. The strongest separation between normalized streams and the other baselines occurs when the residual stream uses a shared tensor scale (Tables~\ref{tab:quant1b} and~\ref{tab:quant}). Per-token scaling sharply reduces the baselines' int8 loss increases and narrows the gap to \gct{}, showing that explicit per-token scale management compensates for much of the measured error in the unnormalized streams.

At 1B, calibrated per-tensor stream int8 increases validation loss by \quantgcthstatoneb{} nats for \gct{}, while the largest baseline increase is \quantworsthstatoneb{} nats. Under per-token 4-bit integer (int4) quantization, \gct{} limits the increase to $+0.739$ nats, compared with $+4.661$ for nGPT and at least $+5.230$ for the other baselines (Table~\ref{tab:quant1b}). On the independent second 1B seed, static per-tensor stream int8 changes the worst baseline, \sqtwoworstname{}, by $+\sqtwoworsthstat{}$ nats and \gct{} by $+\sqtwogcthstat{}$. Under per-token int4, \gct{} changes by $+\sqtwogcthfourtok{}$, compared with at least $+\sqtwobasehfourtokmin{}$ for the other non-nGPT baselines.

\gct{} combines the int8 benefit of normalized token states with substantially lower per-token int4 degradation than nGPT in this test. Both models have $\kappa_{\mathrm{scale}}\approx1$, and $\kappa_{\mathrm{scale}}=1$ exactly in the zero-floor limit. Equal token RMS does not control within-vector amplitude, downstream directional sensitivity, or error accumulation across blocks. The cross-token bound alone cannot explain the models' different int4 loss increases. \gct{} achieves the smaller int4 loss increase with per-token scales. Per-tensor int4 causes large loss increases for every architecture.

On the independently trained second-seed 1B models, we evaluate the same residual-stream sites and granularities through zero-shot accuracy averaged over four tasks (Table~\ref{tab:quantds}). Dynamic per-tensor int8 changes mean accuracy by $-0.1$ points for \gct{} and $-0.8$ for nGPT, compared with $-19.6$ to $-22.7$ for the other baselines. Per-token int8 limits those baseline changes to $-0.5$ to $-1.9$ points. Under per-token int4, \gct{} changes by $-7.4$ points, nGPT by $-17.4$, and the other baselines by $-22.6$ to $-23.1$.

\subsection{PINNs under changes of physical units}
\label{sec:exp-pinn-qual}

We use PINNs to test NDA's loss-aggregation rule on data-fitting and partial differential equation (PDE) residual losses, which can carry different physical units. Following \secref{sec:rules} and Thm.~\ref{thm:lossscalar}, we divide each loss by a characteristic scale with matching units to keep the total objective unchanged under a unit conversion. We derive these scales from the measured magnitude of the data or forcing and the units introduced by the PDE operator, and use the scaled output head of \secref{sec:gct}. We then re-express the same Burgers and Helmholtz problems in different units to test whether this construction preserves physical solution accuracy without retuning the loss weights.

Across the $10^4$ unit sweep, NDA's scaled output head and loss weights keep the worst relative $L_2$ error at \pinnunitworst{}, whereas equal or adaptive weights degrade for at least one rescaling (Table~\ref{tab:pinn}, five seeds). Figure~\ref{fig:qual-pinn-fields} visualizes the recovered Burgers fields. These results show that the loss-aggregation rule, applied together with the scaled output head, preserves accurate physical solutions across the tested unit systems.

Additional diagnostic, width-transfer, optimizer, vision, and deployment results are collected in App.~\ref{app:full}.

\FloatBarrier
\section{Conclusion}

Our NDA framework establishes a typing principle for scale management across the computation graph. The NDA rules unify the scale-control mechanisms used in Transformers, including normalization, residual scaling, attention temperature, loss weighting, and parameter updates. UVI diagnostics and the weight-space audit make these rules testable and actionable by locating scale mismatches and checking the rescaling behavior predicted by NDA. \gct{} implements these scale-control rules in a Transformer architecture, and the typed multi-loss objective makes each loss dimensionless before combining the terms. Experiments spanning 51M to 1B parameters show that \gct{} achieves learning-rate and depth transfer, internal-rescaling invariance, and improved residual-stream quantization robustness while retaining competitive clean validation performance. Vision and PINN studies support the use of NDA beyond language modeling. These results demonstrate the benefits of explicit scale management grounded in NDA for neural network design.

\clearpage
\section*{AI Use Statement}
Generative AI tools were used as assistive tools for writing and editing, mathematical proofreading and consistency checking, and code development and debugging. Any AI-assisted outputs incorporated into this work were reviewed and validated by the authors as appropriate. The authors take full responsibility for the final content of this work, including any text, claims, or artifacts produced with the assistance of generative AI.

\bibliographystyle{apalike}
\bibliography{nda}

\section*{Checklist}
\label{sec:checklist}

\begin{enumerate}
  \item For all models and algorithms presented, check if you include:
  \begin{enumerate}
    \item A clear description of the mathematical setting, assumptions, algorithm, and/or model. [Yes] See Secs.~\ref{sec:units}--\ref{sec:gct}.
    \item An analysis of the properties and complexity (time, space, sample size) of any algorithm. [Yes] See Appendices~\ref{app:supptheory} and~\ref{app:quantresults}.
    \item (Optional) Anonymized source code, with specification of all dependencies, including external libraries. [Yes] See Appendix~\ref{app:gctcode}.
  \end{enumerate}

  \item For any theoretical claim, check if you include:
  \begin{enumerate}
    \item Statements of the full set of assumptions of all theoretical results. [Yes] See Sec.~\ref{sec:units} and Appendices~\ref{app:supptheory} and~\ref{app:rescalingaudits}.
    \item Complete proofs of all theoretical results. [Yes] See Appendices~\ref{app:deriv} and~\ref{app:proofs}.
    \item Clear explanations of any assumptions. [Yes] See Sec.~\ref{sec:units} and Appendices~\ref{app:supptheory}--\ref{app:deriv}.
  \end{enumerate}

  \item For all figures and tables that present empirical results, check if you include:
  \begin{enumerate}
    \item The code, data, and instructions needed to reproduce the main experimental results (either in the supplemental material or as a URL). [Yes] See Appendix~\ref{app:gctcode}.
    \item All the training details (e.g., data splits, hyperparameters, how they were chosen). [Yes] See Appendix~\ref{app:details}.
    \item A clear definition of the specific measure or statistics and error bars (e.g., with respect to the random seed after running experiments multiple times). [Yes] See Appendix~\ref{app:controls} and the captions.
    \item A description of the computing infrastructure used. (e.g., type of GPUs, internal cluster, or cloud provider). [Yes] See Appendix~\ref{app:fullsetup}.
  \end{enumerate}

  \item If you are using existing assets (e.g., code, data, models) or curating/releasing new assets, check if you include:
  \begin{enumerate}
    \item Citations of the creator If your work uses existing assets. [Yes] See Appendix~\ref{app:details}.
    \item The license information of the assets, if applicable. [No] Licenses are not listed.
    \item New assets either in the supplemental material or as a URL, if applicable. [No] No separate asset release.
    \item Information about consent from data providers/curators. [Not Applicable] Existing public datasets only.
    \item Discussion of sensible content if applicable, e.g., personally identifiable information or offensive content. [No] Not discussed in this paper.
  \end{enumerate}

  \item If you used crowdsourcing or conducted research with human subjects, check if you include:
  \begin{enumerate}
    \item The full text of instructions given to participants and screenshots. [Not Applicable] No crowdsourcing or human subjects.
    \item Descriptions of potential participant risks, with links to Institutional Review Board (IRB) approvals if applicable. [Not Applicable] No human subjects.
    \item The estimated hourly wage paid to participants and the total amount spent on participant compensation. [Not Applicable] No participants.
  \end{enumerate}
\end{enumerate}

\clearpage
\appendix
\onecolumn
\thispagestyle{empty}
\aistatstitle{Neural Dimensional Analysis:\\ Foundations of Dimensionless and Scale-Consistent Computation\\ Appendix}
\numberwithin{equation}{section}

\startcontents[apx]
\section*{Appendix Table of Contents}
\begingroup
\footnotesize
\printcontents[apx]{l}{1}{\setcounter{tocdepth}{1}}
\endgroup
\clearpage

\section{Notation}
\label{app:notation}

Table~\ref{tab:notation} collects the symbols that recur across the paper. The subscripts $\ell$, $h$, $t$, $i$, and $b$ index blocks, attention heads, tokens, loss terms, and parameter blocks, respectively.

\begin{table}[t!]
\centering\small
\caption{List of notation. Recurring symbols are grouped by the part of the calculus that introduces each symbol. A bar denotes a dimensionless canonical coordinate, and a star denotes a target value of a diagnostic.}
\label{tab:notation}
\begin{tabular}{ll}
\toprule
Symbol & Meaning\\
\midrule
\multicolumn{2}{l}{\textbf{Units and the rescaling group}}\\
$G$ & declared rescaling group $(\mathbb{R}_+)^K$ of the graph\\
$g$, $c$ & group element; a single positive rescaling factor\\
$\rho_v$ & neural unit of graph node $v$, a representation of $G$\\
$\chi_u$ & scalar character $g\mapsto\prod_kg_k^{u_k}$ with exponent $u$\\
$\uof{x}$ & unit type of $x$; $\uof{x}=1$ means dimensionless\\
$U_{\mathrm{res}}$, $U_\theta$ & residual-stream unit and parameter unit\\
$s_x$ & empirical scale of $x$, with $\uof{s_x}=\uof{x}$\\
$\RMS(x)$ & RMS over a shared-unit block\\
$\delta_v$, $K_{vu}$ & local equivariance defect and Lipschitz gain at node $v$\\
\addlinespace
\multicolumn{2}{l}{\textbf{Canonical gauge and architecture}}\\
$d$, $d_h$, $L$ & model width, head dimension, number of blocks\\
$h_\ell$, $r_\ell$ & residual stream and branch contribution at block $\ell$\\
$s_\ell$, $\bar h_\ell$ & scale state and canonical coordinate, $h_\ell=s_\ell\bar h_\ell$\\
$\canon$, $\canon_0$ & canonicalizer and zero-floor limit $x/\RMS(x)$\\
$\mathbb{S}$, $d_{\mathbb{S}}$ & RMS-one canonical section and chord metric on that section\\
$P^{\perp}_{\bar h}$ & projector removing the component along $\bar h$\\
$\alpha_\ell$, $\alpha^{\star}$ & dimensionless residual step and target step size $1/\sqrt{2L}$\\
$c_\ell$ & cosine similarity between the branch contribution $r_\ell$ and stream $h_\ell$\\
$\tau_{\ell,h}$ & attention temperature $\sqrt{d_h}s_Qs_K$, a logit nondimensionalizer\\
$s_Q$, $s_K$ & per-head query and key coordinate scales\\
$A$, $N$, $\mathcal Q$ & unit-exponent matrix, basis of $\ker A$, complete invariant\\
\addlinespace
\multicolumn{2}{l}{\textbf{Losses, gradients, and optimizer}}\\
$\Lcal$, $\Lcal_i$ & dimensionless objective and constituent $i$th loss term\\
$\lambda_i$ & loss weight, a preference $\alpha_i$ over a characteristic scale $S_i$\\
$\nabla_x\Lcal$ & gradient of the loss, carrying the dual unit $\uof{x}^{-1}$\\
$\Pi_x$ & primal--dual RMS product $\RMS(x)\RMS(\nabla_x\Lcal)$\\
$\theta$, $\Delta\theta$ & parameters and optimizer displacement\\
$P_\theta$, $\eta$ & preconditioner with $\uof{P_\theta}=U_\theta^2$; dimensionless step\\
$p(r)$ & radial preconditioner; gauge-covariant iff $p(r)\propto r^2$\\
$\vartheta$ & rotation angle of the exact great-circle quotient update\\
$\rho_b$ & relative update of parameter block $b$\\
\addlinespace
\multicolumn{2}{l}{\textbf{Diagnostics, perturbation, and quantization}}\\
$\UVI_{\mathrm{res}}$, $\UVI_{\mathrm{attn}}$ & log deviation of $\alpha_\ell$ and of logit variance from target\\
$\UVI_{\mathrm{act}}$, $\UVI_{\mathrm{grad}}$ & depth dispersion of preactivation scale and of $\Pi_\ell$\\
$\UVI_{\mathrm{opt}}$ & cross-block imbalance of the relative updates $\rho_b$\\
$e$, $a$, $\rho$ & perturbation, relative radial coefficient, and tangential size\\
$b$, $q$ & bit width and half-range $2^{b-1}-1$ of the integer grid\\
$\Delta$, $Q_\Delta$ & shared quantization step and associated absmax quantizer\\
$\kappa_{\mathrm{scale}}$ & cross-token spread $\max_tr_t/\min_tr_t$ of the stream RMS $r_t$\\
$\kappa_{\mathrm{amp}}$ & within-vector amplitude ratio $\max_t\|h_t\|_\infty/r_t$\\
$\kappa_{\mathrm{depth}}$ & span of the mean token RMS across block inputs\\
$\mathfrak E_q(\Gamma)$ & worst-case quotient error at scale spread at most $\Gamma$\\
$\kappa$ & rescaling of the data's unit system in the PINN study\\
\bottomrule
\end{tabular}
\end{table}

\section{Limitations and scope}
\label{app:limits}

The theory applies to the declared rescaling group, and the experiments test the theory's consequences in controlled settings. The language-model study spans decoder-only Transformers from 51M to 1.0B parameters trained on FineWeb-Edu under one shared recipe, with the main controlled comparison at 124M and 2.75B tokens, and the generality probes cover two vision benchmarks and two PINN problems. Generality beyond these datasets and model families remains to be tested. The larger models support the deployment claim of \secref{sec:exp-quant}. Compute-optimal scaling requires a separate study. Each setting compares typed and untyped models at a matched budget. Extending these comparisons to further corpora, scales, and architectures is natural future work.

\section{Related work}
\label{app:related}

\subsection{Normalization, residual scaling, and attention temperature}
Batch normalization established the practice of restandardizing internal activations \citep{ioffe2015batchnorm}, and LayerNorm and RMSNorm carry that practice into the Transformer \citep{ba2016layer,zhang2019rmsnorm}, where Pre-LN placement stabilizes depth \citep{xiong2020layer} and NormFormer adds post-attention and mid-MLP normalization with head scaling \citep{shleifer2021normformer}. Residual connections make depth trainable in the first place \citep{he2016resnet}, and DeepNet, ReZero, SkipInit, LayerScale, and Fixup control residual-branch magnitude to sustain that trainability \citep{wang2022deepnet,bachlechner2020rezero,de2020skipinit,touvron2021cait,zhang2019fixup}. A complementary line removes the normalizer instead of tuning the normalizer's placement, through a learnable elementwise squashing \citep{zhu2025dyt} or through analytically scaled residual blocks with adaptive gradient clipping \citep{brock2021nfnets}.

The same accounting governs the attention interface. The $1/\sqrt{d_h}$ factor is the $s_Q=s_K=1$ case of the logit nondimensionalizer $\tau=\sqrt{d_h}s_Qs_K$ \citep{vaswani2017attention}, QK-Norm and its use in a 22-billion-parameter Vision Transformer (ViT) make softmax inputs dimensionless by construction \citep{henry2020qknorm,dehghani2023vit22b}, $\sigma$Reparam bounds logit growth through weight reparameterization \citep{zhai2023stabilizing}, and controlled small-scale reproductions trace large-model training instabilities to attention-logit growth exposed through learning-rate sensitivity \citep{wortsman2024proxies}.

NDA interprets these mechanisms as scale controls at different interfaces. Normalization nondimensionalizes from $U_{\mathrm{res}}$ to $1$. Residual scaling controls the dimensionless step size $\alpha_\ell$, and attention scaling implements the softmax rule of \secref{sec:rules}. In a normalizer-free block, a bounded nonlinearity and a calibrated input scale can provide the corresponding nondimensionalization without a measured normalization statistic.

\subsection{Optimizers, parameterizations, and typed optimization}
Adaptive moment estimation (Adam) normalizes per-coordinate gradient scale \citep{kingma2015adam}, and layer-wise adaptive moments for batch training (LAMB) and layer-wise adaptive rate scaling (LARS) apply relative trust-ratio updates per block, which restore primal units in our terminology \citep{you2020lamb,you2017lars}. $\mu$P chooses width-dependent multipliers to keep the update effect invariant across widths and enable zero-shot hyperparameter transfer \citep{yang2021tensor,yang2021tp5}. Depth-wise $\mu$P scales residual-branch contributions as $1/\sqrt L$ to enable the same transfer across depth \citep{yang2024tp6}. Unit scaling chooses scales for low-precision formats \citep{blake2023unit}, generalizing the loss-scaling constant of mixed-precision training, which is an inserted unit conversion that keeps half-precision gradients representable \citep{micikevicius2018mixed}, and unit-scaled $\mu$P composes the two disciplines \citep{blake2024umup}.

A parallel line makes the metric on weight space explicit. The modular norm equips a network with a norm under which well-scaled updates transfer across width and depth \citep{large2024modular}, and modular duality with the efficient Muon realization casts the optimizer as a duality map that sends dual gradient vectors to primal update directions \citep{bernstein2024duality,jordan2024muon}.

NDA uses the same primal--dual relation, in which a gradient carries the inverse unit of the differentiated variable and an optimizer restores the primal unit. NDA expresses this primal--dual relation for scalar rescaling groups $G=(\mathbb R_+)^K$ as a typing discipline covering the full computation graph, including objectives and diagnostics. A modular norm can supply a shape-aware duality map, while NDA further requires the complete displacement, including the scalar step or reference scale of the update, to obey Prop.~\ref{prop:precondlaw}. Combining Muon with a scalar scale that obeys the required transformation law can make the full update unit-covariant. Our optimizer contribution is the covariance criterion, while the proposed constructions are the UVI diagnostics and the canonical architecture.

\subsection{Rescaling symmetries and normalized architectures}
A classical line studies weight rescalings that leave a network's function invariant. For positive-homogeneous activations, paired rescalings of incoming and outgoing weights form a symmetry group. Path-norm methods and path-normalized variants of stochastic gradient descent (SGD) optimize in the induced quotient \citep{neyshabur2015path}, and the same reparameterizations show that sharpness measures are not invariant \citep{dinh2017sharp}, imprint conservation laws on gradient flow \citep{kunin2021symmetry}, and define the orbits used by teleportation methods \citep{zhao2022teleport}.

Weight normalization exploits the same freedom constructively by splitting each weight vector into a direction and an explicit magnitude \citep{salimans2016weightnorm}, which is the radial coordinate that becomes irrelevant under the downstream-nondimensionalization hypothesis of Prop.~\ref{prop:pgauge}. A normalizer creates this freedom by making the upstream weight block scale-invariant. The scale-invariant weight block's norm then acts only through an effective learning rate \citep{vanlaarhoven2017l2,hoffer2018norm,arora2019autorate}, which App.~\ref{app:optimizergeometry} recovers as the $p\equiv1$ case of Prop.~\ref{prop:quotgeometry}, and weight decay on such a block is equivalent to a learning-rate schedule \citep{li2020exponential}. NDA follows how activation rescalings propagate through the graph. In an untyped graph, a rescaling can change the computed function. The typing rules identify the interfaces that absorb or propagate the rescaling. The two views meet at interfaces, because a weight-space rescaling of adjacent layers induces an activation rescaling between those layers.

The same normalized gauge also appears at the architecture level. nGPT constrains hidden states and weight rows to the unit hypersphere, with large reported convergence speedups \citep{loshchilov2024ngpt}, following earlier work that placed learned latent representations on the unit hypersphere \citep{xu2018spherical}, and the magnitude-preserving diffusion layers of \citet{karras2024edm2} hold activations and weights on fixed-norm shells throughout the backbone, an independent cross-domain instance of computing only in a normalized coordinate.

\gct{} shares the normalized-state mechanics but differs in derivation and content. \gct{} follows from a typing principle that also produces UVI diagnostics, UCT corrective modules, and a treatment of losses and optimizers, and the architecture adds an orthogonalizing projection $P^{\perp}$ motivated by the residual-drift analysis, a learnable dimensionless step initialized at $1/\sqrt{2L}$, and dimensionless gates. For unitful outputs such as regression and PINN predictions, \gct{} retains an explicit scale state to restore the output magnitude and units. The component ablations evaluate proposal canonicalization and the orthogonal projection (Tables~\ref{tab:depth} and~\ref{tab:seeds-depth-abl}).

\subsection{Quantization and activation-scale management}
Quantization research documents substantial variation in the scale of internal language-model activations. Integer inference represents tensors as scales times integer grids \citep{jacob2018quantization,nagel2021whitepaper}, connecting integer inference to the factorization in \secref{sec:gct}. Relevant phenomena include outlier features \citep{dettmers2022llmint8}, massive activations \citep{sun2024massive}, and attention sinks \citep{xiao2024sinks}. These observations motivate examining both cross-token scale variation and within-vector concentration.

Methods for low-bit inference include mixed-precision decomposition \citep{dettmers2022llmint8}, error-compensated weight rounding \citep{frantar2023gptq}, activation-scale migration \citep{xiao2023smoothquant}, activation-aware weight quantization \citep{lin2024awq}, and rotations that reduce coordinate outliers \citep{ashkboos2024quarot}. Logit soft-capping provides a complementary architectural control at attention and output interfaces \citep{riviere2024gemma}. These methods address different aspects of the scale and precision problem.

An architectural approach is taken by \citet{bondarenko2023quantizable}, who associate activation outliers with attention heads implementing a no-op and use clipped softmax and gated attention to improve int8 quantization.

NDA controls cross-token scale variation by constructing a canonical residual stream. Within-vector outliers remain a separate factor, $\kappa_{\mathrm{amp}}$, in Prop.~\ref{prop:quant}. Thm.~\ref{thm:quantminimax} shows that the cross-token penalty persists in a worst-case functional bound after radial rejection. The experiments compare this architectural control with per-token and per-channel stream scales and SmoothQuant migration. Normalized streams reduce the need for fine-grained scale management in these settings while retaining the ordinary per-tensor quantization cost.

\subsection{Typed objectives and units of physical data}
Uncertainty weighting \citep{kendall2018uncertainty}, gradient normalization \citep{chen2018gradnorm}, and gradient-statistics annealing for PINNs \citep{wang2021understanding} adapt loss weights during training, and practitioner guides for PINNs also recommend nondimensionalizing the problem before training \citep{wang2023expert}. NDA instead types the objective. A weight $\lambda_i$ factorizes into a dimensionless preference $\alpha_i$ and an inverse-unit converter $1/S_i$ with $\uof{S_i}=\uof{\Lcal_i}$. Thm.~\ref{thm:lossscalar} shows that every invariant linear rule whose weights depend on the characteristic scales carries this inverse-unit factor, with the remaining scale dependence confined to dimensionless $\pi$ groups, and Cor.~\ref{cor:lossunique} shows that the preferences are constant within this class when the loss units vary independently. Deriving $S_i$ from observable characteristic scales makes the objective invariant to the data's unit system without adaptive weighting. Adaptive methods can then adjust dimensionless preferences among the normalized terms.

This typing connects to the broader use of units in scientific machine learning, where dimensional analysis constrains learned models of physical systems \citep{villar2023dimensionless}, Buckingham--$\pi$ style reductions are classical \citep{buckingham1914}, and recent work learns the dimensionless groups from data \citep{bakarji2022buckingham}. Outside machine learning, units-of-measure type systems check the dimensional consistency of a program at compile time and infer unit-polymorphic types \citep{kennedy2010units}. All of these lines attach externally declared physical units to inputs, outputs, and program variables. NDA instead assigns units to internal tensors as transformation laws under internal rescalings, a gauge structure of the computation graph that applies when no physical unit exists. The gauge terminology follows geometric deep learning, where equivariance is stated relative to a choice of local frame rather than to a single global symmetry \citep{cohen2019gauge}.

\subsection{Positioning summary}
\label{app:positioning}

The related methods address complementary parts of scale management. Parameterization methods such as $\mu$P and unit scaling prescribe initialization, multipliers, and update scales \citep{yang2021tensor,yang2024tp6,blake2023unit}. Normalized architectures control activation magnitude \citep{loshchilov2024ngpt,karras2024edm2}. Weight-space symmetry analyses describe function-preserving reparameterizations \citep{neyshabur2015path,dinh2017sharp,kunin2021symmetry}. Modular norms and duality maps organize optimizer updates \citep{large2024modular,bernstein2024duality}. Loss-balancing and quantization methods address scale in objectives and finite-precision computation, respectively \citep{kendall2018uncertainty,chen2018gradnorm,dettmers2022llmint8,xiao2023smoothquant}. Units-of-measure type systems and units-equivariant learning provide the closest conceptual parallels for externally declared units \citep{kennedy2010units,villar2023dimensionless}.

Our NDA framework unifies these perspectives in one graph-level description. Units are transformation laws under declared internal rescalings. This definition lets NDA apply to tensors without physical units. Forward consistency determines reverse-mode units (Prop.~\ref{prop:dualsound}), while UVI diagnostics and the parameter-gauge audit make the declarations testable on trained networks. Through NDA, we also connect training-time scale control to PTQ. Our contribution is this unified treatment and the resulting diagnostics and constructions.

\section{Discussion}
\label{sec:discussion}

Our NDA framework connects a declared rescaling group to local computation rules, diagnostics, and architectural choices. The following discussion explains how to interpret these declarations and how the resulting constructions relate to UCT, nGPT, parameterization methods, and optimizers.

\paragraph{The status of a unit declaration.} A unit assignment specifies a rescaling group and the intended transformation of each quantity. UVI values describe deviations from the corresponding reference regime. The same declaration organizes the scale-control mechanisms, identifies mismatches, and guides the construction of \gct{}. The diagnostics characterize that regime, while the relation between UVI values and task quality is an empirical question.

\paragraph{Role of UCT in the comparison.} UCT tests local application of the typing rules through three corrective modules: an RMS-matching residual adapter, an empirical attention temperature, and normalized gates. At 124M, UCT's per-site tensor int8 loss increase is $+0.007$ nats (Table~\ref{tab:quant}), but UCT's clean loss of $3.256$ is the highest among the evaluated architectures (Table~\ref{tab:clean124m}). UCT also loses invariance under the strongest down-scaling intervention. The hidden state shrinks, but the small constants in UCT's normalization denominators and the stored UCT scale estimates remain fixed. Dividing by these unchanged quantities no longer cancels the change in magnitude. This comparison distinguishes the benefits of local converters from the guarantees of the canonical construction.

\paragraph{Relation to nGPT.} Because \gct{} and nGPT both use normalized states, we evaluate nGPT as a direct baseline across the sweep, depth, perturbation, quantization, and vision studies under the identical recipe. The language-model comparison between \gct{} and nGPT contrasts \gct{}'s orthogonalizing projection, proposal canonicalization, and scalar step initialized at $1/\sqrt{2L}$ with nGPT's per-dimension eigen learning rate. This language-model comparison concerns dimensionless computation, for which the implementation of \gct{} omits $s_\ell$, as stated in \secref{sec:gct}. Under the fixed-budget depth-transfer protocol at $L=48$ and 400M tokens, \gct{} reaches \depthgctln{}, compared with \depthngptln{} for nGPT (Table~\ref{tab:depth}). Under 1B int4 per-token residual-stream quantization, the two architectures separate more strongly (Table~\ref{tab:quant1b}), as discussed in \secref{sec:exp-quant}. Our contribution beyond nGPT is the derivation through typing rules that also cover losses, gradients, and optimizers, the UVI diagnostic layer, and the local-correction and canonical-design variants.

\paragraph{Relation to $\mu$P.} $\mu$P-style width transfer is orthogonal to and composable with NDA typing, and depth-wise $\mu$P \citep{yang2024tp6} plays the analogous role along depth. The width experiments of \secref{sec:exp-breadth} test this composition directly.

\paragraph{Optimizers and open extensions.} The optimizer contribution is a covariance law for complete updates. Modular-duality optimizers \citep{bernstein2024duality,jordan2024muon} realize the shape-aware duality-map component at scale, but satisfying NDA's rescaling law additionally requires a typed scalar step or reference scale. The Muon pairing experiment in \secref{sec:exp-breadth} evaluates compatibility with \gct{} under a shared training recipe. On the architecture side, the zero-floor construction fixes each token's pre-retraction RMS ratio to $|\alpha_\ell|$, and the chord displacement of the canonical state transition is likewise token-independent. Per-token dimensionless gates on $\alpha$ are a natural extension for tasks requiring residual step sizes that vary strongly across tokens. App.~\ref{app:limits} states the scope of the empirical study.

\section{Additional theoretical statements}
\label{app:supptheory}

\subsection{Quantitative local-to-global typing stability}

Real graphs also contain sources of inexact equivariance, including finite normalization floors, approximate kernels, and quantizers. Thm.~\ref{thm:robustsound} bounds the output error by propagating each local equivariance defect through the sum of the downstream path gains from the defective node. For a frozen canonicalization floor, the local defect is explicitly $O(\epsilon/\RMS(x)^2)$, whereas a correctly co-transforming floor has zero defect. The theorem gives quantitative control for an arbitrary DAG by separating a node's own typing defect from the downstream gains that can amplify that defect.

Fix $g\in G$ and one evaluation $x\coloneqq(x_v)_v$ of a finite DAG with source set $\mathcal S$ and output node $o$. Equip every node space with a norm $\|\cdot\|_v$. A nonsource node $v$ with parents $\mathrm{pa}(v)$ computes $x_v\coloneqq f_v((x_u)_{u\in\mathrm{pa}(v)})$. Let $\widetilde x$ be the graph evaluation obtained by replacing every source value $x_s$ by $\rho_s(g)x_s$. Suppose that for each nonsource $v$ there are constants $K_{vu}\ge0$ such that
\begin{equation}
\label{eq:nodelip}
\|f_v(z)-f_v(z')\|_v \le \sum_{u\in\mathrm{pa}(v)}K_{vu}\|z_u-z'_u\|_u .
\end{equation}
This bound need only hold on a set containing the two parent tuples used in the proof, $z\coloneqq(\widetilde x_u)_{u\in\mathrm{pa}(v)}$ and $z'\coloneqq(\rho_u(g)x_u)_{u\in\mathrm{pa}(v)}$. A global Lipschitz bound is sufficient but not required. Define the local equivariance defect
\begin{equation}
\label{eq:localdefect}
\delta_v(g,x)\coloneqq \|f_v((\rho_u(g)x_u)_{u\in\mathrm{pa}(v)}) -\rho_v(g)x_v\|_v .
\end{equation}
Let $\mathcal P(v,o)$ be the set of directed paths from $v$ to $o$. This set includes the empty path when $v=o$. Then

\begin{theorem}[Quantitative local-to-global typing stability]
\label{thm:robustsound}
Under the preceding setup,
\begin{equation}
\label{eq:pathgain}
\|\widetilde x_o-\rho_o(g)x_o\|_o \le \sum_{v\notin\mathcal S}\delta_v(g,x) \sum_{p\in\mathcal P(v,o)} \prod_{(a\to b)\in p}K_{ba}.
\end{equation}
The proof is given in App.~\ref{app:robustproof}.
\end{theorem}

This theorem also quantifies the finite-floor implementation error in \secref{sec:gct}. For the floored canonicalizer $\canon_\epsilon(x)\coloneqq x/\sqrt{\RMS(x)^2+\epsilon}$ with a fixed numerical $\epsilon\ge0$, a correctly typed floor transforms as $\epsilon\mapsto c^2\epsilon$ and makes the local equivariance defect exactly zero. A frozen implementation floor instead leaves an explicit $O(\epsilon/r^2)$ local defect at $r\coloneqq\RMS(x)>0$, bounded in \ref{eq:canondefect} and derived in App.~\ref{app:floordefect}, which \ref{eq:pathgain} propagates to the output. The bound separates an architectural mismatch from the numerical error caused by a frozen floor.

\subsection{Invariant loss scalarization}
\label{app:lossweights}

\paragraph{Characteristic loss scales.} In \ref{eq:lossfact}, each factor $1/S_i$ converts the corresponding loss to a dimensionless value. The converter can be derived from characteristic scales of the data and operators. For PDE residuals, $S_i$ must account for the operator's output unit, including the units introduced by differentiation. For example, a squared first-order-in-time residual loss uses the characteristic loss scale $S_i\coloneqq(s_u/t^{\ast})^2$.

Thm.~\ref{thm:lossscalar} characterizes all loss weights that preserve this invariance within the class of linear objectives whose weights depend only on the characteristic scales. Every such weight must contain the factor $1/S_i$, with any remaining scale dependence restricted to dimensionless combinations of scales called Buckingham--$\pi$ groups. When the loss units vary independently, Cor.~\ref{cor:lossunique} fixes the preferences to constants and gives $w_i(S)=\alpha_i/S_i$.

To characterize every invariant linear scalarization whose weights depend only on the declared characteristic scales, let $G\coloneqq(\mathbb R_+)^K$. Write $g\coloneqq e^t$ coordinatewise, let $a_i\in\mathbb R^K$ be the character exponent of loss $i$, let $A\in\mathbb R^{K\times m}$ be the matrix with columns $a_1,\ldots,a_m$, and set $D(t)\coloneqq\operatorname{diag}(e^{a_1^\top t},\ldots,e^{a_m^\top t})$. Suppose the loss vector $\ell\in\mathbb R^m$ and the corresponding positive characteristic-scale vector $S\in\mathbb R_+^m$ both transform as $(\ell,S)\mapsto(D(t)\ell,D(t)S)$. Consider an arbitrary scale-dependent linear scalarization $J_S(\ell)\coloneqq\sum_{i=1}^m w_i(S)\ell_i$.

\begin{theorem}[Characterization of characteristic-scale-dependent invariant loss scalarization]
\label{thm:lossscalar}
Let $s\coloneqq\operatorname{rank}A$ and let the columns of $N\in\mathbb R^{m\times(m-s)}$ be a basis of $\ker A$. Under the preceding setup, $J_S$ is unit-system invariant for every loss vector,
\begin{align*}
J_{D(t)S}(D(t)\ell)=J_S(\ell), \text{ for all }S\in\mathbb R_+^m,\ell\in\mathbb R^m,t\in\mathbb R^K,
\end{align*}
if and only if the scalarization weights have the form
\begin{equation}
\label{eq:losscomplete}
w_i(S)=\frac{1}{S_i} \psi_i(N^\top\log S), \qquad i=1,\ldots,m,
\end{equation}
for arbitrary functions $\psi_i:\mathbb R^{m-s}\to\mathbb R$, where $\log S$ is componentwise. The proof is given in App.~\ref{app:lossscalarproof}.
\end{theorem}

The quotient coordinates $N^\top\log S$ form a basis for the logarithms of all dimensionless monomials $\prod_jS_j^{n_j}$ with $n\in\ker A$.

Any scale dependence of a unit-invariant adaptive preference must pass through these dimensionless $\pi$ groups.

\begin{corollary}[Independent loss units force inverse-scale weights]
\label{cor:lossunique}
Under the assumptions of Thm.~\ref{thm:lossscalar}, if the $m$ loss-unit characters are independent, meaning $\operatorname{rank}A=m$, the quotient has no nonconstant dimensionless scale group and
\begin{equation}
\label{eq:lossunique}
w_i(S)=\alpha_i/S_i
\end{equation}
with constants $\alpha_i$ is the unique invariant linear rule in this $S$-dependent class. The proof is given in App.~\ref{app:lossscalarproof}.
\end{corollary}

\subsection{Optimizer covariance and quotient geometry}
\label{app:optimizergeometry}

\paragraph{Relative optimizer updates.} Adam's $m_t/(\sqrt{v_t}+\epsilon_g)$ is dimensionless when $\uof{\epsilon_g}=U_\theta^{-1}$. Multiplying this ratio by a parameter scale restores the update unit, giving the relative form $\Delta\theta=-\eta s_\theta m_t/(\sqrt{v_t}+\epsilon_g)$ with $\uof{s_\theta}=U_\theta$. LARS \citep{you2017lars} and LAMB \citep{you2020lamb} implement this relative-scaling idea through trust ratios based on parameter norms. The comparable diagnostic across blocks is the relative update $\rho_b\coloneqq\|\Delta\theta_b\|/(s^{\mathrm{ref}}_{\theta,b}+\epsilon_{\theta,b})$.

These rules enable us to distinguish a change of parameter units from a change in the optimization trajectory. For an invertible parameter rescaling $R$, displacement covariance is equivalent to the preconditioner law $P^g=RPR^\top$ (Prop.~\ref{prop:precondlaw}). On a scale-invariant block, Prop.~\ref{prop:quotgeometry} shows that the induced quotient update is an exact great-circle step, and Thm.~\ref{thm:quotopt} and Cor.~\ref{cor:quotoptunique} show that $p(r)\propto r^2$ is the unique isotropic radial preconditioner that removes dependence on the arbitrary parameter gauge. In this setting, the squared-scale factor makes the update take the same angular step at every rescaling of the parameter vector.

To make the optimizer rule precise, let $R\in GL(\Theta)$ be an invertible rescaling of a finite-dimensional real inner-product parameter space. Suppose corresponding gradients satisfy $q^g=R^{-\top}q$, as the gradients do when a scalar objective is invariant under $\theta^g=R\theta$. Consider preconditioned displacements $u\coloneqq-\eta Pq$ and $u^g\coloneqq-\eta P^gq^g$ with the same dimensionless learning rate $\eta$.

\begin{proposition}[Covariance law for preconditioned updates]
\label{prop:precondlaw}
Under the preceding setup, $u^g=Ru$ for every $q$ if and only if
\begin{equation}
\label{eq:precondlaw}
P^g=R P R^\top.
\end{equation}
The proof is given in App.~\ref{app:precondproof}.
\end{proposition}

Carrying squared parameter units is both necessary and sufficient for the preconditioner to produce a covariant displacement.

The same distinction applies to nonlinear direction maps. A map $\mathcal D$ from gradients to displacements is unit-covariant precisely when the transformed direction map satisfies $\mathcal D^g(R^{-\top}q)=R\mathcal D(q)$. A duality map satisfies this law only when the resulting full displacement, including any scalar step, co-transforms with the parameters. For example, under a scalar rescaling $R=cI$, a scale-zero-homogeneous orthogonalization obeys $\mathcal D(c^{-1}q)=\mathcal D(q)$, whereas displacement covariance requires $c\mathcal D(q)$. Such an update becomes typed only after supplying an explicit primal-scale factor or an equivalently transformed scalar step.

Two immediate consequences clarify the operator law. Ordinary Euclidean SGD with a fixed common learning rate satisfies that law for every gradient only when $RR^\top=I$. A scalar parameter rescaling $R=cI$ instead requires the transformed learning rate $\eta^g=c^2\eta$.

The preconditioner law in Prop.~\ref{prop:precondlaw} is an operator statement. For the common case of a scale-invariant parameter block and a scalar radial preconditioner, the following proposition and theorem show that the quotient update is an exact geodesic step. These results also show that the squared parameter scale is the unique radial scalar correction that makes that step independent of the arbitrary parameter gauge.

Let $V$ be a finite-dimensional real inner-product space of dimension at least two and let $J:V\setminus\{0\}\to\mathbb R$ be continuously differentiable and scale-invariant, $J(cw)=J(w)$ for every $c>0$. Writing $g(w)\coloneqq\nabla J(w)$, differentiation with respect to $w$ and to $c$ gives
\begin{equation}
\label{eq:scalegradlaw}
g(cw)=c^{-1}g(w), \qquad \langle w,g(w)\rangle=0.
\end{equation}

Fix $a>0$, $\eta>0$, and a scalar radial preconditioner $p:(0,\infty)\to(0,\infty)$. Writing $r\coloneqq\|w\|$, consider
\begin{equation}
\label{eq:radialupdate}
w^+\coloneqq a w-\eta p(r)g(w).
\end{equation}

\begin{proposition}[Exact geometry of a radial quotient update]
\label{prop:quotgeometry}
Under the preceding setup, at every nonstationary point, with $\bar w\coloneqq w/\|w\|$ and $\bar g\coloneqq g(w)/\|g(w)\|$, the canonical update is exactly
\begin{equation}
\label{eq:quotgeodesic}
\frac{w^+}{\|w^+\|} =\cos\vartheta\bar w-\sin\vartheta\bar g, \qquad \tan\vartheta =\frac{\eta p(r)\|g(w)\|}{a r}, \quad 0<\vartheta<\frac{\pi}{2}.
\end{equation}
The proof is given in App.~\ref{app:quotoptproof}.
\end{proposition}

The update in \ref{eq:radialupdate} follows a great circle on the scale quotient exactly, including at finite step sizes. The same orthogonality separates the complementary raw radial evolution:
\begin{equation}
\label{eq:radialnormlaw}
\|w^+\|^2=a^2r^2+\eta^2p(r)^2\|g(w)\|^2.
\end{equation}

\begin{theorem}[Equivalent gauge-covariance criteria for radial updates]
\label{thm:quotopt}
Under the assumptions and notation of Prop.~\ref{prop:quotgeometry}, the following three conditions are equivalent: For every scale-invariant $J$, every nonstationary $w$, and every $c>0$, the updates from $w$ and $cw$ have the same canonical direction; the raw update is equivariant, $(cw)^+=cw^+$, for every such $J,w,c$; and the preconditioner has quadratic covariance, $p(cr)=c^2p(r)$ for all $c,r>0$. The proof is given in App.~\ref{app:quotoptproof}.
\end{theorem}

\begin{corollary}[Uniqueness of squared-scale radial preconditioning]
\label{cor:quotoptunique}
Under the same assumptions, the equivalent conditions of Thm.~\ref{thm:quotopt} hold if and only if there is a constant $\kappa>0$ such that
\begin{equation}
\label{eq:uniquepscale}
p(r)=\kappa r^2.
\end{equation}
For this unique family, $\tan\vartheta=(\eta\kappa/a)r\|g(w)\|$ is dimensionless and gauge-invariant. The proof is given in App.~\ref{app:quotoptproof}.
\end{corollary}

The radial qualifier is essential. For a general direction-dependent scalar preconditioner $\widetilde p:V\setminus\{0\}\to(0,\infty)$, gauge equivariance requires and is implied by $\widetilde p(cw)=c^2\widetilde p(w)$, whose complete family of solutions is $\widetilde p(w)=\|w\|^2\varphi(w/\|w\|)$ for an arbitrary positive function $\varphi$ on the unit sphere, as derived in App.~\ref{app:quotoptproof}. Cor.~\ref{cor:quotoptunique} establishes uniqueness only within isotropic radial rules, where $\varphi$ must be constant.

Ordinary Euclidean gradient descent has $p\equiv1$ and obeys $\tan\vartheta(cw)=c^{-2}\tan\vartheta(w)$. The Euclidean gradient step on the quotient depends quadratically on an arbitrary parameter rescaling. This quadratic dependence is the exact form of the effective learning rate that a normalizer induces on the normalizer's upstream block, where the angular progress of plain gradient descent scales as $\eta/r^2$ and the parameter norm silently retunes the step \citep{hoffer2018norm,arora2019autorate}. Cor.~\ref{cor:quotoptunique} identifies $p(r)=\kappa r^2$ as the unique isotropic radial rule that removes this dependence instead of tracking or annealing the resulting effective learning rate.

Together, Prop.~\ref{prop:quotgeometry}, Thm.~\ref{thm:quotopt}, and Cor.~\ref{cor:quotoptunique} make the optimizer rule operational at block level. Writing $d_V\coloneqq\dim V$, we have $r\|g(w)\|=d_V\RMS(w)\RMS(g(w))=d_V\Pi_w$, making this invariant a fixed-dimension multiple of the primal--dual RMS product in \ref{eq:G}. The normalized update in \ref{eq:quotgeodesic} has exactly the rotation form used by \gct{}.

\subsection{UVI calibration and depth-uniform stability}
\label{app:uvicalibration}

\paragraph{Additional UVI definitions.} In addition to the residual and attention indices in \secref{sec:uvi}, we define, with a small dimensionless floor $\epsilon$,
\begin{align}
\UVI_{\mathrm{act}} &\coloneqq \Var_\ell[\log(\hat s_{a,\ell} + \epsilon)],
\label{eq:uviact}\\
\UVI_{\mathrm{grad}} &\coloneqq \Var_\ell[\log(\Pi_\ell + \epsilon)],
\label{eq:uvigrad}\\
\UVI_{\mathrm{opt}} &\coloneqq \tfrac1B \textstyle\sum_b [\log \tfrac{\rho_b + \epsilon}{\rho^{\star} + \epsilon}]^2,
\label{eq:uviopt}
\end{align}
where $\hat s_{a,\ell}\coloneqq\RMS(a_\ell)/s^{\mathrm{ref}}_{a,\ell}$ is the gate and activation preactivation RMS expressed in a declared reference scale with $\uof{s^{\mathrm{ref}}_{a,\ell}}=\uof{a_\ell}$. The quantity $\Pi_\ell$ is the primal--dual RMS product from \ref{eq:G} at the block output, and $\rho_b$ is the relative update of parameter block $b$. The target $\rho^\star$ is the median of the $\rho_b$ values in log space. \ref{eq:uviopt} measures how far the blocks' relative updates deviate from this common target.

The activation and gradient indices in \ref{eq:uviact} and \ref{eq:uvigrad} measure the log-scale dispersion of normalized preactivation RMS and primal--dual RMS products across depth, respectively. A common log-scale offset cancels in both index variances because adding the same offset to all log-values shifts the mean log-value by the same amount, leaving each deviation from the mean unchanged. To monitor residual-stream scale growth across depth, we additionally measure $\sum_\ell\alpha_\ell^2$ and the aligned drift $\sum_\ell\max(2\alpha_\ell c_\ell,0)$ from \ref{eq:drift}. All five indices and the two additional diagnostics can be computed inexpensively on small batches at intervals during training.

\paragraph{Attention entropy regime.}
\label{sec:entropy}

The calibrated attention regime uses dimensionless, row-centered logit variance $v_i=O(1)$. This paragraph gives the entropy expansion and the limitation of using logit variance to infer saturation. For row logits $z_{ij}$ with row-centered variance $v_i\coloneqq(1/n)\sum_j(z_{ij}-\bar z_i)^2$, a second-order expansion of softmax entropy gives
\begin{equation}
\label{eq:ent}
H(p_i) = \log n - \tfrac12 v_i + O(\max_j |z_{ij} - \bar z_i|^3),
\end{equation}
showing that attention is nearly uniform for $v_i\ll1$ in this local regime. The calibrated dimensionless regime is $v_i=O(1)$. Large variance lies outside the Taylor regime and often accompanies saturation. Inferring low entropy from variance additionally requires an assumption on the logit shape. Entropy and maximum probability measure actual saturation directly. The temperature in \ref{eq:tau} targets the calibrated variance regime, and $\UVI_{\mathrm{attn}}$ measures deviation from that regime.

\paragraph{Depth-uniform residual drift.}

$\UVI_{\mathrm{res}}$ connects the measured step budget to stream-scale drift (\secref{sec:dynamics}). $\UVI_{\mathrm{res}}$ also yields a depth-uniform drift guarantee in the near-orthogonal regime. Prop.~\ref{prop:orthodrift} shows that $|c_k|\le\kappa\alpha_k$ and bounded target-relative step violation control $\log(s_N/s_0)$ independently of depth. This depth-uniform drift guarantee motivates combining $P^\perp$ with a depth-scaled residual step in \gct{}.

Consider any nondegenerate residual trajectory obeying \ref{eq:drift} over $N$ residual additions, with $s_k>0$, steps $\alpha_k>0$, and alignments $c_k\in[-1,1]$. Suppose every addition satisfies $\UVI_{\mathrm{res}}(k)=|\log(\alpha_k/\alpha^\star)|\le B$ in the zero-floor limit $\epsilon\to0$. This bound implies $\alpha_k\le a\coloneqq\alpha^\star e^B$ with $\alpha^\star\coloneqq1/\sqrt N$. For a Transformer with attention and MLP sub-blocks, $N=2L$, giving the target $\alpha^\star=1/\sqrt{2L}$ in \ref{eq:uvires}.

The exact projector of \secref{sec:gct} enforces this near-orthogonal regime for the \gct{} pre-retraction proposal.

\begin{proposition}[Depth-uniform drift under near-orthogonal residual additions]
\label{prop:orthodrift}
Under the preceding setup, suppose additionally that $|c_k| \le \kappa\alpha_k$ with $0\le\kappa < 1/2$. Then the stream can only grow and
\begin{align*}
0 \le \log(s_N/s_0) \le \tfrac{1+2\kappa}{2} \sum_{k} \alpha_k^2 \le \tfrac{(1+2\kappa)e^{2B}}{2}.
\end{align*}
The proof is given in App.~\ref{app:driftproof}.
\end{proposition}

The bound is independent of $N$. For fixed $B$, the drift upper bound is attained to leading order as $N\to\infty$ by $\alpha_k\equiv a$ and $c_k=\kappa\alpha_k$. In this limit, $a=e^B/\sqrt N\to0$.

The near-orthogonal stream's log-scale growth is bounded by $(1+2\kappa)e^{2B}/2$ at every depth under these assumptions, motivating orthogonalization and depth-scaled residual steps in \gct{}.

\subsection{Gauge-canonical factorization}
\label{app:factorization}

\paragraph{What scale separation preserves.} For nonzero inputs with independent positive rescalings, canonical coordinates preserve continuous invariant expressivity (Thm.~\ref{thm:factor}; Cor.~\ref{cor:factorexpressivity}). Targets that depend on absolute magnitude still require the relevant scale information. Coupled unit declarations can also leave meaningful dimensionless scale ratios, which must be retained (Thm.~\ref{thm:coupledquot}). Apps.~\ref{app:factorization} through~\ref{app:perturbstability} give the factorization and perturbation-stability details.

\paragraph{Factorization and expressivity.} For nonzero inputs that can each be rescaled independently by a positive factor, Thm.~\ref{thm:factor} shows that every equivariant computation factors into a function of RMS-normalized inputs and explicit scale factors that supply the required output unit. Choosing RMS-one coordinates makes this factorization unique. For targets that are unchanged by these rescalings, Cor.~\ref{cor:factorexpressivity} shows that normalized coordinates retain all the information needed to represent any continuous invariant function. Targets that depend on absolute magnitude still require the relevant scale information.

On nonzero shared-unit blocks, every map respecting independent positive rescalings has a unique factorization once the RMS-one section is chosen. The following theorem establishes this normal form, and the accompanying corollary shows that gauge fixing preserves continuous invariant expressivity.

For $j=1,\ldots,m$, let $V_j$ be a nonzero finite-dimensional real inner-product space of dimension $d_j$, set $r_j(x_j)\coloneqq\|x_j\|_j/\sqrt{d_j}=\RMS(x_j)$, and define $V_j^\times\coloneqq V_j\setminus\{0\}$, $\mathbb S_j\coloneqq\{z\in V_j:r_j(z)=1\}$, and $\canon_0(x_j)\coloneqq x_j/r_j(x_j)$. Let $G\coloneqq(\mathbb R_+)^m$ act on $X\coloneqq\prod_jV_j^\times$ by $g\cdot x\coloneqq(g_1x_1,\ldots,g_mx_m)$, and let the action of $G$ on a finite-dimensional normed output space $W$ be the scalar character $\chi_q(g)\coloneqq\prod_jg_j^{q_j}$ for a fixed $q\in\mathbb R^m$.

\begin{theorem}[Gauge-canonical factorization]
\label{thm:factor}
Under the preceding setup, a map $F:X\to W$ is equivariant,
\begin{equation}
\label{eq:factorequiv}
F(g\cdot x)=\chi_q(g)F(x)\qquad(g\in G),
\end{equation}
if and only if there is a unique map $f:\prod_j\mathbb S_j\to W$ such that
\begin{equation}
\label{eq:factorform}
F(x_1,\ldots,x_m) =\left(\prod_{j=1}^{m}r_j(x_j)^{q_j}\right) f(\canon_0(x_1),\ldots,\canon_0(x_m)).
\end{equation}
The proof is given in App.~\ref{app:factorproof}.
\end{theorem}

For $q\ne0$, \ref{eq:factorform} shows exactly which explicit scale monomial must restore the output unit.

\begin{corollary}[Invariant expressivity is preserved by gauge fixing]
\label{cor:factorexpressivity}
Under the assumptions of Thm.~\ref{thm:factor}, for $q=0$, pullback by $\canon_0$ is an isometric bijection, in the uniform norm, between continuous functions on the compact product $\prod_j\mathbb S_j$ and continuous $G$-invariant functions on $X$. The proof is given in App.~\ref{app:factorproof}.
\end{corollary}

Any model class uniformly dense on the canonical coordinates is uniformly dense in the entire continuous invariant function class after composition with $\canon_0$, which means that gauge fixing preserves the full continuous invariant function class. The mechanism is the scale-group counterpart of frame averaging and canonicalization for spatial symmetry groups, where a symmetry is likewise discharged by mapping each input to a canonical representative before an unconstrained network is applied \citep{puny2022frame,kaba2023canonical}. The rescaling group here is abelian and acts by a positive scalar on each block. The RMS-one section is a single global slice with a unique canonical representative for each orbit. No frame averaging is needed.

For the single residual-stream action, the invariant case is $F(h)=f(\canon_0(h))$ and the degree-one equivariant case is $F(h)=\RMS(h)f(\canon_0(h))$. The dimensionless head and the explicit scale-state reconstruction are the $q=0$ and $q=1$ instances of the same representation theorem. The corollary covers targets that are unchanged by rescaling. Targets that depend on absolute magnitude require retaining the relevant scale information. \secref{sec:exp-vit} discusses magnitude information as one possible explanation for the additive-noise results.

\subsection{Coupled neural units and the complete quotient}
\label{app:coupledunits}

When unit declarations couple different tensors, relative magnitudes can remain meaningful because two tensors with the same unit have a dimensionless RMS ratio. Thm.~\ref{thm:coupledquot} identifies all the scale combinations that must be retained, namely the dimensionless scale monomials generated by $\ker A$, where $A$ collects the declared unit exponents. Keeping these combinations prevents normalization from discarding information that is unchanged by the declared rescalings.

Thm.~\ref{thm:factor} uses independent rescalings, one rescaling for each block. Neural-unit declarations can instead couple several blocks to the same base units. In that general case, the quotient can contain both canonical directions and dimensionless ratios of empirical scales. Use the following notation for the coupled action.

For $j=1,\ldots,m$, let $V_j$ be a nonzero finite-dimensional real inner-product space, let $V_j^\times\coloneqq V_j\setminus\{0\}$, and define $r_j(x_j)\coloneqq\RMS(x_j)$, $z_j(x_j)\coloneqq\canon_0(x_j)=x_j/r_j(x_j)$, and $\mathbb S_j\coloneqq\{z:r_j(z)=1\}$. Write $g\coloneqq e^t\in G=(\mathbb R_+)^K$ and let the coupled scalar-unit action on $X=\prod_jV_j^\times$ be
\begin{equation}
\label{eq:coupledaction}
(e^t\cdot x)_j\coloneqq e^{a_j^\top t}x_j, \qquad A\coloneqq\begin{bmatrix}a_1 & \cdots & a_m\end{bmatrix}\in\mathbb R^{K\times m}.
\end{equation}
Put $s\coloneqq\operatorname{rank}A$ and let the columns of $N\in\mathbb R^{m\times(m-s)}$ be any basis of $\ker A$. With $\log r(x)\coloneqq(\log r_1(x_1),\ldots,\log r_m(x_m))$, define
\begin{equation}
\label{eq:completequotient}
\mathcal Q(x)\coloneqq (z_1(x_1),\ldots,z_m(x_m),N^\top\log r(x)).
\end{equation}

The next theorem gives the complete quotient. The following corollary determines when nonzero equivariant outputs exist and factorizes those outputs when the output character is represented by the input characters.

\begin{theorem}[Complete quotient for coupled neural units]
\label{thm:coupledquot}
Under the preceding setup, $\mathcal Q$ is a complete invariant: For all $x,x'\in X$,
\begin{equation}
\label{eq:maximalinvariant}
\mathcal Q(x)=\mathcal Q(x') \quad\Longleftrightarrow\quad x'=g\cdot x \text{ for some } g\in G.
\end{equation}
The proof is given in App.~\ref{app:coupledproof}.
\end{theorem}

The $N^\top\log r(x)$ coordinates are logarithms of dimensionless scale monomials generated by $\ker A$. When $\operatorname{rank}A=m$, no scale coordinate remains and canonical directions form the complete quotient. Otherwise, the $m-s$ Buckingham--$\pi$ coordinates in \ref{eq:completequotient} must be retained. The independent action of Thm.~\ref{thm:factor} is the special case $A=I_m$.

\begin{corollary}[Factorization of equivariant outputs]
\label{cor:outputfactor}
Under the assumptions of Thm.~\ref{thm:coupledquot}, let a finite-dimensional output space $W$ carry the scalar character $w\mapsto e^{q^\top t}w$. If $q\notin\operatorname{im}A$, the only equivariant map $F:X\to W$ is $F\equiv0$. If $q\in\operatorname{im}A$, fix any $p\in\mathbb R^m$ satisfying $Ap=q$. Then all and only the equivariant maps $F:X\to W$ are
\begin{equation}
\label{eq:coupledfactor}
F(x)= \left(\prod_{j=1}^m r_j(x_j)^{p_j}\right) f(z_1(x_1),\ldots,z_m(x_m),N^\top\log r(x)),
\end{equation}
where, for the fixed choices of $p$ and $N$, the map $f:(\prod_j\mathbb S_j)\times\mathbb R^{m-s}\to W$ is unique. Moreover, $F$ is continuous if and only if $f$ is continuous. The proof is given in App.~\ref{app:coupledproof}.
\end{corollary}

Thm.~\ref{thm:coupledquot} and Cor.~\ref{cor:outputfactor} give a design rule against over-canonicalization. A common scale is gauge, but a ratio of two quantities carrying the same unit is dimensionless and may be predictive. The ratio must either remain in the canonical computation or be retained as an explicit log-ratio state. The coordinates $N^\top\log S$ in Thm.~\ref{thm:lossscalar} are exactly the scale-only part of this maximal invariant. The independent per-token residual gauge used by \gct{} is the $N$-empty case in which no dimensionless scale coordinate remains.

\subsection{Perturbation stability on the canonical quotient}
\label{app:perturbstability}

\paragraph{Perturbation stability.} RMS normalization cancels a positive multiplier because the vector and the corresponding RMS are multiplied by the same factor. For $x\ne0$ and $\epsilon_x=0$, this cancellation gives $\canon(cx)=\canon(x)$ for every $c>0$. A perturbation that also changes direction can affect the result. Thm.~\ref{thm:tangent} separates a perturbation into a radial component along the input and a tangential component perpendicular to the input. As long as the radial component does not reverse the input's direction, normalization removes pure radial perturbations exactly, and the remaining error is bounded by the tangential component measured relative to the radially rescaled input.

Factorization identifies the information available to an invariant map. The next result quantifies the effect of a perturbation after canonicalization, assuming that the perturbation's radial component preserves orientation, meaning $a>-1$ in the notation below. Pure radial perturbations are removed exactly. For a mixed perturbation, the bound uses the tangential ratio $\rho/(1+a)$, which measures directional error after accounting for radial rescaling.

Let $V$ be a real inner-product space of dimension $d\ge2$, $r(x)\coloneqq\|x\|/\sqrt d$, $\mathbb S\coloneqq\{z:r(z)=1\}$, and $\canon_0(x)\coloneqq x/r(x)$ on $V^\times\coloneqq V\setminus\{0\}$. Equip $\mathbb S$ with the RMS chord metric $d_{\mathbb S}(z,z')\coloneqq r(z-z')$. For $x\ne0$ and a perturbation $e$, define the relative radial coefficient and tangential component of $e$ by $a\coloneqq\langle x,e\rangle/\|x\|^2$, $e_\perp\coloneqq e-a x$, and $\rho\coloneqq\|e_\perp\|/\|x\|$.

\begin{theorem}[Radial rejection and sharp tangential displacement]
\label{thm:tangent}
Under the preceding setup, if $a>-1$, then
\begin{equation}
\label{eq:tangentexact}
d_{\mathbb S}(\canon_0(x+e),\canon_0(x)) = \left[2\left(1-\frac{1+a}{\sqrt{(1+a)^2+\rho^2}}\right)\right]^{1/2} \le \frac{\rho}{1+a}.
\end{equation}
The proof is given in App.~\ref{app:tangentproof}.
\end{theorem}

The unit constant in the upper bound is asymptotically sharp as $\rho/(1+a)\to0$. For an orientation-preserving pure radial error $e=ax$ with $a>-1$, we have $\rho=0$ and the error is rejected exactly.

\begin{corollary}[Functional stability under coordinatewise quantization]
\label{cor:quantfunctional}
Under the notation and assumptions of Thm.~\ref{thm:tangent}, let $f:\mathbb S\to W$ be $K$-Lipschitz into a normed space $W$ and set $F\coloneqq f\circ\canon_0$. If, in a fixed orthonormal coordinate system, a nonclipping coordinatewise quantizer produces $e\coloneqq Q(x)-x$ with $|e_i|\le\Delta/2$ and $\delta\coloneqq\Delta/(2r(x))<1$, then $Q(x)\ne0$ and
\begin{equation}
\label{eq:quantfunctional}
\|F(Q(x))-F(x)\|_W\le K\frac{\delta}{1-\delta}.
\end{equation}
The proof is given in App.~\ref{app:tangentproof}.
\end{corollary}

For a canonical stream $r(x)=1$, this bound is independent of the token's original scale. For a shared quantization step on an unmanaged stream, the bound worsens as $1/r(x)$ and exposes exactly the cross-token scale-spread penalty of Prop.~\ref{prop:quant}.

The perturbation theorem and the quantization corollary separate two robustness claims that are otherwise easy to conflate. Gauge invariance removes multiplicative scale distortion with no Lipschitz constant. Finite-precision rounding generally introduces both radial and tangential components, but only the tangential component can change the canonical computation. The sensitivity to rounding is determined by the dimensionless ratio $\RMS(e)/\RMS(x)$. The rescaling experiment of \secref{sec:exp-perturb} isolates the first claim, while residual-stream quantization in \secref{sec:exp-quant} exercises the second claim.

\subsection{Shared-scale quantization on the quotient}

We next bound the nonclipping rounding error under shared-scale quantization. In symmetric $b$-bit absmax quantization with a shared per-tensor scale, the largest absolute coordinate in the tensor sets the quantization step. Smaller-scale tokens have a larger relative rounding-error bound.

Two conventions fix the scope of the bound. In the following results, the index $t$ ranges only over vectors that share the same quantization step. If a deployment uses a separate tensor scale at each block site, the result applies sitewise with a separate $\kappa_{\mathrm{scale},\ell}$. The stated bound applies directly to dynamic absmax. For a static nonclipping range $A_{\mathrm{cal}}\ge A$, the same proof replaces $A$ by $A_{\mathrm{cal}}$ and multiplies the dynamic-absmax right-hand sides by the calibration-slack factor $\gamma_{\mathrm{cal}}\coloneqq A_{\mathrm{cal}}/A$. If $A_{\mathrm{cal}}<A$, clipping contributes an additional error not covered by the proposition.

Formally, let $b\ge2$, $q\coloneqq2^{b-1}-1$, and let nonzero vectors $h_1,\dots,h_T\in\mathbb R^d$ have $r_t\coloneqq\RMS(h_t)$. Set $A\coloneqq\max_{t,i}|h_{t,i}|$, $\Delta\coloneqq A/q$, and $\kappa_{\mathrm{amp}}\coloneqq\max_t\|h_t\|_\infty/r_t$. Let $Q_\Delta$ be coordinatewise symmetric absmax quantization to the grid $\Delta\{-q,\ldots,q\}$.

\begin{proposition}[Shared-scale absmax rounding bound]
\label{prop:quant}
Under the preceding setup, no coordinate clips and, for every token,
\begin{align*}
\frac{\|Q_\Delta(h_t)-h_t\|_2}{\|h_t\|_2} \le \frac{\kappa_{\mathrm{amp}}}{2q}\frac{\max_s r_s}{r_t} \le \kappa_{\mathrm{amp}}2^{-(b-1)}\frac{\max_s r_s}{r_t}.
\end{align*}
The proof is given in App.~\ref{app:quantproof}.
\end{proposition}

For fixed normalized directions, the worst tokenwise multiplier in Prop.~\ref{prop:quant} is the scale spread $\kappa_{\mathrm{scale}}\coloneqq\max_t r_t/\min_t r_t$. A zero-floor canonical stream has $r_t\equiv1$ under \ref{eq:gctblock}, while per-token absmax scaling removes the shared-token factor for an arbitrary stream. With a finite frozen floor, an input of RMS $r$ is mapped to RMS $(1+\epsilon/r^2)^{-1/2}$. Equal output scales are only approximate when the input scales differ. The associated rescaling defect is bounded in \ref{eq:canondefect}. Under the usual high-resolution power-of-two convention, the shared-scale factor costs $\log_2\kappa_{\mathrm{scale}}$ effective bits. The factor is sharp in general.

The preceding sharpness statement concerns representation error. To characterize functional error after radial rejection, we next bound the displacement on the canonical quotient. The matching lower bound shows that shared-scale rounding can produce an unavoidable tangential error even when within-token amplitudes are bounded.

Let $q\ge2$ be an integer, let $h_1,\ldots,h_T\in\mathbb R^d\setminus\{0\}$, and use the shared absmax quantizer of Prop.~\ref{prop:quant}, with $r_t\coloneqq\RMS(h_t)$, $r_{\min}\coloneqq\min_t r_t$, $A\coloneqq\max_t\|h_t\|_\infty$, and $\Delta\coloneqq A/q$. Put $\kappa_{\mathrm{amp}}\coloneqq\max_t\|h_t\|_\infty/r_t$, $\kappa_{\mathrm{scale}}\coloneqq\max_t r_t/r_{\min}$, $\delta_*\coloneqq A/(2qr_{\min})$, and $\mathbb S_d\coloneqq\{z:\RMS(z)=1\}$.

\begin{lemma}[Shared-scale quotient upper bound]
\label{lem:quantquotupper}
Under the preceding setup, if $\delta_*<1$, then every $Q_\Delta(h_t)$ is nonzero and, for every $K$-Lipschitz $f:\mathbb S_d\to W$ into a normed space $W$ under the RMS chord metric,
\begin{equation}
\label{eq:quantquotupper}
\max_t\| f(\canon_0(Q_\Delta(h_t))) -f(\canon_0(h_t)) \|_W \le K\frac{\delta_*}{1-\delta_*}, \qquad \delta_*\le \frac{\kappa_{\mathrm{amp}}\kappa_{\mathrm{scale}}}{2q}.
\end{equation}
The proof is given in App.~\ref{app:quantminimaxproof}.
\end{lemma}

For $q\ge4$ and $2\le\Gamma\le q/2$, define the worst-case quotient error
\begin{align*}
\mathfrak E_q(\Gamma)\coloneqq \sup_{\substack{d\ge2,T\ge2\\ h_1,\ldots,h_T\ne0\\ \kappa_{\mathrm{scale}}\le\Gamma, \kappa_{\mathrm{amp}}\le\sqrt2\\ f:\mathbb S_d\to\mathbb R,\operatorname{Lip}(f)\le1}} \max_t |f(\canon_0(Q_\Delta(h_t))) -f(\canon_0(h_t))|,
\end{align*}
where, for each tensor in the supremum, $r_t$, $\kappa_{\mathrm{scale}}$, $\kappa_{\mathrm{amp}}$, $A$, $\Delta\coloneqq A/q$, and $Q_\Delta$ are defined as in Lem.~\ref{lem:quantquotupper}. These constraints imply $\delta_*\le\sqrt2\Gamma/(2q)\le\sqrt2/4<1$. Lem.~\ref{lem:quantquotupper} ensures that every canonicalized quantized vector in this definition is nonzero and the supremum is well-defined.

\begin{theorem}[Worst-case effective-bit law on the scale quotient]
\label{thm:quantminimax}
The dependence in Lem.~\ref{lem:quantquotupper} is worst-case sharp up to universal constants even for scalar outputs and $\kappa_{\mathrm{amp}}\le\sqrt2$: For every $q\ge4$ and $2\le\Gamma\le q/2$,
\begin{equation}
\label{eq:quantminimax}
\frac{\Gamma}{6\pi q} \le \mathfrak E_q(\Gamma) \le \frac{\sqrt2\Gamma}{2q-\sqrt2\Gamma}.
\end{equation}
The proof is given in App.~\ref{app:quantminimaxproof}.
\end{theorem}

Thus $\mathfrak E_q(\Gamma)=\Theta(\Gamma/q)$ in the stated high-resolution regime even when $\kappa_{\mathrm{amp}}$ is bounded. The lower-bound construction has $\kappa_{\mathrm{scale}}\in[\Gamma/2,\Gamma]$, so the same order is equivalently linear in the attained worst-case scale spread up to universal constants. Since $q=2^{b-1}-1$, the high-resolution interpretation of the worst-case shared-scale law is a loss of $\log_2\kappa_{\mathrm{scale}}$ effective bits even after exact radial rejection. At finite $b$, the exact statement is the factor in $q$ of \ref{eq:quantminimax}. A zero-floor canonical stream sets $\kappa_{\mathrm{scale}}=1$ and leaves only the within-vector term $O(\kappa_{\mathrm{amp}}2^{-b})$.

The deployment protocol assigns a separate scale to every block site. The theoretical bound applies across tokens within each site. The zero-floor \gct{} formulation enforces $r_{\ell,t}=1$ for every token and hence $\kappa_{\mathrm{scale},\ell}=1$. A finite-floor canonicalization instead maps an input RMS $r_{\ell,t,\mathrm{in}}$ to $(1+\epsilon/r_{\ell,t,\mathrm{in}}^2)^{-1/2}$ and has the intervention defect of \ref{eq:canondefect}. An unmanaged stream imposes no such constraint. The static-quantization columns directly measure the resulting deployment error, including calibration slack and any clipping, and the per-token control removes the shared-token-scale factor empirically.

Prop.~\ref{prop:quant} bounds how much rounding error a shared scale creates. Thm.~\ref{thm:tangent} shows that the downstream canonical graph discards the radial part of the rounding error. Thm.~\ref{thm:quantminimax} establishes that the remaining tangential functional error still has the same scale-spread order in the worst case. Together, these results identify two contributors to the deployment gap. An unmanaged stream inflates $\RMS(e)/\RMS(h_t)$ for small-scale tokens, while a canonical stream fixes the denominator and removes the scale direction.

In the high-resolution interpretation, unmanaged scale consumes $\log_2 \kappa_{\mathrm{scale}}$ effective bits. This cost occupies a larger fraction of the budget as the bit width decreases. In the int4 stream columns of Table~\ref{tab:quant}, the worst baseline changes by \quantworsthfour{} nats with a per-tensor scale, compared with $+\quantgcthfour{}$ for \gct{}. Per-token scales reduce the worst change to $+\quantworsthfourtok{}$, while \gct{} changes by $+\quantgcthfourtok{}$ and nGPT by $+\quantngpthfourtok{}$. This remaining gap is consistent with within-vector amplitude, downstream directional sensitivity, and accumulation across sites, none of which canonicalization removes.

\section{Derivations}
\label{app:deriv}

\subsection{Finite floors of the canonical gauge}
\label{app:floordefect}

Numerical stabilizers are also typed. A denominator $s_x+\epsilon_s$ requires $\uof{\epsilon_s}=\uof{x}$, and $\sqrt{s_x^2+\epsilon_{\mathrm{var}}}$ requires $\uof{\epsilon_{\mathrm{var}}}=\uof{x}^2$. Adding a fixed dimensionless $\epsilon$ to a unitful quantity breaks exact equivariance in the floor-dominated regime and serves only as numerical regularization.

For the canonicalizer in \secref{sec:gct}, setting $\epsilon_x=0$ recovers exact division by RMS for $x\ne0$. In this case, positive multiples such as $x$ and $2x$ give the same RMS-one vector, and applying $\canon$ again leaves that vector unchanged. The implementation adds a small positive $\epsilon_x$ to prevent division by zero. This numerical floor makes the output RMS less than one. Exact unit consistency with a finite floor requires $\uof{\epsilon_x}=\uof{x}^2$ and $\epsilon_x\mapsto c^2\epsilon_x$ when $x\mapsto cx$. Keeping $\epsilon_x$ numerically fixed during our rescaling interventions can change the normalized vector. The following bound quantifies the rescaling error in \ref{eq:canondefect}.

For $\canon_\epsilon(x)\coloneqq x/\sqrt{\RMS(x)^2+\epsilon}$ with a fixed numerical $\epsilon\ge0$, $r\coloneqq\RMS(x)>0$, and $c>0$, direct calculation and the mean-value theorem give
\begin{equation}
\label{eq:canondefect}
\RMS(\canon_\epsilon(cx)-\canon_\epsilon(x)) =\left|\frac{1}{\sqrt{1+\epsilon/(c^2r^2)}} -\frac{1}{\sqrt{1+\epsilon/r^2}}\right| \le \frac{\epsilon}{2r^2}|c^{-2}-1|.
\end{equation}
A correctly typed floor transforms as $\epsilon\mapsto c^2\epsilon$ and makes this defect exactly zero, whereas a frozen implementation floor leaves the explicit $O(\epsilon/r^2)$ local defect that \ref{eq:pathgain} propagates to the output.

The same floor also prevents an exact RMS of one. If $r=\RMS(x)>0$, then $\RMS(\canon(x))=r/\sqrt{r^2+\epsilon_x}=(1+\epsilon_x/r^2)^{-1/2}$. The canonicalizer of \secref{sec:gct} returns RMS one only in the zero-floor limit.

\paragraph{Finite-floor implementation.} For numerical regularization, the implementation uses $\|\bar h\|^2+\epsilon_\perp$ in the projection denominator. The display below derives the resulting residual alignment. The exact orthogonality statement in \secref{sec:gct} and the rotation identity in App.~\ref{app:residualderiv} apply to $P^\perp_{\bar h}$ or to the zero-projection-floor limit. Comparing relative errors for finite floors additionally requires a nondegenerate tangential proposal.

This projection floor behaves analogously to the canonicalization floor. Replacing $\|\bar h\|^2$ by $\|\bar h\|^2+\epsilon_\perp$ generally leaves a residual component along $\bar h$:
\begin{align*}
\left\langle \bar h, a-\frac{\langle a,\bar h\rangle}{\|\bar h\|^2+\epsilon_\perp}\bar h \right\rangle =\frac{\epsilon_\perp}{\|\bar h\|^2+\epsilon_\perp} \langle a,\bar h\rangle.
\end{align*}

\subsection{Dot-product variance and the calibrated temperature in \ref{eq:tau}}
\label{app:attentionderiv}

Suppose each $q_a$ is independent of $k_a$, both coordinates are zero-mean with $\Var(q_a)=s_Q^2$ and $\Var(k_a)=s_K^2$, and the products $q_ak_a$ are uncorrelated across channels. Channelwise independence is one sufficient condition. Then $\Var(q^{\top}k)=\sum_a\Var(q_ak_a)=d_hs_Q^2s_K^2$ exactly. With merely weak cross-channel correlations, the same equality is an approximation with the omitted cross-covariances added explicitly. Dividing logits by $\tau=\sqrt{d_h}s_Qs_K$ yields dimensionless logits of unit variance under the exact assumptions and $O(1)$ variance in the weak-correlation regime. In the zero-floor \gct{} formulation, per-head-vector canonicalization enforces $\|\bar q\|=\|\bar k\|=\sqrt{d_h}$. For weakly dependent directions, $\bar q^{\top}\bar k/\sqrt{d_h}$ again has $O(1)$ variance, and the learnable per-head gain tunes the regime multiplicatively.

\paragraph{Attention in canonical coordinates.} Queries and keys are canonicalized per head vector, $\bar q\coloneqq\canon(q)$ and $\bar k\coloneqq\canon(k)$. When $\epsilon_x=0$, RMS normalization removes the magnitude of each nonzero query and key before the dot product. The remaining $1/\sqrt{d_h}$ factor accounts for the head dimension. If the query and key directions are approximately isotropic and weakly dependent at initialization, we have $\Var(\bar q^{\top}\bar k/\sqrt{d_h})\approx1$, giving softmax dimensionless logits on an $O(1)$ scale. A per-head dimensionless gain $g_h$, initialized at $1$, tunes the attention entropy within these canonical coordinates.

\paragraph{Empirical calibration.} If the coordinates are centered with $\Var(q_a)=s_Q^2$ and $\Var(k_a)=s_K^2$, each $q_a$ is independent of $k_a$, and the products $q_ak_a$ are uncorrelated across channels, then $\Var(q^{\top}k)=d_hs_Q^2s_K^2$ exactly. Empirical per-head centered-RMS estimates give a calibrated approximation. Centered RMS is the RMS after subtracting the mean, $\RMS(x-\bar x\mathbf{1})$, with $\bar x$ computed over the same entries. Under that averaging convention, centered RMS equals the standard deviation. Ordinary RMS serves as a proxy when coordinate means are negligible.

\subsection{Gradient dual units}
For dimensionless $\Lcal$ and $\uof{x}=U$, consider $x\mapsto\chi_U(g)x$. Invariance of $\dl\Lcal=\langle\nabla_x\Lcal,\dl x\rangle$ forces $\nabla_x\Lcal\mapsto\chi_U(g)^{-1}\nabla_x\Lcal$. The products $x\odot\nabla_x\Lcal$ and $\RMS(x)\RMS(\nabla_x\Lcal)$ are dimensionless invariants, whereas the raw gradient norm scales inversely with $x$. The same argument for parameters gives $\uof{\nabla_\theta\Lcal}=U_\theta^{-1}$ and types Adam moments as $\uof{m}=U_\theta^{-1}$ and $\uof{v}=U_\theta^{-2}$. The ratio $m/(\sqrt v+\epsilon_g)$ is dimensionless when $\uof{\epsilon_g}=U_\theta^{-1}$.

\subsection{Residual scale dynamics for \ref{eq:drift}}
\label{app:residualderiv}

Let $s_\ell\coloneqq\|h_\ell\|_2/\sqrt d$ and assume $h_\ell\ne0$. If $r_\ell=0$, then $h_{\ell+1}=h_\ell$ and $s_{\ell+1}=s_\ell$, with no change in stream scale. For $r_\ell\ne0$, write $r_\ell=\alpha_\ell s_\ell z_\ell$ with $\RMS(z_\ell)=1$. The coefficient $\alpha_\ell=\RMS(r_\ell)/\RMS(h_\ell)>0$ is the dimensionless residual step size. Expanding $\|h_\ell+r_\ell\|_2^2$ shows how the branch's squared magnitude and alignment with the stream change the squared stream scale in \ref{eq:drift}.

With $h_{\ell+1} = h_\ell + r_\ell$, $\|h_{\ell+1}\|^2 = \|h_\ell\|^2 + \|r_\ell\|^2 + 2\langle h_\ell, r_\ell\rangle$. Dividing by $d$ and using $s_\ell=\|h_\ell\|/\sqrt d$, the definition $\|r_\ell\|=\alpha_\ell\|h_\ell\|$, and the cosine similarity $c_\ell$ gives \ref{eq:drift}.

The cosine similarity $c_\ell$ measures the directional alignment between the stream and the branch contribution. A positive value means that the contribution points partly along the stream, zero means that the two vectors are perpendicular, and a negative value means that the contribution points partly against the stream. If branch contributions align persistently with the stream, say $c_\ell\ge c_0>0$, bounded stream scale forces $\sum_\ell\alpha_\ell=O(1)$. If all $L$ steps are equal or remain uniformly comparable, each must be $O(L^{-1})$. This per-layer bound need not hold when only a few layers have appreciable contributions. Those layers may take larger steps, provided that $\sum_\ell\alpha_\ell$ stays bounded. If $c_\ell=0$ exactly, boundedness follows from $\sum_\ell\alpha_\ell^2=O(1)$ and permits the uniform schedule $\alpha_\ell=\Theta(L^{-1/2})$, which is the branch scale that depth-wise $\mu$P prescribes for depth transfer \citep{yang2024tp6}. Prop.~\ref{prop:orthodrift} extends this order to near-orthogonal contributions with $|c_\ell|\le\kappa\alpha_\ell$ and $0\le\kappa<1/2$. The alignment term $2\alpha_\ell c_\ell$ then has magnitude at most $2\kappa\alpha_\ell^2$. Exact orthogonality is unnecessary as long as alignment shrinks with the residual step size.

For persistent positive alignment $c_\ell\ge c_0>0$, the term $2\alpha_\ell c_\ell$ causes scale growth that is first order in the residual step size. Orthogonalizing the branch contribution removes this linear term and leaves $\alpha_\ell^2$, permitting larger uniform residual steps of order $L^{-1/2}$ under the same bounded scale budget. Geometrically, an orthogonal contribution changes the hidden state's direction, while the resulting change in the state's magnitude begins only at second order in $\alpha_\ell$. \secref{sec:gct} implements this orthogonalization by projecting each branch output perpendicular to the stream.

If $c_\ell\ge c>0$ for all $\ell$, then $s_L^2\ge s_0^2\prod_\ell(1+2c\alpha_\ell)$. Uniform boundedness of this product also bounds every $\alpha_\ell$. On that bounded interval, $\log(1+2c\alpha)\ge c'\alpha$ for some $c'>0$. Boundedness therefore forces $\sum_\ell\alpha_\ell=O(1)$. Only a uniform or uniformly comparable schedule implies the per-layer order $O(L^{-1})$. If $c_\ell=0$, then $s_L^2=s_0^2\prod_\ell(1+\alpha_\ell^2)\le s_0^2e^{\sum_\ell\alpha_\ell^2}$, so $\sum_\ell\alpha_\ell^2=O(1)$ suffices. With uniform steps, the bounded quadratic step budget permits $\alpha_\ell=\Theta(L^{-1/2})$ instead of $\Theta(L^{-1})$. With the exact projector, a nonzero projected branch output, and zero canonicalization floors, \gct{} adds a contribution orthogonal to the current state. The subsequent canonicalization normalizes the sum of the state and this contribution, giving $\bar h_{\ell^+} = (\bar h_\ell + \alpha_\ell \hat a_\ell)/\sqrt{1+\alpha_\ell^2}$. The finite projection-floor qualification is stated in App.~\ref{app:floordefect}.

\paragraph{Canonical residual update.} In \ref{eq:gctblock}, $\Phi_\ell$ is the attention or MLP branch, and $\ell^+$ denotes the state after that sub-block. When the projected branch output $P^\perp_{\bar h}\bar a$ is nonzero and the canonicalization floors are set to zero, $\hat a\perp\bar h$ and both vectors have RMS one. The final normalization, which we call the retraction, returns the updated vector to RMS one. The next canonical state is the exact rotation
\begin{align*}
\bar h_{\ell^+} =\frac{\bar h_\ell+\alpha_\ell\hat a_\ell}{\sqrt{1+\alpha_\ell^2}},
\end{align*}
which changes direction while keeping RMS fixed. Before retraction, the vector actually added to the state is $r_\ell^{\mathrm{pre}}\coloneqq\alpha_\ell\hat a_\ell$. This contribution is orthogonal to $\bar h_\ell$, matching the orthogonal residual case analyzed in \secref{sec:dynamics}. When $\alpha_\ell\ne0$, the pre-retraction contribution has cosine similarity $c_\ell^{\mathrm{pre}}=0$ with the stream and an RMS ratio to the stream of exactly $|\alpha_\ell|$. The learned scalar directly controls the relative size of the added contribution. When $\alpha_\ell=0$, no branch contribution is added, and exact RMS normalization leaves the state unchanged. The final displacement $\bar h_{\ell^+}-\bar h_\ell$ generally includes a component along $\bar h_\ell$ because retraction rescales the sum. With zero canonicalization floors, the resulting state still satisfies $\RMS(\bar h_{\ell^+})=1$ exactly.

\subsection{Attention entropy expansion for \ref{eq:ent}}
Let $z_j=\bar z+\delta_j$ with $(1/n)\sum_j\delta_j=0$, put $v\coloneqq(1/n)\sum_j\delta_j^2$, and write $M\coloneqq\max_j|\delta_j|$. Then $p_j = e^{\delta_j} / \sum_k e^{\delta_k}$ and a Taylor expansion of $H(p) = -\sum_j p_j \log p_j$ in $\delta$ gives $H=\log n-v/2+O(M^3)$. To obtain this expansion, write $\sum_ke^{\delta_k}=n(1+v/2+O(M^3))$ and $\log p_j=\delta_j-\log n-v/2+O(M^3)$. Substituting these expansions gives
\begin{align*}
H=\log n+\frac v2-\frac1n\sum_j\left(1+\delta_j+\frac{\delta_j^2}{2}\right)\delta_j+O(M^3) =\log n-\frac v2+O(M^3).
\end{align*}
Centered logit variance controls the local attention entropy to leading order. Small $v$ gives near-uniform attention. Large $v$ flags over-dispersed logits. Outside the Taylor regime, entropy and maximum probability identify whether attention is saturated. $\UVI_{\mathrm{attn}}$ measures under- and over-dispersion symmetrically in log space.

\subsection{Exact gauge invariance of \gct{}}
\label{app:exactgauge}

\paragraph{Boundary invariance.} Assume that the canonicalization floors are zero and every vector being normalized is nonzero. Replacing a block input $h$ by $ch$ for $c>0$ then leaves the network's dimensionless output unchanged because $\canon(ch)=\canon(h)$. This invariance holds at every block-input boundary, including $h_0$, and at the final block-output boundary because the stated head also canonicalizes the final block output.

\paragraph{Derivation.}

In the zero-floor limit, positive homogeneity of RMS gives $\canon(ch)=\canon(h)$ for every $c>0$. A \gct{} block including entry canonicalization has the same output on $h$ and $ch$ wherever the block's canonicalizations are defined. Applying this identity at the first downstream block shows that a network with the stated dimensionless head is invariant to positive rescaling at every block-input boundary, including $h_0$. The head's entry canonicalization also removes a rescaling at the final block-output boundary.

For an output with units, the output value must change with the declared scale. Under $(s_\ell,\bar h_\ell)\mapsto(c s_\ell,\bar h_\ell)$, the reconstructed output changes from $s_\ell\bar h_\ell$ to $c s_\ell\bar h_\ell$, while the normalized coordinate and the canonical-coordinate computation stay unchanged. This behavior is called equivariance.

For unitful outputs, the corresponding statement is equivariance rather than invariance. Under the scale-state action $(s_\ell,\bar h_\ell)\mapsto(c s_\ell,\bar h_\ell)$, the reconstruction map $R(s,\bar h)\coloneqq s\bar h$ obeys $R(cs,\bar h)=cR(s,\bar h)$, while the canonical coordinate is fixed. The reconstruction $s_\ell\bar h_\ell$ follows the declared scale change exactly.

\subsection{Further examples of local typing rules}
\label{app:typingexamples}

\paragraph{Independent scale factors.} Each element $g=(g_1,\ldots,g_K)\in G$ assigns a multiplier $g_k$ to each chosen scale. Each multiplier can be chosen independently as any positive real number. Examples include the residual-stream scale of each stage, the target scale of each task head, and per-block parameter scales.

\paragraph{A single residual-stream unit.} The simplest rescaling group has one independent scale factor, $K=1$. Let the single factor $g$ rescale the residual stream of a Pre-LN block, and write $U_{\mathrm{res}}$ for the resulting stream unit. Under $h_\ell\mapsto gh_\ell$, the identity $N(gh_\ell)=N(h_\ell)$ shows that the normalizer output stays unchanged and is dimensionless. To produce an output that can be added to $h_\ell$, the branch $F_\ell$ must restore the stream unit through a map $1\to U_{\mathrm{res}}$. This declaration concerns the chosen activation rescaling and applies even when the input data, such as token identifiers, have no physical units. \secref{sec:rules} generalizes the example into local typing rules.

\paragraph{Unit types and empirical scales.} We distinguish a tensor's unit type from the tensor's empirical scale. The type $\uof{x}$ specifies how $x$ transforms under $G$ and can be read from the graph. The empirical scale $s_x$ is a measured magnitude with the same type as the tensor, $\uof{s_x}=\uof{x}$, making $x/s_x$ dimensionless. For example, multiplying $h_\ell$ by $g$ also multiplies the tensor's RMS value by $g$, giving $\RMS(gh_\ell)=g\RMS(h_\ell)$. The tensor and the corresponding RMS share the unit $U_{\mathrm{res}}$, while a forward pass supplies the actual RMS value. The attention scales $s_Q,s_K$, stream scale $s_\ell$, and characteristic loss scales $S_i$ used below are empirical scales in this sense. A single RMS scale applies to a block whose coordinates share one scalar unit. Mixed-unit tensors require separate scales for shared-unit blocks, or a unit-compatible metric.

\paragraph{Parameters and explicit converters.} Parameters follow the same rule. In $y\coloneqq f_\phi(x)$, we treat $\phi$ as another input and declare the parameter transformation $\phi\mapsto\rho_\phi(g)\phi$. For example, in a linear map $y=Wx$ between scalar-unit blocks, the weights must convert the input unit $U_{\mathrm{in}}$ into the output unit $U_{\mathrm{out}}$, which requires $\uof{W}=U_{\mathrm{out}}U_{\mathrm{in}}^{-1}$. Fixing these units before checking the graph makes mismatches detectable because an addition with unequal input units requires a converter. We keep unit conversion explicit by separating dimensionless mixing weights $\bar W$ from scale converters $s$. The factorization $W=s\bar W$ has $\uof{\bar W}=1$ and $\uof{s}=U_{\mathrm{out}}U_{\mathrm{in}}^{-1}$, as developed in \secref{sec:gct}.

\paragraph{Normalization and output units.} Consider $N(x)\coloneqq x/\RMS(x)$ on a nonzero shared-unit block. Here a gauge is a convention for representing tensors that differ only by the declared rescaling. Gauge fixing chooses one representative from each such family. Here we divide by RMS, which maps every $cx$, $c>0$, to the same RMS-one tensor. We call this mapping to a fixed representative canonicalization. RMSNorm applies this division, yielding dimensionless coordinates before the learnable RMSNorm gain. LayerNorm also removes a common coordinate shift by subtracting the mean. Learnable gains specify the output unit. For an affine normalization $\gamma\odot N(x)+\beta$, the output unit is the unit of $\gamma$, provided that the bias $\beta$ is assigned the same unit.

\paragraph{Activations and transition scales.} The rectified linear unit (ReLU) is positively homogeneous and satisfies $\mathrm{ReLU}(cx)=c\mathrm{ReLU}(x)$ for $c\ge0$. Under the positive rescalings in $G$, the ReLU output scales by the same factor as the ReLU input, preserving the scalar unit.

Standard non-homogeneous activations, including sigmoid, tanh, GELU, and SiLU, require dimensionless arguments. The formula for each function contains a fixed transition scale. Changing a unitful input's units can move the same underlying value to a different part of the nonlinear response curve. Evaluating the nonlinear response at $x/s$ avoids this change because $x$ and $s$ rescale together. Unit-consistent forms are $\mathrm{GELU}_s(x)\coloneqq x\Phi(x/s)$ and $\mathrm{SiLU}_s(x)\coloneqq x\sigma(x/s)$ with $\uof{s}=\uof{x}$.

\section{Proofs}
\label{app:proofs}

\subsection{Proof of Proposition~\ref{prop:dualsound}}
\label{app:dualsound}
\begin{proof}[Proof of Proposition~\ref{prop:dualsound}]
Evaluating the graph on source values transformed by the declared source actions places every node at the corresponding transformed state, by topological substitution through the unit-consistent primitives. Fix $g\in G$ and abbreviate $R_i\coloneqq\rho_i(g)$ and $R_y\coloneqq\rho_y(g)$. Here $\top$ denotes the adjoint with respect to the fixed inner products on the node spaces, or the usual matrix transpose in orthonormal coordinates. At a differentiability point, differentiate the local equivariance identity in \ref{eq:unitconsistent} with respect to the primitive's $i$th argument. With $x'\coloneqq(R_1x_1,\ldots,R_mx_m)$, the chain rule gives
\begin{align}
D_i f(x')R_i=R_yD_i f(x), \qquad\text{hence}\qquad D_i f(x')=R_yD_i f(x)R_i^{-1}.
\label{eq:dualjacobian}
\end{align}
Let $\bar y$ denote the reverse-mode cotangent at $y$. The cotangent at the dimensionless loss is the scalar $1$, which is fixed by the dual action. Suppose, in reverse topological order, that $\bar y'=R_y^{-\top}\bar y$. The contribution of the edge $x_i\to y$ to the transformed parent cotangent follows from \ref{eq:dualjacobian}:
\begin{align*}
D_i f(x')^\top\bar y' =R_i^{-\top}D_i f(x)^\top R_y^\top R_y^{-\top}\bar y =R_i^{-\top}D_i f(x)^\top\bar y.
\end{align*}
If a node has several children, the node's reverse-mode cotangent is the sum of the child contributions, and every summand has the same factor $R_i^{-\top}$. Reverse induction yields $\bar x_v'=\rho_v(g)^{-\top}\bar x_v$ at every node, which is the claimed dual transformation law for $\nabla_{x_v}\Lcal$.
\end{proof}

\subsection{Proof of Theorem~\ref{thm:robustsound}}
\label{app:robustproof}
\begin{proof}[Proof of Theorem~\ref{thm:robustsound}]
For every node, define the equivariance error $E_v\coloneqq\|\widetilde x_v-\rho_v(g)x_v\|_v$. At a source $s\in\mathcal S$, the transformed evaluation is defined by $\widetilde x_s\coloneqq\rho_s(g)x_s$, hence $E_s=0$. For a nonsource node, insert and subtract the value of the node's primitive on the ideally transformed parent tuple. The triangle inequality, \ref{eq:nodelip}, and \ref{eq:localdefect} give
\begin{align}
E_v &=\|f_v((\widetilde x_u)_u)-\rho_v(g)x_v\|_v\nonumber\\
&\le \|f_v((\widetilde x_u)_u) -f_v((\rho_u(g)x_u)_u)\|_v +\delta_v(g,x)\nonumber\\
&\le \delta_v(g,x) +\sum_{u\in\mathrm{pa}(v)}K_{vu}E_u .
\label{eq:robustrecurrence}
\end{align}
All parent indices in this recurrence precede $v$ in a topological ordering.

We now unroll \ref{eq:robustrecurrence}. We claim, for every node $v$, that
\begin{equation}
E_v\le \sum_{w\notin\mathcal S}\delta_w(g,x) \sum_{p\in\mathcal P(w,v)} \prod_{(a\to b)\in p}K_{ba},
\label{eq:robustunroll}
\end{equation}
where a sum over an empty path set is zero and the empty path from $v$ back to $v$ has product one. The claim is immediate for a source. Suppose the claim holds for all parents of a nonsource $v$. Substitution into \ref{eq:robustrecurrence} appends the edge $u\to v$ to every path ending at each parent $u$. Every nonempty directed path ending at $v$ has a unique last edge and is obtained exactly once by this path extension. The term $\delta_v$ supplies the empty path. This construction proves \ref{eq:robustunroll} by topological induction. Setting $v=o$ yields \ref{eq:pathgain}.
\end{proof}

\subsection{Proofs of Theorem~\ref{thm:lossscalar} and Corollary~\ref{cor:lossunique}}
\label{app:lossscalarproof}
\begin{proof}[Proof of Theorem~\ref{thm:lossscalar}]
Expanding the invariance identity and comparing the coefficient of each $\ell_i$ gives
\begin{align*}
w_i(D(t)S)=e^{-a_i^\top t}w_i(S).
\end{align*}
We now solve this covariance equation completely. Define $\beta_i(S)\coloneqq S_iw_i(S)$. Because the $i$th coordinate of $D(t)S$ is $e^{a_i^\top t}S_i$, the covariance equation is equivalent to $\beta_i(D(t)S)=\beta_i(S)$ for all $S\in\mathbb R_+^m$ and $t\in\mathbb R^K$. Pass to logarithmic coordinates $y\coloneqq\log S\in\mathbb R^m$ and put $b_i(y)\coloneqq\beta_i(e^y)$, with the exponential also componentwise. Since $\log(D(t)S)=y+A^\top t$, this identity says precisely that $b_i$ is constant on every affine orbit $y+\operatorname{im}A^\top$.

We now identify those orbits without any regularity assumption on $w$. The columns of $N$ span $\ker A$, so $\ker N^\top=(\operatorname{im}N)^\perp=(\ker A)^\perp=\operatorname{im}A^\top$, where the last equality is the finite-dimensional fundamental theorem of linear algebra. Therefore
\begin{align*}
N^\top y=N^\top y' \quad\Longleftrightarrow\quad y'-y\in\operatorname{im}A^\top.
\end{align*}
Thus the fibers of $y\mapsto N^\top y$ are exactly the $G$-orbits. Since $N$ has full column rank, $N^\top:\mathbb R^m\to\mathbb R^{m-s}$ is surjective. We may define, for each $z\in\mathbb R^{m-s}$, $\psi_i(z)\coloneqq b_i(y)$ for any $y$ satisfying $N^\top y=z$. The orbit characterization makes this definition independent of the chosen $y$, and the definition yields
\begin{align*}
\beta_i(S)=\psi_i(N^\top\log S), \qquad w_i(S)=\frac{1}{S_i}\psi_i(N^\top\log S),
\end{align*}
which proves necessity of \ref{eq:losscomplete}.

Conversely, let the weights have the form in \ref{eq:losscomplete}. Since $AN=0$, we have $N^\top\log(D(t)S)=N^\top\log S+N^\top A^\top t=N^\top\log S$. The argument of every $\psi_i$ is invariant, while the prefactor transforms as $(D(t)S)_i^{-1}=e^{-a_i^\top t}S_i^{-1}$. Hence the coefficient covariance holds, and expanding the scalarization proves the invariance identity. This direction establishes sufficiency and completes the characterization.

Finally, for any $n\in\ker A$,
\begin{align*}
\prod_{j=1}^m(D(t)S)_j^{n_j} =\exp(t^\top An)\prod_{j=1}^mS_j^{n_j} =\prod_{j=1}^mS_j^{n_j}.
\end{align*}
Taking logarithms gives $n^\top\log S$. Conversely, every invariant linear combination of the log-scales has coefficient vector in $\ker A$, so the coordinates supplied by a basis $N$ exhaust the dimensionless monomial groups.
\end{proof}

\begin{proof}[Proof of Corollary~\ref{cor:lossunique}]
If $\operatorname{rank}A=m$, then $m-s=0$ and $\mathbb R^{m-s}=\mathbb R^0$ is a one-point space. Every $\psi_i$ is a constant $\alpha_i$, giving \ref{eq:lossunique}. The coefficient comparison at the start of the theorem proof shows there are no other invariant linear rules.
\end{proof}

\subsection{Proof of Proposition~\ref{prop:precondlaw}}
\label{app:precondproof}
\begin{proof}[Proof of Proposition~\ref{prop:precondlaw}]
For the preconditioned update, the displacement computed in the transformed run is $u^g=-\eta P^gR^{-\top}q$, whereas equivariance requires the displacement to equal $Ru=-\eta RPq$. Equality for every covector $q$ is equivalent to the operator identity $P^gR^{-\top}=RP$. Right multiplication by $R^\top$ gives $P^g=RPR^\top$, proving both necessity and sufficiency of \ref{eq:precondlaw}.

Ordinary SGD with a common learning rate has $P^g=P=I$, so \ref{eq:precondlaw} reduces to $I=RR^\top$. If scalar learning rates may differ, the corresponding calculation is $\eta^gI=\eta RR^\top$. For $R=cI$, the condition is exactly $\eta^g=c^2\eta$.
\end{proof}

\subsection{Proofs of Proposition~\ref{prop:quotgeometry}, Theorem~\ref{thm:quotopt}, and Corollary~\ref{cor:quotoptunique}}
\label{app:quotoptproof}
\begin{proof}[Proof of Proposition~\ref{prop:quotgeometry}]
Suppose $g(w)\ne0$. The unit vectors $\bar w=w/r$ and $\bar g=g(w)/\|g(w)\|$ are orthogonal by \ref{eq:scalegradlaw}, and the update can be written
\begin{align*}
w^+=ar\bar w-\eta p(r)\|g(w)\|\bar g.
\end{align*}
Pythagoras gives \ref{eq:radialnormlaw}. Define $\vartheta\in(0,\pi/2)$ by
\begin{align*}
\cos\vartheta=\frac{ar}{\|w^+\|}, \qquad \sin\vartheta=\frac{\eta p(r)\|g(w)\|}{\|w^+\|}.
\end{align*}
The ratio of the displayed sine and cosine is the tangent in \ref{eq:quotgeodesic}, and division of the preceding expression for $w^+$ by the norm of $w^+$ proves that equation. Because $\bar w$ and $-\bar g$ span a two-dimensional plane through the origin, the curve $\cos\theta\bar w-\sin\theta\bar g$ is the unit sphere's great circle through $\bar w$ in direction $-\bar g$. The step is exactly geodesic.
\end{proof}

\begin{proof}[Proof of Theorem~\ref{thm:quotopt}]
The gradient scaling law gives the update from $cw$ as
\begin{align*}
(cw)^+=acw-\eta p(cr)c^{-1}g(w).
\end{align*}
If $p(cr)=c^2p(r)$, the right side of this update equals $c[aw-\eta p(r)g(w)]=cw^+$. Thus quadratic covariance implies raw equivariance, which immediately implies canonical-direction invariance.

Conversely, at a nonstationary point the normalized gradient direction is unchanged by positive rescaling, and \ref{eq:quotgeodesic} applied at $cw$ gives
\begin{align*}
\tan\vartheta(cw) =\frac{\eta p(cr)\|c^{-1}g(w)\|}{a\|cw\|} =\frac{p(cr)}{c^2p(r)}\tan\vartheta(w).
\end{align*}
Both angles lie in $(0,\pi/2)$ and use the same ordered orthonormal pair $(\bar w,-\bar g)$, so equality of the canonical directions forces equality of the angles. Thus canonical-direction invariance implies $p(cr)=c^2p(r)$ wherever the test point is nonstationary. The invariance hypothesis applies at every radius. Because $\dim V\ge2$, for any $r>0$ choose orthonormal $u,v$, put $w=ru$, and use the smooth scale-invariant objective $J_v(x)\coloneqq\langle v,x\rangle/\|x\|$. At $w$, the objective's gradient is $v/r\ne0$. Hence the angle identity proves quadratic covariance for every $c,r>0$. This implication closes the cycle and proves the equivalence of the three conditions.
\end{proof}

\begin{proof}[Proof of Corollary~\ref{cor:quotoptunique}]
Quadratic covariance with $r=1$ and $c=r'$ gives $p(r')=r'^2p(1)$. Thus quadratic covariance implies the squared-scale form with $\kappa\coloneqq p(1)>0$, and the reverse implication is immediate. For $p(r)=\kappa r^2$, the scaled-update calculation proves raw equivariance directly and \ref{eq:quotgeodesic} reduces to $\tan\vartheta=(\eta\kappa/a)r\|g(w)\|$. The product is invariant because $\|cw\|\|g(cw)\|=r\|g(w)\|$. This computation proves the claimed unique family and the gauge invariance of the quotient angle.
\end{proof}

\paragraph{Direction-dependent extension and Euclidean case.} For $p\equiv1$, the scaled-angle identity in the theorem proof gives $\tan\vartheta(cw)=c^{-2}\tan\vartheta(w)$, proving the Euclidean comparison of App.~\ref{app:optimizergeometry}. Now replace the radial rule $p(r)$ by a general scalar rule $\widetilde p(w)>0$. The same scaled-update calculation shows that raw gauge equivariance is equivalent to $\widetilde p(cw)=c^2\widetilde p(w)$. Writing $w=ru$ with $r=\|w\|$ and $\|u\|=1$ then gives $\widetilde p(w)=r^2\widetilde p(u)=\|w\|^2\varphi(w/\|w\|)$, where $\varphi$ is an arbitrary positive function on the unit sphere. Every such rule satisfies the covariance law. This derivation proves the qualification following Cor.~\ref{cor:quotoptunique} and shows why radiality forces $\varphi$ to be constant.

\subsection{Proof of Proposition~\ref{prop:orthodrift}}
\label{app:driftproof}
\begin{proof}[Proof of Proposition~\ref{prop:orthodrift}]
The UVI hypothesis implies $\alpha_k\le\alpha^\star e^B=a$. Iterating the exact recursion in \ref{eq:drift} gives
\begin{align}
\left(\frac{s_N}{s_0}\right)^2 =\prod_{k=1}^N(1+\alpha_k^2+2\alpha_kc_k),
\label{eq:driftproduct}
\end{align}
and hence, for a nondegenerate trajectory,
\begin{align}
2\log\frac{s_N}{s_0} =\sum_{k=1}^N\log(1+\alpha_k^2+2\alpha_kc_k).
\label{eq:driftlog}
\end{align}

The near-orthogonality assumption gives the two-sided estimate
\begin{align}
1+(1-2\kappa)\alpha_k^2 \le1+\alpha_k^2+2\alpha_kc_k \le1+(1+2\kappa)\alpha_k^2.
\label{eq:drifttwosided}
\end{align}
Because $\kappa<1/2$, the left side is at least one, so every factor in \ref{eq:driftproduct} is at least one and $s_N\ge s_0$. For the upper bound, use \ref{eq:driftlog}, the right side of \ref{eq:drifttwosided}, and again $\log(1+x)\le x$:
\begin{align*}
2\log\frac{s_N}{s_0} &\le\sum_{k=1}^N\log(1+(1+2\kappa)\alpha_k^2)\\
&\le(1+2\kappa)\sum_{k=1}^N\alpha_k^2 \le(1+2\kappa)Na^2 =(1+2\kappa)e^{2B}.
\end{align*}
Division by two proves the claim.

Finally, choose $\alpha_k=a$ and $c_k=\kappa a$ for every $k$. This choice is admissible for all sufficiently large $N$ at fixed $B$. The right inequality in \ref{eq:drifttwosided} is then exact and
\begin{align*}
\log\frac{s_N}{s_0} =\frac N2\log(1+(1+2\kappa)a^2) =\frac{1+2\kappa}{2}e^{2B}+O(Na^4).
\end{align*}
Since $Na^4=e^{4B}/N\to0$, the chosen steps and alignments attain the stated upper constant to leading order.
\end{proof}

\subsection{Proofs of Theorem~\ref{thm:factor} and Corollary~\ref{cor:factorexpressivity}}
\label{app:factorproof}
\begin{proof}[Proof of Theorem~\ref{thm:factor}]
Assume first that $F$ satisfies \ref{eq:factorequiv}. On $\mathbb S\coloneqq\prod_j\mathbb S_j$, define $f$ by restriction, $f(z)\coloneqq F(z)$. Given an arbitrary $x\in X$, put $s_j\coloneqq r_j(x_j)>0$ and $z_j\coloneqq x_j/s_j=\canon_0(x_j)\in\mathbb S_j$. Then $x=s\cdot z$ for $s\coloneqq(s_1,\ldots,s_m)\in G$, so equivariance gives
\begin{align*}
F(x)=F(s\cdot z) =\chi_q(s)F(z) =\left(\prod_{j=1}^{m}r_j(x_j)^{q_j}\right) f(\canon_0(x_1),\ldots,\canon_0(x_m)),
\end{align*}
which is \ref{eq:factorform}.

Conversely, suppose \ref{eq:factorform} holds. Positive homogeneity of RMS implies, for $g_j>0$, $r_j(g_jx_j)=g_jr_j(x_j)$ and $\canon_0(g_jx_j)=\canon_0(x_j)$. Substituting these identities into \ref{eq:factorform} yields
\begin{align*}
F(g\cdot x) =\left(\prod_j(g_jr_j(x_j))^{q_j}\right) f(\canon_0(x_1),\ldots,\canon_0(x_m)) =\chi_q(g)F(x),
\end{align*}
so $F$ is equivariant. Uniqueness is forced by restriction to the canonical section. If $z\in\mathbb S$, then every $r_j(z_j)=1$ and \ref{eq:factorform} gives $f(z)=F(z)$.
\end{proof}

\begin{proof}[Proof of Corollary~\ref{cor:factorexpressivity}]
Now take $q=0$. The product $\mathbb S$ is compact because each $V_j$ is finite-dimensional, and $\canon_0:X\to\mathbb S$ is continuous and surjective. Define the pullback $P:C(\mathbb S,W)\to C(X,W)^G$ by $(Pf)(x)\coloneqq f(\canon_0(x_1),\ldots,\canon_0(x_m))$. The already-proved factorization says that restriction to $\mathbb S$ is the inverse of $P$, so $P$ is a bijection. Moreover, for any $f,h\in C(\mathbb S,W)$, surjectivity of $\canon_0$ gives
\begin{align*}
\|Pf-Ph\|_{\infty,X} &=\sup_{x\in X} \|f(\canon_0(x))-h(\canon_0(x))\|_W\\
&=\sup_{z\in\mathbb S}\|f(z)-h(z)\|_W =\|f-h\|_{\infty,\mathbb S}.
\end{align*}
Thus $P$ is an isometric bijection. If a class $\mathcal H$ is uniformly dense in $C(\mathbb S,W)$, approximate the unique restriction $f\coloneqq F|_{\mathbb S}$ of any continuous invariant $F$ by elements of $\mathcal H$. The preceding norm identity transfers the same approximation error unchanged to all of $X$. This transfer proves the expressivity assertion.
\end{proof}

\subsection{Proofs of Theorem~\ref{thm:coupledquot} and Corollary~\ref{cor:outputfactor}}
\label{app:coupledproof}
\begin{proof}[Proof of Theorem~\ref{thm:coupledquot}]
Write $y(x)\coloneqq\log r(x)\in\mathbb R^m$. Positive homogeneity of RMS and \ref{eq:coupledaction} give $y(e^t\cdot x)=y(x)+A^\top t$ and $z_j((e^t\cdot x)_j)=z_j(x_j)$. Since $AN=0$, multiplication of the first identity by $N^\top$ shows that $\mathcal Q(e^t\cdot x)=\mathcal Q(x)$. Thus $\mathcal Q$ is invariant.

To prove completeness, suppose $\mathcal Q(x)=\mathcal Q(x')$ and put $y\coloneqq y(x)$, $y'\coloneqq y(x')$. Equality of the last coordinates gives $N^\top(y'-y)=0$. The columns of $N$ span $\ker A$, hence the finite-dimensional fundamental theorem of linear algebra gives $\ker N^\top=(\ker A)^\perp=\operatorname{im}A^\top$. There is therefore a $t\in\mathbb R^K$ such that $y'-y=A^\top t$. Equality of the canonical directions and the identity $x_j=e^{y_j}z_j(x_j)$ now imply
\begin{align*}
x'_j=e^{y'_j}z_j(x'_j) =e^{a_j^\top t}e^{y_j}z_j(x_j) =e^{a_j^\top t}x_j
\end{align*}
for every $j$, so $x'=e^t\cdot x$. The reverse implication is the invariance already proved, establishing \ref{eq:maximalinvariant}.
\end{proof}

\begin{proof}[Proof of Corollary~\ref{cor:outputfactor}]
If $q\notin\operatorname{im}A$, finite-dimensional orthogonality gives a vector $t\in\ker A^\top$ with $q^\top t\ne0$. Then $e^t\cdot x=x$ for every $x\in X$, while equivariance would require $F(x)=F(e^t\cdot x)=e^{q^\top t}F(x)$. Since $e^{q^\top t}\ne1$, the equivariance identity forces $F\equiv0$.

Now suppose $q\in\operatorname{im}A$. Write $y(x)\coloneqq\log r(x)$, fix $p$ with $Ap=q$, and define the positive scale monomial $M_p(x)\coloneqq e^{p^\top y(x)}=\prod_{j=1}^m r_j(x_j)^{p_j}$. The scale transformation above yields $M_p(e^t\cdot x)=e^{p^\top A^\top t}M_p(x)=e^{q^\top t}M_p(x)$. If $F$ is equivariant, $H\coloneqq F/M_p$ is consequently invariant. By the completeness of $\mathcal Q$, $H$ is constant on each fiber of $\mathcal Q$, so there is a unique function $f$ on the image of $\mathcal Q$ such that $H=f\circ\mathcal Q$.

That image is the entire stated domain. Indeed, when $m-s>0$, let $C\coloneqq N(N^\top N)^{-1}\in\mathbb R^{m\times(m-s)}$, which satisfies $N^\top C=I_{m-s}$. When $m-s=0$, interpret $C$ as the unique empty linear map. Given arbitrary $z=(z_1,\ldots,z_m)\in\prod_j\mathbb S_j$ and $u\in\mathbb R^{m-s}$, set $y\coloneqq Cu$ and $x_j\coloneqq e^{y_j}z_j$. Then $z_j(x_j)=z_j$ and $N^\top y=u$, proving $\mathcal Q(x)=(z,u)$. This construction establishes \ref{eq:coupledfactor} and uniqueness of $f$ on all of $(\prod_j\mathbb S_j)\times\mathbb R^{m-s}$.

Conversely, if \ref{eq:coupledfactor} holds for any $f$, then $\mathcal Q$ is invariant and the preceding transformation identity shows immediately that $F(e^t\cdot x)=e^{q^\top t}F(x)$. This converse direction completes the necessary-and-sufficient factorization.

Continuity in the forward direction follows from the explicit continuous section just constructed:
\begin{align*}
f(z,u)=e^{-p^\top Cu} F(e^{(Cu)_1}z_1,\ldots,e^{(Cu)_m}z_m).
\end{align*}
The displayed formula makes $f$ continuous whenever $F$ is continuous. The converse follows directly from \ref{eq:coupledfactor}, since RMS, canonicalization on nonzero inputs, logarithm, and the exponential prefactor are continuous.
\end{proof}

\subsection{Proofs of Theorem~\ref{thm:tangent} and Corollary~\ref{cor:quantfunctional}}
\label{app:tangentproof}
\begin{proof}[Proof of Theorem~\ref{thm:tangent}]
By construction, $e_\perp\perp x$ and $x+e=(1+a)x+e_\perp$. Writing $\rho=\|e_\perp\|/\|x\|$, we obtain
\begin{align*}
\|x+e\|^2=\|x\|^2((1+a)^2+\rho^2), \qquad \frac{\langle x,x+e\rangle}{\|x\|\|x+e\|} =\frac{1+a}{\sqrt{(1+a)^2+\rho^2}}.
\end{align*}
The assumption $a>-1$ makes $x+e$ nonzero and fixes the positive orientation of the radial component of $x+e$. Both $\canon_0(x)$ and $\canon_0(x+e)$ have squared norm $d$. The polarization identity gives
\begin{align*}
d_{\mathbb S}(\canon_0(x+e),\canon_0(x))^2 &=\frac1d\|\canon_0(x+e)-\canon_0(x)\|^2\\
&=2\left(1- \frac{\langle x,x+e\rangle}{\|x\|\|x+e\|}\right) =2\left(1-\frac{1+a}{\sqrt{(1+a)^2+\rho^2}}\right),
\end{align*}
which proves the equality in \ref{eq:tangentexact}.

Set $t\coloneqq\rho/(1+a)\ge0$. For $s\ge0$, the function $\phi(s)\coloneqq s-2+2/\sqrt{1+s}$ satisfies $\phi(0)=0$ and $\phi'(s)=1-(1+s)^{-3/2}\ge0$. Thus $2(1-(1+t^2)^{-1/2})\le t^2$. Taking square roots proves the inequality in \ref{eq:tangentexact}. Moreover, $[2(1-(1+t^2)^{-1/2})]^{1/2}=t+O(t^3)$ as $t\to0^+$, and every sufficiently small $t>0$ is realizable in dimension $d\ge2$ by taking $a=0$ and a tangential perturbation with $\|e_\perp\|/\|x\|=t$. Hence no constant smaller than one can replace the unit constant in \ref{eq:tangentexact} uniformly. When $e=ax$ with $a>-1$, $e_\perp=0$ and the exact formula is zero, proving radial rejection without an approximation.
\end{proof}

\begin{proof}[Proof of Corollary~\ref{cor:quantfunctional}]
The coordinatewise error hypothesis implies
\begin{align*}
\RMS(e)=\frac{\|e\|}{\sqrt d}\le\frac\Delta2, \qquad \frac{\|e\|}{\|x\|}\le\frac{\Delta}{2r(x)}=\delta.
\end{align*}
Because $\delta<1$, the same bound also gives $\|e\|<\|x\|$ and hence $Q(x)=x+e\ne0$. Cauchy--Schwarz and orthogonal projection then give $|a|\le\delta$ and $\rho\le\delta$. Since $\delta<1$, we have $a>-1$ and $1+a\ge1-\delta$. The $K$-Lipschitz property of $f$ and Thm.~\ref{thm:tangent} give
\begin{align*}
\|F(Q(x))-F(x)\|_W \le Kd_{\mathbb S}(\canon_0(Q(x)),\canon_0(x)) \le K\frac{\delta}{1-\delta},
\end{align*}
which is \ref{eq:quantfunctional}.
\end{proof}

\subsection{Proof of Proposition~\ref{prop:pgauge}}
\label{app:pgproof}
\begin{proof}[Proof of Proposition~\ref{prop:pgauge}]
Fix the input batch and all parameters other than $W$, and set $W_t\coloneqq e^tW$. Independence of $u$ from $W$ implies $x_t=W_tu=e^tWu=e^t x$. The assumed downstream invariance gives $\Lcal(W_t)=\Lcal(W)$ for every $t\in\mathbb R$. This equality is a functional identity along the whole positive ray and uses no stationarity assumption.

Let $f(t)\coloneqq\Lcal(W_t)$. Wherever the first derivative exists, the chain rule for the Frobenius inner product gives
\begin{align*}
0=f'(t)=\langle\nabla_W\Lcal(W_t),W_t\rangle_F.
\end{align*}
At $t=0$, the identity is the claimed radial Euler pairing.
\end{proof}

\begin{proof}[Derivation of the untyped comparison]
When downstream invariance is absent, because $x=Wu$ is the only dependence on $W$, matrix calculus gives $\nabla_W\Lcal=(\nabla_x\Lcal)u^\top$. Taking the Frobenius pairing with $W$ gives
\begin{align*}
\langle W,\nabla_W\Lcal\rangle_F =\operatorname{tr}(W^\top(\nabla_x\Lcal)u^\top) =(\nabla_x\Lcal)^\top Wu =\langle x,\nabla_x\Lcal\rangle,
\end{align*}
which is the corresponding activation-space radial pairing.
\end{proof}

\subsection{Proof of Proposition~\ref{prop:quant}}
\label{app:quantproof}
\begin{proof}[Proof of Proposition~\ref{prop:quant}]
Let rounding ties be resolved arbitrarily. By definition of $A$, every coordinate lies in $[-A,A]=[-q\Delta,q\Delta]$, so absmax quantization does not clip. Rounding to the nearest grid point gives the coordinatewise estimate $|Q_\Delta(h_{t,i})-h_{t,i}|\le\Delta/2$. Summing the squared coordinate errors yields
\begin{align}
\|Q_\Delta(h_t)-h_t\|_2 \le \frac{\sqrt d\Delta}{2} =\frac{\sqrt dA}{2q}.
\label{eq:quantcoordbound}
\end{align}
On the other hand,
\begin{align}
A=\max_s\|h_s\|_\infty \le \left(\max_s\frac{\|h_s\|_\infty}{r_s}\right)\max_s r_s =\kappa_{\mathrm{amp}}\max_s r_s.
\label{eq:quantampbound}
\end{align}
Since $\|h_t\|_2=\sqrt dr_t$, division of \ref{eq:quantcoordbound} by this norm followed by \ref{eq:quantampbound} gives
\begin{align*}
\frac{\|Q_\Delta(h_t)-h_t\|_2}{\|h_t\|_2} \le \frac{\kappa_{\mathrm{amp}}}{2q}\frac{\max_s r_s}{r_t}.
\end{align*}
For $b\ge2$, $[2(2^{b-1}-1)]^{-1}\le2^{-(b-1)}$, proving the second bound in the proposition.
\end{proof}

\paragraph{Sharpness and effective-bit interpretation.} The scale-spread factor is attained by a simple construction. Take $d=1$, choose $A>0$, set $h_1=A$, and for any integer $m\in\{0,\ldots,q-1\}$ set $h_2=(m+1/2)\Delta$ with $\Delta=A/q$. The first token fixes the shared absmax and the second token lies halfway between two grid points, hence
\begin{align*}
\frac{|Q_\Delta(h_2)-h_2|}{|h_2|} =\frac{1}{2m+1} =\frac{1}{2q}\frac{A}{|h_2|} =\frac{\kappa_{\mathrm{scale}}}{2q}.
\end{align*}
This construction attains the linear dependence on $\kappa_{\mathrm{scale}}$, establishing the sharpness of the scale-spread factor. In the power-of-two relaxation $[2(2^{b-1}-1)]^{-1}\le2^{-(b-1)}$, multiplying the resolution by $\kappa_{\mathrm{scale}}=2^{\log_2\kappa_{\mathrm{scale}}}$ is algebraically the same as reducing the effective resolution budget by $\log_2\kappa_{\mathrm{scale}}$ bits. For finite $b$, the exact level-count statement remains the factor $\kappa_{\mathrm{scale}}/(2q)$ above.

\subsection{Proofs of Lemma~\ref{lem:quantquotupper} and Theorem~\ref{thm:quantminimax}}
\label{app:quantminimaxproof}
\begin{proof}[Proof of Lemma~\ref{lem:quantquotupper}]
For token $t$, write $e_t\coloneqq Q_\Delta(h_t)-h_t$. Nearest-grid rounding and the absmax choice imply $|(e_t)_i|\le\Delta/2$ coordinatewise. Summing the squared coordinate errors gives
\begin{align}
\RMS(e_t)\le\frac{\Delta}{2}, \qquad \frac{\|e_t\|_2}{\|h_t\|_2} =\frac{\RMS(e_t)}{r_t} \le\delta_t\coloneqq\frac{\Delta}{2r_t} \le\delta_*.
\label{eq:quottokenerror}
\end{align}
If $Q_\Delta(h_t)=0$, the middle ratio in \ref{eq:quottokenerror} equals one, contradicting $\delta_*<1$. Hence every canonicalized quantized token is well-defined.

Use the radial--tangential decomposition from Thm.~\ref{thm:tangent}, $a_t\coloneqq\langle h_t,e_t\rangle/\|h_t\|_2^2$, $(e_t)_\perp\coloneqq e_t-a_th_t$, and $\rho_t\coloneqq\|(e_t)_\perp\|_2/\|h_t\|_2$. Cauchy--Schwarz and contractivity of orthogonal projection give $|a_t|\le\|e_t\|_2/\|h_t\|_2\le\delta_t$ and $\rho_t\le\|e_t\|_2/\|h_t\|_2\le\delta_t$. Thus $a_t>-1$ and $1+a_t\ge1-\delta_t$. Applying the exact quotient perturbation bound in \ref{eq:tangentexact} and then the Lipschitz property of $f$ yields
\begin{align*}
\|f(\canon_0(Q_\Delta(h_t)))-f(\canon_0(h_t))\|_W \le K\frac{\rho_t}{1+a_t} \le K\frac{\delta_t}{1-\delta_t} \le K\frac{\delta_*}{1-\delta_*}.
\end{align*}
Taking the maximum over $t$ proves the first inequality in \ref{eq:quantquotupper}. Moreover, $A=\max_t\|h_t\|_\infty\le\kappa_{\mathrm{amp}}\max_t r_t$, and therefore
\begin{align*}
\delta_*=\frac{A}{2q\min_t r_t} \le\frac{\kappa_{\mathrm{amp}}\kappa_{\mathrm{scale}}}{2q},
\end{align*}
which proves the second inequality.
\end{proof}

\begin{proof}[Proof of Theorem~\ref{thm:quantminimax}]
We first upper-bound the worst-case quantity. Every tensor in the definition of $\mathfrak E_q(\Gamma)$ satisfies $\delta_*\le\sqrt2\Gamma/(2q)\le\sqrt2/4<1$. The already-proved bound with $K=1$, together with monotonicity of $x\mapsto x/(1-x)$ on $[0,1)$, gives
\begin{align*}
\mathfrak E_q(\Gamma) \le \frac{\sqrt2\Gamma/(2q)}{1-\sqrt2\Gamma/(2q)} =\frac{\sqrt2\Gamma}{2q-\sqrt2\Gamma}.
\end{align*}

For the matching lower bound, fix $q\ge4$ and $2\le\Gamma\le q/2$, and put $k\coloneqq\lceil q/\Gamma\rceil$. Then $2\le k\le q-1$. For an arbitrary $\Delta>0$, take $d=T=2$ and
\begin{align}
h_1=(q\Delta,0), \qquad h_2=(\Delta/3,k\Delta).
\label{eq:quantlowerconstruction}
\end{align}
The shared absmax $A=q\Delta$ gives a shared quantization step of exactly $\Delta$. Nearest-grid rounding is unambiguous at $\Delta/3$, and
\begin{align}
Q_\Delta(h_1)=h_1, \qquad Q_\Delta(h_2)=(0,k\Delta).
\label{eq:quantlowerround}
\end{align}
The two RMS scales are $r_1=q\Delta/\sqrt2$ and $r_2=\Delta\sqrt{k^2+1/9}/\sqrt2$. The ratio of these RMS scales gives the scale spread of the two tokens,
\begin{align}
\kappa_{\mathrm{scale}}=\frac{r_1}{r_2} =\frac{q}{\sqrt{k^2+1/9}}.
\label{eq:quantlowerspread}
\end{align}
Because $k\ge q/\Gamma$, we have $\kappa_{\mathrm{scale}}\le q/k\le\Gamma$. For the reverse comparison, set $x\coloneqq q/\Gamma\ge2$. Since $k\le x+1$ and $(x+1)^2+1/9\le4x^2$ for $x\ge2$, we obtain $\sqrt{k^2+1/9}\le2q/\Gamma$, hence $\Gamma/2\le\kappa_{\mathrm{scale}}\le\Gamma$. The amplitude factor satisfies $\|h_1\|_\infty/r_1=\sqrt2$ and $\|h_2\|_\infty/r_2=\sqrt2k/\sqrt{k^2+1/9}<\sqrt2$, so $\kappa_{\mathrm{amp}}=\sqrt2$ and the constructed tensor belongs to the supremum defining $\mathfrak E_q(\Gamma)$.

We now exhibit a scale-invariant Lipschitz head sensitive to the rounding error. Let $z\coloneqq\canon_0(h_2)$ and $z'\coloneqq\canon_0(Q_\Delta(h_2))$. In two dimensions, canonicalization multiplies the Euclidean unit direction by $\sqrt2$, so the RMS chord distance between $z$ and $z'$ is the ordinary chord distance between those unit directions. The angle between those directions is $\theta=\arctan(1/(3k))$, and hence $d_{\mathbb S}(z,z')=2\sin(\theta/2)$. For $0\le y\le\pi/2$, $\sin y\ge2y/\pi$, and for $0\le u\le1$, $\arctan u\ge u/2$. Since $1/(3k)\le1$, these inequalities and \ref{eq:quantlowerspread} give
\begin{align*}
d_{\mathbb S}(z,z') &\ge\frac{2\theta}{\pi} \ge\frac{1}{3\pi k} \ge\frac{\kappa_{\mathrm{scale}}}{3\pi q} \ge\frac{\Gamma}{6\pi q},
\end{align*}
where the last step uses $\kappa_{\mathrm{scale}}\ge\Gamma/2$.

Finally, put $v\coloneqq(z'-z)/\|z'-z\|_2$ and define $f:\mathbb S_2\to\mathbb R$ by $f(u)\coloneqq\langle v,u\rangle/\sqrt2$. For all $u,u'\in\mathbb S_2$, we have $|f(u)-f(u')|\le\|u-u'\|_2/\sqrt2=d_{\mathbb S}(u,u')$, so $f$ is $1$-Lipschitz, and the choice of $v$ makes $|f(z')-f(z)|=\|z'-z\|_2/\sqrt2=d_{\mathbb S}(z,z')$. Therefore the scale-invariant map $f\circ\canon_0$ attains a functional error at least $\Gamma/(6\pi q)$ on the admissible tensor in \ref{eq:quantlowerconstruction}. This construction proves the lower inequality in \ref{eq:quantminimax}. Together with the upper inequality, the lower bound establishes the claimed worst-case order.
\end{proof}

\section{Experimental details}
\label{app:details}

\subsection{Language-model training and controlled comparisons}
\label{app:fullsetup}

Under matched controls, we test how unit consistency affects training robustness, functional rescaling invariance, deployment quantization, and unit-system-invariant multi-loss optimization. All runs use a single NVIDIA GH200-96GB GPU. Within each controlled comparison, architectures share data order, budget, and recipe. Only the declared intervention changes, which in the main comparisons is the block structure.

\paragraph{Main language models.} The main language models are decoder-only Transformers with $d=768$, $L=12$, $12$ heads, and context length $1024$. These models have approximately 124M parameters, with all variants matched to within $0.3\%$. We train the models from scratch on FineWeb-Edu \citep{penedo2024fineweb} with the Generative Pre-trained Transformer 2 (GPT-2) byte-pair encoding (BPE) tokenizer \citep{radford2019gpt2} for 2.75B tokens. This budget is approximately Chinchilla-optimal for the model size \citep{hoffmann2022chinchilla}. A disjoint shard supplies a held-out 43M-token validation split.

\paragraph{Training recipe.} The training recipe follows standard practice at this scale \citep{karpathy2022nanogpt,touvron2023llama}. We use Adam with decoupled weight decay (AdamW), setting $\beta=(0.9,0.95)$ and weight decay $0.1$ on matrices only \citep{loshchilov2019adamw}. The peak learning rate is $6\times10^{-4}$, the warmup occupies $2.7\%$ of training and lasts 151 steps at 124M, and cosine decay ends at $10\%$ of the peak rate. The global batch contains 491,520 tokens, the gradient clip is $1.0$, and training uses bf16 autocast and weight tying. Every variant uses rotary positional encoding (RoPE) \citep{su2021roformer}, which strengthens the Pre-LN baseline relative to learned positions.

All eight architectures are replicated with three seeds at 124M. Unless a table explicitly reports a multi-seed aggregate, perturbation and quantization analyses evaluate one model per architecture trained with the primary seed. The remaining seeds quantify training variability. App.~\ref{app:controls} states the seed protocol for every scale.

\paragraph{Architectures compared.} The comparison includes Pre-LN with LayerNorm and GELU in the GPT-2 style \citep{radford2019gpt2,xiong2020layer}. LLaMA-style blocks use RMSNorm and SwiGLU \citep{touvron2023llama}. QK-Norm adds per-head RMS normalization of $Q$ and $K$ \citep{henry2020qknorm,dehghani2023vit22b}. We also include NormFormer \citep{shleifer2021normformer} and LayerScale initialized at $0.1$ \citep{touvron2021cait}.

Three further variants control scale through normalized or explicitly corrected representations. nGPT is the closest normalized-representation prior to \gct{} and uses a sphere-normalized state, per-dimension eigen-learning-rate linear-interpolation updates, unit-norm weights, and QK-Norm \citep{loshchilov2024ngpt}. UCT uses the local-correction modules of App.~\ref{app:uctmodules}, and \gct{} is the canonical design derived in \secref{sec:gct}.

\paragraph{Learning-rate robustness and seeds at 51M.} The robustness study uses $d=512$, $L=8$ models with parameter counts from 50.9M to 51.5M, referred to as 51M models. Small-scale learning-rate sweeps are used to study training instabilities that also arise at larger scales \citep{wortsman2024proxies}. Pre-LN, LLaMA-style, QK-Norm, nGPT, and \gct{} train for 600M tokens with learning rates in $\{3,6,12,24,48\}\times10^{-4}$, spanning a factor of $16$. The protocol includes the strong QK-Norm baseline in every stress test. Each architecture also has three seeds at the default learning rate.

\paragraph{UVI-based early stopping.} At the additional learning rate $9.6\times10^{-3}$, we apply the same preset stopping rule to Pre-LN, LLaMA-style, nGPT, and \gct{}. At diagnostic steps within the first $10\%$ of the training budget, excluding initialization, the monitor terminates a run if $\UVI_{\mathrm{res}}+\UVI_{\mathrm{attn}}>\intervtau$. Pre-LN and LLaMA-style cross this threshold at step 100 of 1220, while nGPT and \gct{} complete training under the same rule. These early stops record threshold crossings, not observed numerical divergence.

\paragraph{Depth probes and ablations.} Separate fixed-budget depth probes use $L=24$ and $L=48$, which are respectively $3\times$ and $6\times$ the tuning depth. These probes use 400M tokens with unchanged hyperparameters for the same five architectures, and the depth table reports the primary seed. The \gct{} component ablations remove $P^\perp$, proposal canonicalization, or both components under the 600M-token $L=8$ protocol. We additionally repeat the $P^\perp$ ablation under the 400M-token depth protocol to measure how the projection's effect changes during depth transfer.

\paragraph{Larger language models.} The same shared recipe uses learning rate $6\times10^{-4}$ with an identical schedule and data order at larger scales. The 350M models use $d=1024$, $L=24$, 2B tokens, and three seeds. The 1.0B models use $d=1792$, $L=24$, 4B tokens, and two seeds. Pre-LN, LLaMA-style, QK-Norm, nGPT, and \gct{} are all represented. These models extend the residual-stream quantization and parameter-fingerprint studies beyond the 124M setting.

\paragraph{Scale perturbation.} Using the trained 124M models, we measure validation loss under $h\to ch$ at the embedding output and at mid-depth for $c\in[10^{-2},10^2]$. We also measure per-block equivariance or invariance error in fp32 (Figure~\ref{fig:perturb}; Table~\ref{tab:perturb}).

\subsection{Quantization and deployment evaluations}

\paragraph{PTQ.} We apply PTQ to the same trained models to evaluate sensitivity to finite-precision storage and computation. The protocols include symmetric per-channel weight quantization at 8 and 4 bits and int8 activation quantization at every linear input, using both dynamic and calibration-based static scales in the standard integer-inference format \citep{jacob2018quantization}. The per-tensor, per-channel, and per-token granularities compared in \secref{sec:exp-quant} are the standard PTQ options \citep{nagel2021whitepaper}. Weight quantization uses round-to-nearest rather than error-compensated rounding \citep{frantar2023gptq}, because the comparison isolates activation- and stream-scale management rather than weight-rounding quality.

We also quantize the residual stream at every block boundary to int8 and int4. This setting arises when hidden states are cached, offloaded, or transmitted across pipeline stages. The stream protocols use static or dynamic per-tensor scales and per-token scales. We additionally evaluate scale migration through SmoothQuant with $\alpha=0.5$ and per-site static calibration on the linear path \citep{xiao2023smoothquant}, together with calibrated per-channel static scales for the stored stream. The logit head remains in full precision as specified in App.~\ref{app:auditprotocols}. A microbenchmark compares the deployment cost of per-token and per-tensor activation-scale management on the same device.

\paragraph{Off-the-shelf audit.} To examine the quantization pattern in independently trained models, we record the depth-scale span $\kappa_{\mathrm{depth}}\coloneqq\max_\ell\bar r_\ell/\min_\ell\bar r_\ell$, where $\bar r_\ell$ is the mean token RMS at block input $\ell$. We then apply identical int8 and int4 quantization hooks to the four publicly available pretrained OpenAI GPT-2 models from 124M to 1.5B parameters \citep{radford2019gpt2}. The depth-scale span $\kappa_{\mathrm{depth}}$ is a descriptive depth-drift statistic and is distinct from the within-tensor $\kappa_{\mathrm{scale}}$ of Prop.~\ref{prop:quant}. We evaluate these pretrained models on the same FineWeb-Edu validation bins with the same GPT-2 BPE through an independent PyTorch reimplementation. The implementation reproduces the published WikiText-103 \citep{merity2016pointer} validation perplexities \otspplladder{} for the 124M, 355M, 774M, and 1.5B pretrained models on non-overlapping blocks, confirming faithful weight loading.

\paragraph{Quantized downstream accuracy.} The 1B models trained independently with a second seed are evaluated before and after residual-stream quantization on four tasks. Language Modeling Broadened to Account for Discourse Aspects (LAMBADA) uses the OpenAI variant and greedy final-word accuracy \citep{paperno2016lambada}. Harder Endings, Longer Contexts, and Low-shot Activities for Situations With Adversarial Generations (HellaSwag) uses byte-length-normalized four-way accuracy \citep{zellers2019hellaswag}. The remaining tasks are Physical Interaction: Question Answering (PIQA) \citep{bisk2020piqa} and the easy subset of the Allen Institute for AI Reasoning Challenge (ARC-Easy) \citep{clark2018arc}. The evaluation uses lm-evaluation-harness-compatible prompts and log-likelihood scoring under the training tokenizer.

\subsection{Vision, optimizer, width, and multi-loss evaluations}

\paragraph{First vision benchmark.} The benchmark on the 100-class Canadian Institute for Advanced Research (CIFAR) dataset uses ViTs at the scale of the tiny Data-efficient Image Transformer (DeiT) variant, denoted DeiT-Ti, with $d=192$, $L=12$, and patch size $4$. Models train from scratch for 300 epochs with the standard heavy-augmentation recipe \citep{touvron2021deit}. The recipe uses AdamW with learning rate $10^{-3}$, batch size 512, weight decay $0.05$, a 10-epoch warmup, cosine decay, RandAugment$(2,9)$, mixup $0.8$, CutMix $1.0$, label smoothing $0.1$, and random erasing $0.25$. Variants are Pre-LN, QK-Norm, nGPT, UCT, and \gct{}. Evaluation covers clean top-1 accuracy, the Gaussian-noise, contrast, and brightness corruptions of the standard common-corruption robustness protocol \citep{hendrycks2019benchmarking}, and multiplicative rescaling of the normalized input as an input-unit shift.

\paragraph{Vision benchmark on Tiny-ImageNet.} A second benchmark trains small ViT (ViT-S) models from scratch on Tiny-ImageNet-200 for 200 epochs. These models use $d=384$, $L=12$, patch size $8$, images of size $64\times64$, 100k training images, and 10k validation images. All eight architectures use the same optimizer and augmentation recipe with a 15-epoch warmup.

\paragraph{Optimizer and width evaluations.} Two further axes test generality beyond one optimizer and one width. The optimizer-pairing axis uses a 51M experiment that replaces AdamW on hidden matrices with the shape-aware orthogonalized Muon optimizer \citep{jordan2024muon}. We test $\eta_{\mathrm{Muon}}\in\{0.01,0.02\}$ for \gct{}, Pre-LN, and QK-Norm, while auxiliary parameters retain the AdamW recipe. This comparison tests compatibility with a duality-map optimizer.

The width axis repeats the learning-rate robustness protocol at $d\in\{256,1024\}$ with $L=8$ and fixed $d_h=64$. The same axis also composes a learning-rate-only $\mu$P rule with \gct{} and Pre-LN at the respective $d=512$-tuned learning rates \citep{yang2021tp5}. Hidden-matrix learning rates scale by $512/d$, while initialization and multipliers remain unchanged.

\paragraph{PINN multi-loss.} The PINN study covers Burgers with $\nu=0.01/\pi$ and a Cole--Hopf reference, together with Helmholtz using $a_1=1$ and $a_2=4$ \citep{raissi2019pinn,wang2021understanding}. Both problems use a $4\times64$ tanh MLP, Adam with learning rate $10^{-3}$, 20K steps with a cosine tail, and 10K collocation points. The field is re-expressed in rescaled units as $\hat u\coloneqq\kappa u$ for $\kappa\in\{10^{-2},1,10^2\}$. The PDE is rewritten accordingly, leaving the physical problem unchanged.

Methods include equal weights and the common manual choice $\lambda_{\mathrm{sup}}=100$, which is tuned at $\kappa=1$ and then transferred. We also evaluate uncertainty weighting \citep{kendall2018uncertainty}, gradient-statistics annealing, known as learning-rate annealing (LRA) \citep{wang2021understanding}, neural tangent kernel (NTK) trace balancing \citep{wang2022ntk}, and the NDA-typed objective. NDA types the whole graph through the output head $\hat u\coloneqq s_{\mathrm{out}}\hat{\bar u}$ from \secref{sec:gct} and the objective in \ref{eq:lossfact}. All characteristic scales are derived from observables such as data RMS, forcing RMS and forcing length scale, and operator units. The method requires no tuning, adaptivity, or access to the solution. App.~\ref{app:auditprotocols} gives the characteristic scales and baseline settings. We use five seeds per cell and report relative $L_2$ error of the recovered physical field $\hat u/\kappa$.

\subsection{Controls and model implementation}
\label{app:controls}

\paragraph{Seed protocol.} Every language-model family draws training seeds from one shared list, and each architecture inside a controlled comparison reuses that same list to isolate the effect of the declared intervention. At 51M and at 124M, each architecture trains three seeds at the default learning rate, the 350M family trains three seeds, and the 1.0B family trains two seeds. Learning-rate sweep points, component ablations, and depth probes are single-seed. One member of the shared list serves as the primary seed at every scale.

Analyses that report one result per architecture evaluate the model trained with the primary seed unless the table states otherwise. These analyses include the scale-perturbation study of Table~\ref{tab:perturb}, the PTQ results of Tables~\ref{tab:quant1b} and~\ref{tab:quant}, the parameter-gauge fingerprint of Table~\ref{tab:paramgauge}, the fixed-budget depth probe of Table~\ref{tab:depth}, and the width and $\mu$P grid of Table~\ref{tab:widthmup}. The remaining seeds supply the multi-seed aggregates and quantify training variability.

We use models trained independently with a second seed for the 350M column of Table~\ref{tab:quantscale}, the repeat of the 1B stream-quantization and parameter-fingerprint results in \secref{sec:exp-quant}, the downstream-accuracy evaluation of Table~\ref{tab:quantds}, and the serving-latency benchmark. The PINN study uses five seeds per cell and reports the mean across those five seeds, and the post-layer-normalization (Post-LN) reference at $L=8$ in App.~\ref{app:trainingresults} uses five seeds.

\paragraph{Weight-decay controls.} Every architecture in a controlled comparison uses weight decay $0.1$ on matrices. The effect of weight decay can nevertheless differ across architectures, because nGPT normalizes weight rows and \gct{} canonicalizes branch proposals. On scale-invariant parameter blocks, weight decay affects the effective learning rate through the parameter norm \citep{vanlaarhoven2017l2,li2020exponential}.

We assess sensitivity to this shared setting with weight-decay controls. Re-running nGPT without weight decay, as recommended by \citet{loshchilov2024ngpt}, changes nGPT's 51M loss by \ngptwddelta{} nats, and \wdfairnessclaim{}. \ngptbespokeclaim{}

\paragraph{UCT corrective modules.}
\label{app:uctmodules}
UCT makes residual additions unit-consistent and gives softmax and gates dimensionless inputs through three added modules: an RMS-matching residual adapter $h+\alpha_\ell[\RMS(h)/\RMS(a)]a$, the empirical attention temperature from \ref{eq:tau}, and normalized gates. We use UCT to test whether these three local corrections improve a conventional Transformer's performance and robustness to changes in scale.

\paragraph{UCT details.} Empirical attention scales $s_Q$ and $s_K$ are per-head stop-gradient RMS batch statistics with an exponential moving average (EMA) of momentum $0.99$ used at evaluation. The gate scale $s_a$ is a per-layer scalar EMA. The residual adapter uses per-token RMS matching with a learnable scalar $\alpha_\ell$ for each branch. The pre-normalizers and configured converter denominators use fixed numerical floors of $\epsilon=10^{-6}$. Both these floors and the evaluation-time EMA scales stay numerically fixed under an injected rescaling. When the hidden state is scaled down, the fixed floors contribute more to the normalization denominators, and the stored EMA scales no longer match the activations. Dividing rescaled activations by these unchanged quantities does not undo the rescaling, which explains the loss of equivariance under the most severe down-scaling in Figure~\ref{fig:perturb}.

\paragraph{\gct{} details.} Canonicalizer floors are fixed numerically at $10^{-12}$ inside the square root. The floors have the formal squared unit $\uof{\epsilon_x}=\uof{x}^2$. The numerical canonicalizer floors stay fixed under an injected rescaling, producing the finite defect in \ref{eq:canondefect}. Exact invariance refers to the zero-floor or co-transforming-floor statement. The implementation uses the projection denominator $\|\bar h\|^2+\epsilon_\perp$ with $\epsilon_\perp=10^{-6}$, whereas the exact orthogonality and rotation results use $\epsilon_\perp=0$ as stated in App.~\ref{app:floordefect}. Per-head gains and the logit gain initialize at $1.0$. Attention and MLP sub-blocks are sequential, and each sub-block has a separate $\alpha_\ell$. Embeddings are canonicalized at entry, followed by a final canonicalization before the tied head.

\paragraph{Heads, scales, and parameters.} The classification head produces dimensionless logits from $\canon(\bar h_L)$ with a learnable dimensionless logit gain. For regression and PINN predictions that require units, the head first produces a dimensionless value $\hat{\bar y}$. Multiplying by an explicit output scale gives $\hat y\coloneqq s_y\hat{\bar y}$, which restores the magnitude and units. We parameterize each learned scale as $s\coloneqq s_0e^{\ell_s}$, where the fixed reference $s_0$ has the required unit and the trainable log multiplier $\ell_s$ is dimensionless. Within each block, linear maps mix dimensionless coordinates using dimensionless weights. Any required unit conversion is performed by an explicit scale factor, such as $s_y$ at the output. For tasks with dimensionless outputs, such as language modeling, the computation does not depend on the scale state, and we implement the residual stream using only canonical coordinates.

\subsection{Reference PyTorch implementation of \gct{}}
\label{app:gctcode}

Listing~\ref{lst:gctcode} is a compact extraction of the implementation used in the experiments. The reference block fixes the reported configuration, which uses scalar branch steps and enables both the tangential projection and proposal canonicalization. Only diagnostic collection, ablation switches, and standard initialization boilerplate are removed. The listing retains every scale-sensitive operation: fp32 canonicalization with a return to the input dtype, per-head canonical $Q$ and $K$, the dimensionless normalized-SiLU gate, the regularized tangential projection, and the two sequential canonical residual additions. RoPE is included and is norm preserving, while \texttt{scaled\_dot\_product\_attention} supplies the standard $1/\sqrt{d_h}$ factor.

\begin{lstlisting}[style=ndapython,float=t!,caption={Finite-floor PyTorch reference implementation of the \gct{} block used in the experiments.},label={lst:gctcode}]
import torch
import torch.nn as nn
import torch.nn.functional as F

def canon(x, eps=1e-12):
    xf = x.float()
    inv_rms = torch.rsqrt(xf.square().mean(-1, keepdim=True) + eps)
    return (xf * inv_rms).type_as(x)

def apply_rope(x, cos, sin):
    x1, x2 = x[..., 0::2], x[..., 1::2]
    c, s = cos[None, None], sin[None, None]
    y = torch.stack((x1 * c - x2 * s, x1 * s + x2 * c), dim=-1)
    return y.flatten(-2)

class GCTBlock(nn.Module):
    def __init__(self, dim, n_heads, mlp_width, depth,
                 eps_canon=1e-12, eps_perp=1e-6):
        super().__init__()
        assert dim % n_heads == 0
        self.dim, self.n_heads = dim, n_heads
        self.head_dim = dim // n_heads
        self.eps_canon, self.eps_perp = eps_canon, eps_perp
        self.qkv = nn.Linear(dim, 3 * dim, bias=False)
        self.attn_out = nn.Linear(dim, dim, bias=False)
        self.gain = nn.Parameter(torch.ones(n_heads))
        self.w_gate = nn.Linear(dim, mlp_width, bias=False)
        self.w_up = nn.Linear(dim, mlp_width, bias=False)
        self.mlp_out = nn.Linear(mlp_width, dim, bias=False)
        alpha0 = 1.0 / (2.0 * depth) ** 0.5
        self.alpha_att = nn.Parameter(torch.tensor(alpha0))
        self.alpha_mlp = nn.Parameter(torch.tensor(alpha0))

    def _attention(self, x, cos, sin, causal):
        B, T, D = x.shape
        H, dh = self.n_heads, self.head_dim
        qkv = self.qkv(x).view(B, T, 3, H, dh)
        q, k, v = qkv.permute(2, 0, 3, 1, 4).unbind(0)
        q = canon(q, self.eps_canon) * self.gain.view(1, H, 1, 1)
        k = canon(k, self.eps_canon)
        q = apply_rope(q, cos, sin).type_as(v)
        k = apply_rope(k, cos, sin).type_as(v)
        y = F.scaled_dot_product_attention(q, k, v, is_causal=causal)
        y = y.transpose(1, 2).contiguous().view(B, T, D)
        return self.attn_out(y)

    def _mlp(self, x):
        a = canon(self.w_gate(x), self.eps_canon)
        gate = a * torch.sigmoid(a)
        return self.mlp_out(gate * self.w_up(x))

    def _update(self, h, proposal, alpha):
        hf, pf = h.float(), proposal.float()
        coeff = (pf * hf).sum(-1, keepdim=True)
        coeff = coeff / (hf.square().sum(-1, keepdim=True)
                         + self.eps_perp)
        tangent = (pf - coeff * hf).type_as(proposal)
        tangent = canon(tangent, self.eps_canon)
        return canon(h + alpha * tangent, self.eps_canon)

    def forward(self, h, cos, sin, causal=True):
        h = canon(h, self.eps_canon)
        h = self._update(
            h, self._attention(h, cos, sin, causal), self.alpha_att)
        h = self._update(h, self._mlp(h), self.alpha_mlp)
        return h
\end{lstlisting}

The two floors in Listing~\ref{lst:gctcode} are the reported numerical values. Setting \texttt{eps\_canon} and \texttt{eps\_perp} to zero recovers the exact canonicalizer and projector on the respective nondegenerate domains. The displayed finite-floor code has precisely the defects derived in App.~\ref{app:floordefect}. At network level, the embedding output is canonicalized before the first block, the last block output is canonicalized again before the tied language-model head, and a learnable dimensionless scalar multiplies the logits. These two wrapper operations are omitted from the listing because the wrapper operations repeat \texttt{canon} and contain no additional scale-sensitive mechanism.

\subsection{Task-specific and audit protocols}
\label{app:auditprotocols}

\paragraph{Parameter-gauge audit predictions.} \gct{} nondimensionalizes every probed output. The architecture applies $\canon$ to $q$, $k$, and the gate preactivation, and applies $\canon\circ P^{\perp}$ to both branch proposals. All seven roles are null. Normalizing every weight row in nGPT's forward pass makes all roles null. QK-Norm applies per-head RMSNorm only to $q$ and $k$. In NormFormer, LayerNorm after the attention out-projection types $v$ and the out-projection. NormFormer's mid-MLP LayerNorm acts after the GELU, leaving the GELU's input scale uncorrected. Pre-LN, LLaMA-style, and LayerScale have no nulls because none places a nondimensionalizer between a probed matrix and the residual stream. In evaluation mode, UCT's live-RMS residual adapter types both branch outputs, covering $v$, the attention output, the up-projection, and the MLP output. The $q$, $k$, and gate interfaces retain radial sensitivity because the corrections at these interfaces use frozen EMA statistics.

\paragraph{Parameter-gauge audit protocol.} We use fp32 evaluation mode and the first 8 fixed validation batches. One backward pass yields every $\langle W,\nabla_W\Lcal\rangle$, with dot products accumulated in 64-bit floating-point precision and reported as the maximum over layers for each role. A central finite difference at the middle layer uses $c=e^{\pm0.1}$ and cross-checks two roles per architecture. This secant over $\pm0.1$ log-units includes curvature. For nonnull cells, the secant checks the sign and order of magnitude, while for null cells, the pairing and secant both vanish. The identical audit runs on the five trained 1B models, yielding 34 cells because UCT is absent at 1B and Pre-LN's GELU MLP has no up-projection.

Blocks that realize the same activation rescaling must carry equal pairings. The $q$ and $k$ rows both rescale the logits, the $v$ rows and out-projection both rescale the attention write, and the up- and down-projections both rescale the MLP write. The measured columns agree, providing an internal consistency check on the pairing measurement. For stop-gradient scale estimators, as in UCT with batch statistics active, the finite difference and the autograd pairing measure function invariance and the optimizer-visible gradient, respectively. The two measurements are reported separately in App.~\ref{app:rescalingaudits}.

\paragraph{PINN details.} Both problems use the standard two-term split between the PDE residual and combined supervision to give the adaptive balancers a well-conditioned pair. Burgers uses $N_f=10^4$ uniformly sampled interior points and combines 200 initial-condition points with 200 boundary-condition points into one supervision term. The Burgers reference is computed by Cole--Hopf with 100-node Gauss--Hermite quadrature on a $256\times101$ grid. Helmholtz uses $N_f=10^4$, 400 boundary points, and the manufactured solution $\sin(\pi x)\sin(4\pi y)$.

The adaptive baselines follow the published method specifications. LRA follows \citet{wang2021understanding}. Every 10 steps, $\lambda_{\mathrm{sup}}$ tracks $\|\nabla_\theta\Lcal_{\mathrm{pde}}\|/\|\nabla_\theta\Lcal_{\mathrm{sup}}\|$ as an EMA with weight $0.1$, clamped to $[10^{-2},10^2]$. Uncertainty weighting optimizes $\sum_i e^{-s_i}\Lcal_i/2+s_i/2$. NTK balancing follows \citet{wang2022ntk}. Every 100 steps, $\lambda_k\coloneqq\operatorname{tr}(K)/\operatorname{tr}(K_k)$, with $\operatorname{tr}(K_k)$ estimated as $N_k\times$ the mean squared parameter-gradient norm of 16 randomly sampled per-point residuals of term $k$.

The NDA-typed method types both the head and the objective. Characteristic scales are computed from observables only. For Burgers, $s_u\coloneqq\RMS(\text{initial-condition data})$, which gives output scale $s_{\mathrm{out}}\coloneqq s_u$, combined-supervision scale $S_{\mathrm{sup}}\coloneqq s_u^2$, and $S_{\mathrm{pde}}\coloneqq(s_u/t^\ast)^2$ with domain time scale $t^\ast=1$.

Because the Helmholtz boundary data are zero, we estimate the field scale from the forcing magnitude and forcing length scale. Define $s_q\coloneqq\RMS(q)$ and $\kappa_q^2\coloneqq\E_x\|\nabla q(x)\|_2^2/\E_x|q(x)|^2$, with the expectations estimated over the forcing sample. Our single-mode manufactured forcing is nonresonant and satisfies $\kappa_q^2\ne k^2$. The operator balance is $|\Delta u+k^2u|\sim|\kappa_q^2-k^2|u^\ast\sim s_q$, giving $u^\ast=s_q/|\kappa_q^2-k^2|$. For a representative forcing sample, this observable estimate is $0.515$. Across the five experimental seeds, the estimate ranges from $0.488$ to $0.523$, around the true continuous-field RMS of $0.5$. We then set $s_{\mathrm{out}}\coloneqq u^\ast$, $S_{\mathrm{sup}}\coloneqq u^{\ast2}$, and $S_{\mathrm{pde}}\coloneqq s_q^2$. The scales come directly from data and operator statistics, without using the reference solution or tuning the loss weights.

\paragraph{Quantization details.} All schemes use simulated quantization in the standard PTQ protocol. Evaluation uses the first 32 fixed batches of the validation stream, which are the same deterministic batches as in the perturbation study. The shared base points agree to numerical precision, as noted in the accompanying discussion of Table~\ref{tab:perturb}. Weights use symmetric per-output-channel absmax scaling at 8 or 4 bits, applied to every linear layer except the tied logit head. For nGPT, whose forward pass normalizes weight rows, quantization applies to the effective normalized weights that a deployment would store.

W8A8 activations use symmetric per-tensor int8 at every linear input. The dynamic variant recomputes the absmax per call, while the static variant fixes per-call-site scales from absmax calibration over 8 held-out training batches. Residual-stream quantization maps $h$ to int8 at every block input using dynamic per-tensor absmax, calibrated per-layer static absmax, or per-token absmax scales. We intercept calls at the torch-function level to treat functional and module linear calls identically across architectures.

\paragraph{Migration variants.} SmoothQuant is applied at each linear site with $\alpha=0.5$. Per-channel activation absmax values $a_j$ are calibrated over the same 8 batches, and migration uses $s_j\coloneqq a_j^\alpha/\max_i|W_{ij}|^{1-\alpha}$. Weights $W\operatorname{diag}(s)$ are requantized per output channel. Activations $x\operatorname{diag}(s)^{-1}$ use per-tensor static quantization with the scale implied by the calibrated $a_j/s_j$. For the stored stream, we calibrate per-channel static absmax scales at each block site, which is the storage analogue of migration. The logit head stays in full precision throughout.

\paragraph{Quantized downstream-accuracy details.} Table~\ref{tab:quantds} uses standard lm-evaluation-harness formats. LAMBADA-OpenAI accuracy requires greedy decoding of every token in the final word. HellaSwag validation uses the activity label and context, preprocessed endings, and byte-length-normalized scoring. PIQA validation uses the prompt template \texttt{Question:\ldots Answer:}. ARC-Easy test choices are scored as completions. We use right padding under causal masking, truncate contexts from the left to 1024 tokens, and score with fp32 log-softmax over bf16 logits.

\paragraph{Breadth-evaluation details.} Muon uses five Newton--Schulz iterations and orthogonalized Nesterov momentum with $\mu=0.95$ on hidden two-dimensional matrices. The optimizer applies shape-aware step scaling $\sqrt{\max(1,d_{\mathrm{out}}/d_{\mathrm{in}})}$ and decoupled weight decay 0.1. Embeddings, gains, and scalars use AdamW with the recipe settings. Both groups follow the shared cosine schedule. We report the better result over $\eta_{\mathrm{Muon}}\in\{0.01,0.02\}$.

\subsection{Reproducibility and numerical verification}

\paragraph{Reproducibility.} Reported values are computed from the corresponding run logs, and protocol descriptions follow the executed configurations. Experimental protocols are described in this appendix, and Listing~\ref{lst:gctcode} provides a compact reference implementation of the \gct{} block. The parameter-gauge audit of Prop.~\ref{prop:pgauge} uses one backward pass to measure radial sensitivity by weight-matrix role.

\paragraph{Numerical consistency checks.} Automated fp32 tests check the numerical idempotency of $\canon$. The maximum coordinate error is below $10^{-5}$, and the maximum deviation of the mean square from $1$ is below $10^{-4}$. The tests verify \gct{} block invariance under $h\to ch$ with relative error below $10^{-4}$. The same tests also verify full-network invariance at the embedding output for $c=10^2$ and in the middle of the stream for $c=10^{-2}$, with maximum logit error below $10^{-3}$. UCT block equivariance under the tested up-scaling $c=10^2$ has relative error below $10^{-3}$. Attention logit variance at \gct{} initialization remains between $0.5$ and $2.0$. The NDA-typed PINN objective is numerically invariant at initialization, with relative spread below $10^{-3}$ for $\kappa\in\{10^{-2},1,10^2\}$. The small observed variation across $\kappa$ is consistent with nondeterministic GPU reductions in second-order autograd rather than systematic unit dependence.

\clearpage
\section{Full results}
\label{app:full}
\suppressfloats[t]

This section collects the detailed training, intervention, deployment, breadth, and parameter-gauge results summarized in the main text. Table~\ref{tab:paramgauge} reports the per-role parameter-gauge fingerprint of Prop.~\ref{prop:pgauge}, Table~\ref{tab:widthmup} reports the width-transfer and $\mu$P composition grid, and Figure~\ref{fig:vitrobust} reports the CIFAR-100 robustness grid.

\paragraph{Results guide.} Figure~\ref{fig:qual-uvi-fingerprint} shows how UVI distinguishes attention-scale control from residual-stream scale growth. The useful high-rate behavior across widths is reported in Table~\ref{tab:widthmup}, and App.~\ref{sec:exp-breadth} also evaluates compatibility with Muon. Tables~\ref{tab:vit} and~\ref{tab:vittin} report the two vision benchmarks. Additional quantization evidence covers downstream accuracy, publicly available pretrained GPT-2 models, and the 350M models (Tables~\ref{tab:quantds},~\ref{tab:offtheshelf}, and~\ref{tab:quantscale}). \secref{sec:exp-quant} reports the independent-seed PTQ repeat and downstream accuracy. App.~\ref{app:quantresults} collects kernel verification, scale-management costs, serving measurements, migration controls, and the KV-cache control. App.~\ref{app:rescalingaudits} details the UCT finite-floor failure and the full rescaling results.

\subsection{Learning-rate tables, seeds, ablations, and transfer}
\label{app:trainingresults}

\begin{table}[t!]
\centering\small
\caption{Clean validation loss for the matched 124M models trained on 2.75B tokens. Each row reports the mean and standard deviation over three seeds under the shared recipe. Values are rounded to three decimal places. A displayed standard deviation of $0.000$ denotes variation below that precision.}
\label{tab:clean124m}
\begin{tabular}{lc}
\toprule
architecture & validation loss \\
\midrule
Pre-LN & $3.172 \pm 0.002$ \\
LLaMA-style & $3.175 \pm 0.003$ \\
QK-Norm & $\mathbf{3.147 \pm 0.002}$ \\
NormFormer & $3.198 \pm 0.008$ \\
LayerScale & $3.167 \pm 0.003$ \\
nGPT & $3.170 \pm 0.003$ \\
Corrective UCT (ours) & $3.256 \pm 0.004$ \\
\gct{} (ours) & $3.170 \pm 0.000$ \\
\bottomrule
\end{tabular}
\end{table}

\begin{table}[t!]
\centering\small
\caption{Final validation loss across a $16\times$ learning-rate range for 51M models trained on 600M tokens. The additional octave at $9.6\times10^{-3}$ is reported wherever training runs to completion. \textsc{abort} marks a run terminated early because the run's UVI score exceeded the preset threshold, as specified in App.~\ref{app:details}.}
\label{tab:sweep}
\begin{tabular}{lcccccc}
\toprule
 & \multicolumn{6}{c}{peak learning rate}\\
\cmidrule(lr){2-7}
architecture & $3\times10^{-4}$ & $6\times10^{-4}$ & $1.2\times10^{-3}$ & $2.4\times10^{-3}$ & $4.8\times10^{-3}$ & $9.6\times10^{-3}$ \\
\midrule
Pre-LN & 4.232 & 3.943 & 3.750 & 3.876 & 3.881 & \textsc{abort} \\
LLaMA-style & 4.248 & 3.892 & 3.714 & 3.802 & 3.883 & \textsc{abort} \\
QK-Norm & 4.153 & 3.840 & 3.700 & 3.740 & 3.879 & 3.897 \\
\midrule
nGPT & 4.383 & 3.893 & 3.649 & 3.563 & 3.507 & 3.500 \\
\gct{} (ours) & 4.373 & 3.885 & 3.658 & 3.548 & 3.517 & 3.513 \\
\bottomrule
\end{tabular}
\end{table}

\begin{table}[t!]
\centering\small
\caption{Seed variability and \gct{} component ablations. Both tables follow the Table~\ref{tab:sweep} protocol at learning rate $6\times10^{-4}$. The seed table reports mean $\pm$ standard deviation over three seeds, while the ablations use the primary seed.}
\label{tab:seeds-depth-abl}
\begin{tabular}[t]{lc}
\toprule
\multicolumn{2}{c}{\textbf{Seed variability at 51M}}\\
architecture & val loss\\
\midrule
Pre-LN & 3.9405 $\pm$ 0.0039 \\
LLaMA-style & 3.9024 $\pm$ 0.0167 \\
QK-Norm & 3.8445 $\pm$ 0.0037 \\
\midrule
nGPT & 3.8934 $\pm$ 0.0004 \\
\gct{} (ours) & 3.8944 $\pm$ 0.0096 \\
\bottomrule
\end{tabular}\qquad
\begin{tabular}[t]{lc}
\toprule
\multicolumn{2}{c}{\textbf{\gct{} component ablations}}\\
variant & val loss\\
\midrule
\gct{} (full) & 3.885 \\
without $P^\perp$ & 3.882 \\
without proposal canon. & 4.486 \\
without either component & 4.083 \\
\bottomrule
\end{tabular}
\end{table}

\begin{figure}[t!]
\centering
\includegraphics[width=\textwidth]{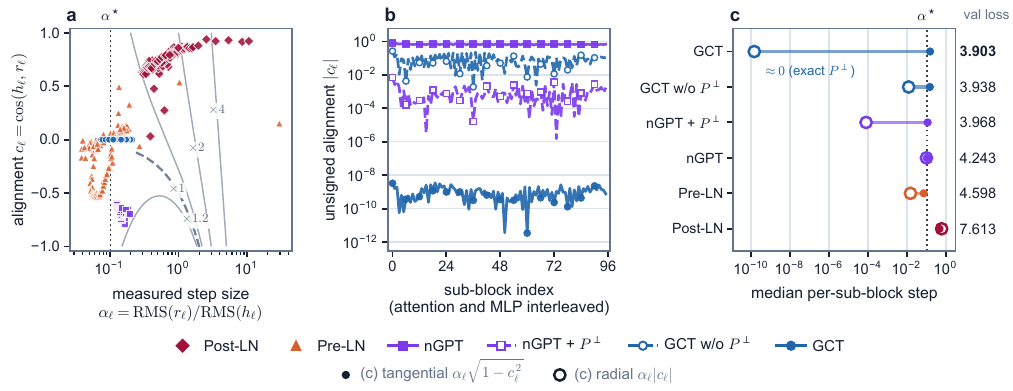}
\caption{Measured residual steps and alignments $(\alpha_\ell,c_\ell)$ during depth transfer to $L=48$, six times the tuning depth. Results use primary-seed fixed-budget probes. The shared legend below the panels names every architecture. (a) Measured step size against alignment. Grey contours give the per-sub-block stream-scale gain $s_{\ell+1}/s_\ell=(1+\alpha_\ell^2+2\alpha_\ell c_\ell)^{1/2}$ of \ref{eq:drift}, the dashed $\times1$ contour being the scale-preserving locus, and the dotted line marks $\alpha^\star=1/\sqrt{2L}$. (b) Removing $P^\perp$ from \gct{} increases alignment, whereas adding the projection to nGPT reduces alignment and changes loss from $4.243$ to $3.968$. (c) Tangential--radial step decomposition, with the final validation loss of each run in the right-hand column. Points are medians over the final three measurements. Hyperparameters transfer unchanged from $L=8$.}
\label{fig:qual-depth-schedule}
\end{figure}

\paragraph{Seed variability.} The three-seed results in Table~\ref{tab:seeds-depth-abl} show similar mean losses for \gct{} and nGPT relative to the observed seed variability of both models. The observed seed variability leaves the ordering of the two architectures unresolved.

\paragraph{Residual schedules in the evaluated models.} \secref{sec:dynamics} identifies the stability-relevant quantity as the positive measured RMS ratio $\alpha_\ell$. In the zero-floor formulation of \gct{}, proposal canonicalization makes that pre-retraction ratio exactly the magnitude $|\alpha_\ell|$ of the learned scalar coefficient, initialized at $1/\sqrt{2L}$. With the fixed $10^{-12}$ implementation floor, that identity is exact only up to the finite-floor defect documented in \ref{eq:canondefect}.

Residual-scaling methods control the measured ratio $\alpha_\ell$, often indirectly through skip or branch coefficients. These coefficients should not be identified with the RMS ratio unless branch and stream scales are matched. The original Post-LN addition uses raw coefficients of one. ReZero and SkipInit initialize a branch gate at zero \citep{bachlechner2020rezero,de2020skipinit}. DeepNet uses the skip coefficient $(2N)^{1/4}$ together with a matching initialization down-scale calibrated for the very-deep regime \citep{wang2022deepnet}. We next evaluate the coefficient-one Post-LN reference and report the corresponding depth sensitivity.

In Post-LN, the conventional raw coefficient of one need not produce a measured ratio of $\alpha_\ell=1$. Under the shared recipe, the five $L=8$ anchor runs span losses from $4.14$ to $7.60$, with mean $5.47\pm1.64$. The $L=24$ and $L=48$ runs reach $7.62$ and $7.61$. \gct{} reaches $3.973$ and $3.903$ under the same fixed-budget depth protocol (Table~\ref{tab:depth}). These values illustrate why the coefficient-one anchor requires a different depth schedule or training recipe.

Our experiments use the coefficient-one Post-LN anchor and do not evaluate the full DeepNet recipe. The theoretical comparison tracks the measured residual-to-stream RMS ratio, which depends on both the raw coefficients and the branch and stream scales.

\begin{figure}[t!]
\centering
\includegraphics[width=\textwidth]{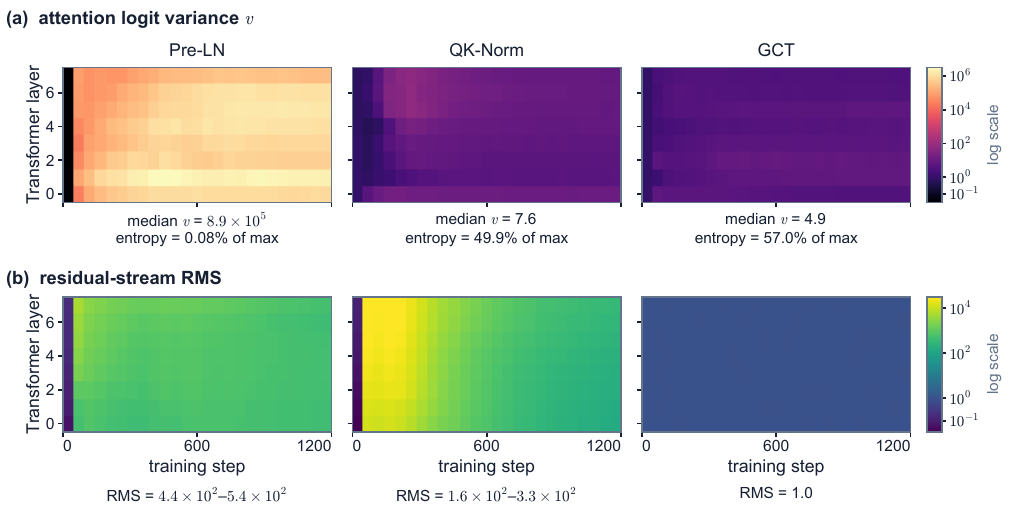}
\caption{UVI fingerprint for matched 51M stress runs at seed 1337 and 600M tokens, without warmup or clipping, at peak learning rate $9.6\times10^{-3}$. (a) QK-Norm repairs attention variance and entropy, while (b) QK-Norm's residual RMS still reaches values between $1.6\times10^2$ and $3.3\times10^2$. \gct{} keeps both readouts controlled. The annotations report final-step statistics: the layer median of $v$, the mean attention entropy as a fraction of the maximum attention entropy, and the across-layer range of the stream RMS. This figure is a single-seed localization case, while Figure~\ref{fig:lrsweep} and Table~\ref{tab:sweep} provide the sweep evidence.}
\label{fig:qual-uvi-fingerprint}
\end{figure}

\paragraph{UVI fingerprint.} Figure~\ref{fig:qual-uvi-fingerprint} illustrates how the diagnostics localize scale mismatches in matched stress runs. QK-Norm controls attention variance and preserves attention entropy, while QK-Norm's residual-stream RMS grows by orders of magnitude. \gct{} controls both readouts. Residual-stream RMS is an auxiliary drift measurement, distinct from $\UVI_{\mathrm{res}}$. Together with the attention diagnostic, the RMS measurement distinguishes local attention control from control of the whole stream.

\begin{table}[t!]
\centering\small
\caption{Sweep-to-scale transfer. Each architecture selects a learning rate from the original five-point $16\times$ grid in Table~\ref{tab:sweep} and applies the selected learning rate unchanged at 124M. The $9.6\times10^{-3}$ stress point is excluded. The selected-rate \gct{} entry reports mean $\pm$ standard deviation over two seeds, while other loss entries use the primary seed. $\Delta$ is computed from unrounded values relative to the primary-seed result at $6\times10^{-4}$.}
\label{tab:transfer}
\begin{tabular}{lcccc}
\toprule
 & & \multicolumn{2}{c}{124M val loss} & \\
\cmidrule(lr){3-4}
architecture & 51M-selected rate & at $6\times10^{-4}$ & at selected rate & $\Delta$\\
\midrule
Pre-LN & \multirow{3}{*}{$1.2\times10^{-3}$} & 3.172 & 3.119 & $-0.052$ \\
LLaMA-style & & 3.178 & 3.105 & $-0.072$ \\
QK-Norm & & 3.149 & 3.091 & $-0.058$ \\
\midrule
nGPT & \multirow{2}{*}{$4.8\times10^{-3}$} & 3.171 & 3.043 & $-0.127$ \\
\gct{} (ours) & & 3.170 & 3.045 $\pm$ 0.002 & $-0.125$ \\
\bottomrule
\end{tabular}
\end{table}

\paragraph{Interpreting learning-rate transfer.} The sweep is consistent with the scale-management interpretation in \secref{sec:exp-lr}. Canonicalization controls residual-stream magnitude, making the forward computation less sensitive to one source of variation across depth and training time. Both \gct{} and nGPT retain low validation loss across a wide range of high learning rates (Table~\ref{tab:sweep}). The transfer experiments test whether a rate selected from this range remains effective at a larger model size.

\paragraph{Sweep-to-scale transfer.} We select a learning rate for each architecture using the original five-point 51M sweep and apply the selected learning rate unchanged at 124M (Table~\ref{tab:transfer}). Under this protocol, the transferred learning rate \lrtransferclaim{}.

\subsection{Additional rescaling and weight-space audits}
\label{app:rescalingaudits}

\begin{figure}[t!]
\centering
\includegraphics[width=\textwidth]{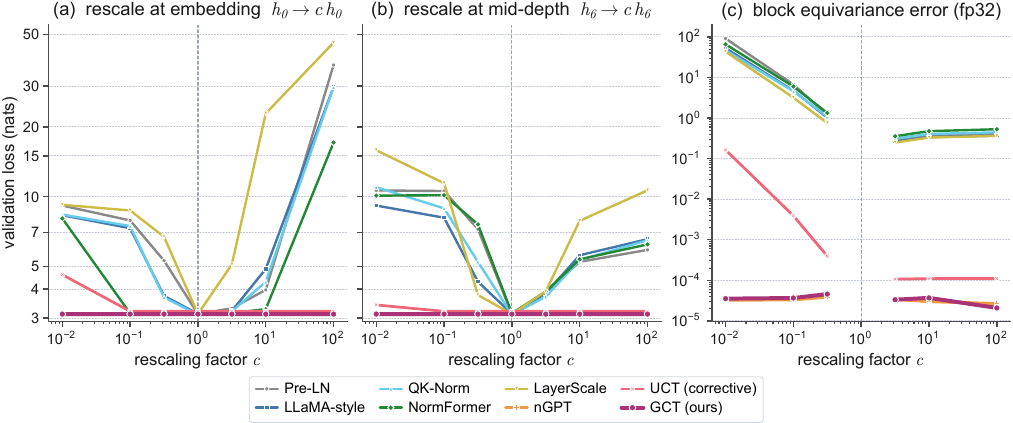}
\caption{Validation loss under internal rescaling $h\to ch$ at the embedding output and mid-depth, with fp32 block error. The intervention changes the hidden state's magnitude while preserving the state's direction. \gct{} and nGPT preserve validation loss, with block invariance errors near the numerical floor, showing insensitivity to the imposed magnitude change. UCT loses invariance at $c=10^{-2}$ because the hidden state shrinks while the small constants in UCT's normalization denominators and the stored evaluation-time scale estimates remain fixed. The fixed constants account for a larger share of the normalization denominators, and the stored scales no longer match the activations. These mismatches prevent exact cancellation of the rescaling. Baselines with unnormalized residual streams also degrade.}
\label{fig:perturb}
\end{figure}

\begin{table}[t!]
\centering\small
\caption{Validation loss of the trained 124M models at the $c=0.1$ and $c=10$ points of Figure~\ref{fig:perturb}, together with block equivariance error at $c=10$.}
\label{tab:perturb}
\begin{tabular}{lcccccc}
\toprule
 & & \multicolumn{2}{c}{rescaled at embedding} & \multicolumn{2}{c}{rescaled at mid-depth} & \\
\cmidrule(lr){3-4}\cmidrule(lr){5-6}
architecture & base & $c=0.1$ & $c=10$ & $c=0.1$ & $c=10$ & eq error\\
\midrule
Pre-LN & 3.118 & 7.901 & 3.964 & 10.581 & 5.239 & $3.7\times10^{-1}$ \\
LLaMA-style & 3.126 & 7.317 & 4.884 & 8.118 & 5.592 & $3.4\times10^{-1}$ \\
QK-Norm & 3.098 & 7.474 & 4.299 & 8.909 & 5.333 & $4.0\times10^{-1}$ \\
NormFormer & 3.139 & 3.141 & 3.277 & 10.153 & 5.370 & $4.7\times10^{-1}$ \\
LayerScale & 3.120 & 8.723 & 22.860 & 11.438 & 7.875 & $3.3\times10^{-1}$ \\
\midrule
nGPT & 3.119 & 3.119 & 3.119 & 3.119 & 3.119 & $3.0\times10^{-5}$ \\
Corrective UCT (ours) & 3.204 & 3.205 & 3.204 & 3.204 & 3.204 & $1.1\times10^{-4}$ \\
\gct{} (ours) & 3.118 & 3.118 & 3.118 & 3.118 & 3.118 & $3.7\times10^{-5}$ \\
\bottomrule
\end{tabular}
\end{table}

The corrective UCT is nearly flat at the $c\in\{0.1,10\}$ points of Table~\ref{tab:perturb}. At the stronger down-scaling $c=10^{-2}$ in Figure~\ref{fig:perturb}, UCT's loss changes by \perturbuctembdown{} nats at the embedding intervention and \perturbuctmiddown{} nats at mid-depth, with block equivariance error \perturbucteqdown{}. At this smaller activation scale, the fixed constants account for a larger share of the normalization denominators, while the stored evaluation scales do not adjust to the smaller activations. UCT's local corrections no longer cancel the injected rescaling.

The baselines without normalized residual streams are more sensitive to the injected rescaling. The largest baseline increase is \perturbworst{} among the tabulated points and \perturbfullworst{} over the full sweep in Figure~\ref{fig:perturb}.\footnote{Table~\ref{tab:perturb} uses the same fixed 32-batch evaluation subset as the quantization study. The base losses in the perturbation table coincide with the base column of Table~\ref{tab:quant}. These subset losses can differ from full-validation losses, but each reported change is computed within the same subset.}

\paragraph{The weight-space shadow.} A bias-free matrix product $x\coloneqq Wu$ is $1$-homogeneous in $W$. When this product is the block's only dependence on $W$, with no tied use or additive $W$-dependent path, the ray $W\mapsto e^tW$ realizes the activation rescaling $x\mapsto e^tx$. If the downstream graph nondimensionalizes that activation, the same ray becomes an exact symmetry of the loss.

\begin{proposition}[Weight-space shadow of activation typing]
\label{prop:pgauge}
Let $W$ be a parameter block whose only effect on the loss is through $x\coloneqq Wu$, where $u$ is independent of $W$. Suppose the graph nondimensionalizes $x$ such that the downstream computation is invariant to $x\mapsto cx$ for $c>0$ when all other quantities are fixed. Then $\Lcal(e^tW)=\Lcal(W)$ for all $t\in\mathbb R$, and the radial Euler pairing vanishes exactly wherever the derivative is defined:
\begin{align*}
\langle W,\nabla_W\Lcal\rangle_F=0.
\end{align*}
The proof is given in App.~\ref{app:pgproof}, and the architecture-specific predictions and measurement protocol are given in App.~\ref{app:auditprotocols}.
\end{proposition}

If instead the graph fails to nondimensionalize $x$, the weight-space radial pairing equals the corresponding activation-space pairing $\langle x,\nabla_x\Lcal\rangle$ and is generally nonzero. App.~\ref{app:pgproof} gives the derivation.

\paragraph{Measured fingerprint.} We also identify weight matrices whose magnitude can change without changing the loss. Prop.~\ref{prop:pgauge} predicts zero radial sensitivity $\langle W,\nabla_W\Lcal\rangle_F$ when the downstream computation cancels rescaling of a matrix's output. We measure this pairing to check whether multiplying the matrix by a common factor affects the loss. Under $W\mapsto e^tW$, this pairing is the loss derivative at $t=0$, measured in nats per $e$-fold. One $e$-fold is a factor of $e\approx2.718$ in weight magnitude, corresponding to a one-unit increase in log-scale. Table~\ref{tab:paramgauge} reports the predicted zero and nonzero roles and the largest measured sensitivity across layers for each architecture at 124M. All \pgmatch{} architecture-by-role predictions match across the eight trained models, with predicted-null cells at the fp32 floor of at most \pgtypedmax{} nats per $e$-fold. The 1B audit likewise matches \pgmatchoneb{} cells, with predicted-null values at most \pgtypedmaxoneb{} nats per $e$-fold. This agreement shows that NDA predicts which weight matrices can be rescaled without changing the loss. For example, QK-Norm cancels rescaling of the query and key projections but retains sensitivity to the other probed matrices.

\secref{sec:exp-pgauge} reports the aggregate fingerprint. Table~\ref{tab:paramgauge} gives the full rolewise audit, reporting the worst layer in every architecture-by-role cell. \gct{} and nGPT are null on every role. QK-Norm is null exactly on $q$ and $k$ and nowhere else. NormFormer's post-attention LayerNorm types NormFormer's attention write through $v$ and the output projection, but no other role. Pre-LN, LLaMA-style, and LayerScale have no null direction, with pairings from \pgminnonnull{} to \pgworst{} nats per $e$-fold.

The local-correction UCT is null where UCT's live-RMS adapter acts, but the $q,k$ and gate interfaces corrected with frozen EMA statistics lose nullity at deployment. When batch statistics are active, the finite-difference slope vanishes at \pguctfd{} while the stop-gradient autograd pairing remains \pguctpair{}. Here local correction restores forward invariance with live batch statistics, while the stop-gradient training computation and the frozen deployment graph retain radial sensitivity.

\begin{table}[t!]
\centering\small
\caption{The Prop.~\ref{prop:pgauge} fingerprint on the trained 124M models in fp32 evaluation mode. Entries report $\max_{\ell}|\langle W_{\ell},\nabla_{W_{\ell}}\Lcal\rangle|$ in nats per $e$-fold for each weight-matrix role. The symbol $\dagger$ marks a cell predicted to be an exact null by the architecture's typing, with the proof in App.~\ref{app:pgproof} and the rolewise predictions in App.~\ref{app:auditprotocols}. Every marked cell sits at the fp32 floor, while every other cell is orders of magnitude above that floor. Equal-pairing roles agree as the theory requires for $q$ and $k$, for $v$ and the attention output, and for the up-projection and MLP output. A double dash denotes an absent role because a GELU MLP has no separate up-projection.}
\label{tab:paramgauge}
\resizebox{\linewidth}{!}{\begin{tabular}{lccccccc}
\toprule
 & \multicolumn{4}{c}{attention} & \multicolumn{3}{c}{MLP}\\
\cmidrule(lr){2-5}\cmidrule(lr){6-8}
architecture & $q$ & $k$ & $v$ & out & in & up & out\\
\midrule
Pre-LN & $4.1\times10^{-3}$ & $4.1\times10^{-3}$ & $3.5\times10^{-3}$ & $3.5\times10^{-3}$ & $2.6\times10^{-2}$ & -- & $1.6\times10^{-2}$ \\
LLaMA-style & $4.2\times10^{-3}$ & $4.2\times10^{-3}$ & $1.2\times10^{-2}$ & $1.2\times10^{-2}$ & $2.3\times10^{-2}$ & $1.6\times10^{-2}$ & $1.6\times10^{-2}$ \\
QK-Norm & $6.4\times10^{-9}\dagger$ & $1.5\times10^{-8}\dagger$ & $4.6\times10^{-3}$ & $4.6\times10^{-3}$ & $2.1\times10^{-2}$ & $1.3\times10^{-2}$ & $1.3\times10^{-2}$ \\
NormFormer & $4.3\times10^{-3}$ & $4.3\times10^{-3}$ & $2.4\times10^{-7}\dagger$ & $2.4\times10^{-7}\dagger$ & $9.2\times10^{-3}$ & -- & $1.3\times10^{-2}$ \\
LayerScale & $4.5\times10^{-3}$ & $4.5\times10^{-3}$ & $6.7\times10^{-3}$ & $6.7\times10^{-3}$ & $2.0\times10^{-2}$ & $1.1\times10^{-2}$ & $1.1\times10^{-2}$ \\
\midrule
nGPT & $2.9\times10^{-10}\dagger$ & $3.9\times10^{-10}\dagger$ & $5.0\times10^{-10}\dagger$ & $1.0\times10^{-9}\dagger$ & $3.5\times10^{-10}\dagger$ & $3.7\times10^{-10}\dagger$ & $8.4\times10^{-10}\dagger$ \\
Corrective UCT (ours) & $1.3\times10^{-2}$ & $1.3\times10^{-2}$ & $4.3\times10^{-8}\dagger$ & $4.4\times10^{-8}\dagger$ & $1.9\times10^{-2}$ & $3.0\times10^{-8}\dagger$ & $2.7\times10^{-8}\dagger$ \\
\gct{} (ours) & $1.1\times10^{-9}\dagger$ & $1.9\times10^{-9}\dagger$ & $3.7\times10^{-9}\dagger$ & $1.1\times10^{-9}\dagger$ & $2.0\times10^{-9}\dagger$ & $3.7\times10^{-9}\dagger$ & $2.5\times10^{-9}\dagger$ \\
\bottomrule
\end{tabular}
}
\end{table}

The following PTQ evaluations test how the same architectures respond to rounding errors during finite-precision storage and computation.

\subsection{Extended PTQ results}
\label{app:quantresults}

Integer quantization stores integer coordinates together with a scale that restores the activation magnitudes. A shared tensor scale must cover large activation values, making the integer grid relatively coarse for tokens with smaller magnitudes. Per-token scaling assigns a separate range of representable values to each token. Tokens with smaller activations can then use more closely spaced values instead of the coarse spacing set by the largest activations in the tensor.

\begin{table}[t!]
\centering\small
\caption{PTQ of the trained 124M models. Values are changes in validation loss relative to bf16, measured in nats, and positive values are worse. The 8-bit weight-only setting is not tabulated because that setting changes loss by $0.000$ nats for every architecture, both here and at 1B. W8A8 uses static absmax calibration at every linear input. The $h\to\mathrm{int8}$ and $h\to\mathrm{int4}$ columns quantize the residual stream at every block boundary with per-layer static tensor scales or per-token scales. The logit head remains in full precision.}
\label{tab:quant}
\begin{tabular}{lccccccc}
\toprule
 & & \multicolumn{2}{c}{weights and activations} & \multicolumn{2}{c}{stream $h\to\mathrm{int8}$} & \multicolumn{2}{c}{stream $h\to\mathrm{int4}$}\\
\cmidrule(lr){3-4}\cmidrule(lr){5-6}\cmidrule(lr){7-8}
architecture & base & W4 & W8A8 & tensor & token & tensor & token\\
\midrule
Pre-LN & 3.118 & +0.067 & +0.016 & +0.085 & +0.002 & +6.723 & +0.912 \\
LLaMA-style & 3.126 & +0.063 & +0.270 & +1.058 & +0.001 & +6.212 & +1.044 \\
QK-Norm & 3.098 & +0.061 & +0.237 & +1.099 & +0.001 & +7.715 & +0.386 \\
NormFormer & 3.139 & +0.125 & +0.234 & +0.034 & +0.001 & +4.164 & +0.761 \\
LayerScale & 3.120 & +0.061 & +0.178 & +0.896 & +0.002 & +6.214 & +0.691 \\
\midrule
nGPT & 3.119 & +0.070 & +0.024 & +0.005 & +0.001 & +2.951 & +0.522 \\
Corrective UCT (ours) & 3.204 & +0.104 & +0.056 & +0.007 & +0.000 & +2.033 & +0.152 \\
\gct{} (ours) & 3.118 & +0.056 & +0.026 & +0.001 & +0.000 & +0.828 & +0.107 \\
\bottomrule
\end{tabular}
\end{table}

\begin{figure}[t!]
\centering
\includegraphics[width=\linewidth]{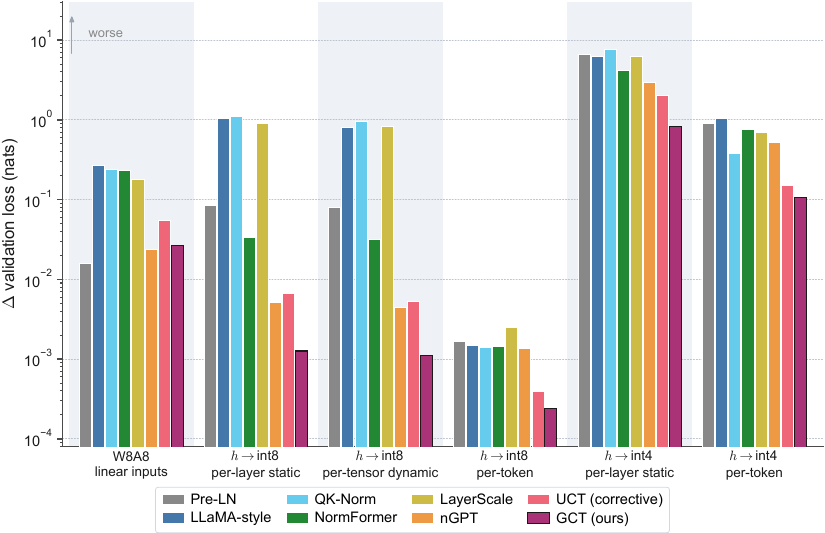}
\caption{Quantization degradation by scheme on a logarithmic scale. Per-tensor int8 on the residual stream separates the architectures by orders of magnitude. Per-token scales remove token-to-token scale spread from the quantizer's relative resolution and substantially reduce the int8 gap.}
\label{fig:quant}
\end{figure}

Quantizing the same trained functions in Table~\ref{tab:quant} and Figure~\ref{fig:quant} differentiates architectures along the scale-management axis predicted by the typing rules. Static per-site, per-tensor int8 changes the normalized-gauge models by at most $+0.005$ nats, while the other 124M baselines change by $+0.034$ to $+1.099$ nats. Per-token scales reduce the worst 124M change to $+0.002$ nats. At 1B under per-token int4, \gct{} has the smallest increase, at $+0.739$ nats.

The control schemes distinguish sensitivity to the quantization site and scale granularity. Weight-only int8 changes loss by $0.000$ nats at the reported precision at both model scales, whereas weight-only int4 increases loss by at most $0.125$ nats. Static W8A8 produces larger, architecture-dependent changes, including $+0.833$ nats for QK-Norm at 1B (Table~\ref{tab:quant1b}). The KV-cache control below has much smaller changes across all architectures. The strongest separation between normalized streams and the other baselines occurs under per-tensor residual-stream int8. Per-token scales substantially reduce this separation, supporting cross-token scale variation as an important contributor.

\paragraph{Kernel fidelity.} Simulated quantization followed by dequantization and a floating-point general matrix multiplication (GEMM) could diverge from deployed integer arithmetic. To check fidelity, we reran the W8A8 setting of Table~\ref{tab:quant} through the actual int8 tensor-core path, \texttt{torch.\_int\_mm}, with int8$\times$int8 products accumulated in 32-bit integers and identical scale computation. Every architecture's loss delta agrees with the simulated delta for the same architecture to within \ieightgapmax{} nats across all \ieightnarch{} trained 124M models, while the deltas span up to \ieightdeltamax{} nats. This agreement supports the fidelity of the simulated W8A8 evaluation. The residual-stream protocols evaluate quantization and dequantization without replacing matrix multiplication.

\paragraph{Cost of fine-grained scale management.} On the same device, the per-token quantization and dequantization step takes \ptcostratio{} as long as the per-tensor step and \ptcostfrac{} as long as the feed-forward GEMM surrounded by the quantization and dequantization operations. Scale metadata for cached or offloaded activations grows from 48~B for the per-tensor path to \ptcostmetakb{}~KB per 1024 tokens for a 24-layer stream. A normalized-gauge stream still incurs the per-tensor round trip but avoids this additional fine-grained management cost.

\paragraph{The migration alternative.} Scale migration provides another way to manage activation ranges, as reviewed in App.~\ref{app:related}. SmoothQuant moves calibrated per-channel activation scales into adjacent weights offline \citep{xiao2023smoothquant}, while calibrated per-channel scales provide analogous range control for a stored stream. Rotation-based schemes redistribute coordinate magnitudes by changing the basis \citep{ashkboos2024quarot}. All of these methods can be applied to trained models.

Migration reduces the measured quantization error on the same trained models. SmoothQuant with $\alpha=0.5$ and per-site static calibration reduces the worst-baseline W8A8 envelope of Table~\ref{tab:quant} from \sqwstatworst{} to \sqwsmoothworst{} nats, with \sqwsmoothworstname{} worst after migration. Per-channel static scales reduce the worst-baseline residual-stream envelope from \quantworsthstat{} to \sqhchanworst{} nats, with \sqhchanworstname{} worst after migration. At 1B the worst value is \sqhchanworstoneb{}, compared with \quantworsthstatoneb{} before this treatment. For \gct{}, the two treatments produce loss changes of \sqgctwsmooth{} and \sqgcthchan{} nats, respectively. At 1B, the stream treatment changes \gct{}'s loss by \sqgcthchanoneb{} nats.

The serving stack can manage scale externally, or the architecture can control scale before deployment. Static migration and channel scales require a calibration pass for each deployment target, using 8 held-out batches here, and may require recalibration under distribution drift. The dynamic per-token route carries fine-grained metadata on every cache and offload path, with the \ptcostratio{} runtime ratio and \ptcostmetakb{}~KB per 1024 tokens measured above. Both routes require kernel support for the chosen granularity. The normalized-gauge architectures meet the int8 loss threshold with per-tensor scales in this setting, while retaining the ordinary quantization and dequantization cost.

The KV cache gives a negative control. KV-cache compression is a separate deployment target \citep{liu2024kivi}. Every architecture computes keys and values from normalized block inputs, removing accumulated radial stream drift before the projections. Equal projected key and value norms require additional conditions because a learned projection can still produce direction-dependent norms. We measure $\kappa_{\mathrm{scale}}\approx1$ empirically.

Consistent with that observation, int8 KV-cache quantization is near-lossless for all eight architectures. The worst increases are $+\kvworst{}$ nats with per-tensor scales and $+\kvworsttok{}$ with per-token scales, with no separation between normalized-gauge and baseline blocks. \kvonebsentence{} These controls localize the measured deployment gap of Table~\ref{tab:quant} to residual-stream storage.

\paragraph{Persistence at scale.} We repeat the study on the trained 1B models (Table~\ref{tab:quant1b}). The separation persists over the tested $8\times$ parameter-count increase, with \gct{} at \quantgcthstatoneb{} nats compared with \quantworsthstatoneb{} nats for \quantworsthstatonebname{} under static per-tensor stream int8.

\begin{figure}[t!]
\centering
\includegraphics[width=0.62\linewidth]{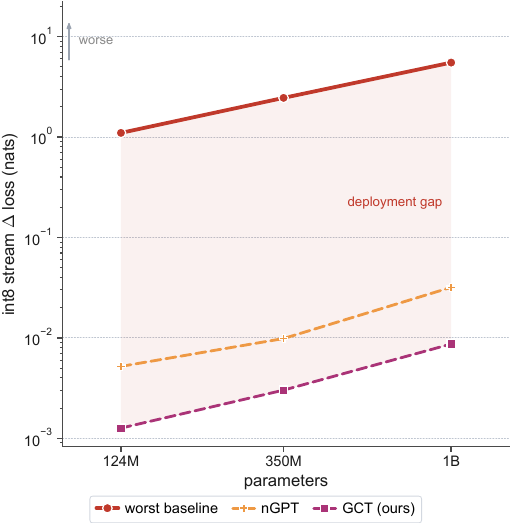}
\caption{Table~\ref{tab:quantscale} on log--log axes. Along the tested model-size ladder, the worst-baseline increase reaches $+\quantworsthstatoneb{}$ nats at 1B, while nGPT and \gct{} remain near the int8 floor.}
\label{fig:quantscale}
\end{figure}

Table~\ref{tab:quantscale} and Figure~\ref{fig:quantscale} add the 350M point between 124M and 1B. \quantscaletrend{}. Across these model sizes, the largest baseline loss increase grows with parameter count, while nGPT and \gct{} retain small int8 loss increases.

The int4 results also reflect factors beyond cross-token scale spread. In the zero-floor theory, canonicalization fixes \gct{}'s cross-token RMS spread at $\kappa_{\mathrm{scale}}\equiv1$, and the finite-floor implementation fixes this spread to numerical precision. The remaining error can depend on within-vector amplitude $\kappa_{\mathrm{amp}}$, which the canonical gauge leaves untouched, together with downstream directional sensitivity and accumulation across sites, neither of which the proposition quantifies. Measured per-tensor int4 losses increase substantially for every stream. Under per-token int4, \gct{} has the smallest measured increase, at $+0.739$ nats (Table~\ref{tab:quant1b}). Explaining this empirical ordering requires analysis of the additional factors beyond cross-token scale spread. On the independent second 1B seed, the parameter-fingerprint audit again matches all \pgmatchonebstwo{} predictions; the corresponding PTQ repeat is reported in \secref{sec:exp-quant}.

Exact internal-rescaling invariance of the zero-floor construction removes the radial component of a scale distortion. Robustness to quantization additionally depends on the tangential and within-vector terms quantified by Thm.~\ref{thm:tangent} and Prop.~\ref{prop:quant}. The deployment tables measure the combined effects of radial rejection, tangential perturbations, and within-vector amplitude.

\paragraph{Residual-stream storage at fixed loss thresholds.} The residual-stream cache occupies \servingbfgib{}~GiB in bf16 for an illustrative 1B configuration with 32 concurrent sequences and context length 2048. At a $0.05$-nat threshold, nGPT and \gct{} use calibrated per-site int8 tensor scales and occupy \servingeightgib{}~GiB, whereas the other baselines require per-token int8 scales. At a $1$-nat threshold, \gct{} reaches $\servinggctcomp{}\times$ compression at \servingfourgib{}~GiB with a \servinggctdelta{}-nat increase, whereas the other evaluated architectures exceed that threshold under int4. These precision choices use the measured loss increases in Table~\ref{tab:quant1b}.

These storage figures include only the residual-stream cache and allow per-token scaling for every architecture. Weight and KV-cache storage are excluded. KV-cache quantization is evaluated separately above. The $4\times$ result uses per-token int4 and has the measured \servinggctdelta{}-nat loss increase. The option to use per-tensor scales applies to the $2\times$ int8 configuration.

\paragraph{Measured latency.} We time the dynamic-absmax residual-stream storage-and-retrieval round trip on device during a forward pass of the independent second-seed 1B model, with $d=1792$, $L=24$, $B=\servingbenchbatch{}$ sequences, and context length $T=\servingbenchctx{}$. The preceding accuracy budget uses the calibrated static primary-seed column of Table~\ref{tab:quant1b}. Separately measured dynamic per-tensor deltas on both 1B seeds induce the same $0.05$-nat tier assignment. \gct{} and nGPT meet this threshold, whereas the other baselines exceed the loss budget.

At this near-lossless granularity, the per-tensor paths add $6.2\%$ and $7.4\%$ forward latency for \gct{} and nGPT, respectively. The per-token paths required by Pre-LN, LLaMA-style, and QK-Norm add $12.3\%$, $12.9\%$, and $9.9\%$. Every path reduces the bf16 stream cache from \servingbfgib{} to \servingeightgib{}~GiB. Across these configurations, the measured latency overhead ranges from $6\%$ to $13\%$.

At the stated loss threshold, the normalized-gauge streams need only coarse per-tensor scales, whereas the other baselines require per-token scales. The primary-seed int4 per-token deltas are $+0.74$ nats for \gct{}, $+4.66$ for nGPT, and $+5.23$ to $+7.10$ for the other baselines. Timing and accuracy use different protocols because static calibration and dynamic absmax are different operations.

\begin{table}[t!]
\centering\small
\caption{Downstream accuracy under residual-stream quantization on the 1B models trained independently with a second seed. We report zero-shot accuracy with the harness formatting for \quantdstasks{}. The stream is stored at the same block-boundary sites and granularities as in Table~\ref{tab:quant1b}, using dynamic scales without calibration, and the evaluation is inference-only, with no retraining or fine-tuning. The bf16 column gives mean accuracy over the four tasks in percent, and the remaining columns give accuracy changes in points, where negative values indicate lower accuracy.}
\label{tab:quantds}
\begin{tabular}{lcccc}
\toprule
& & \multicolumn{2}{c}{int8} & int4\\
\cmidrule(lr){3-4}\cmidrule(lr){5-5}
architecture & bf16 (\%) & tensor & token & token\\
\midrule
Pre-LN & 48.6 & $-20.5$ & $-1.9$ & $-22.7$ \\
LLaMA-style & 48.7 & $-22.7$ & $-0.8$ & $-22.6$ \\
QK-Norm & 49.4 & $-19.6$ & $-0.5$ & $-23.1$ \\
\midrule
nGPT & 49.0 & $-0.8$ & $-0.3$ & $-17.4$ \\
\gct{} (ours) & 49.0 & $-0.1$ & $+0.2$ & $-7.4$ \\
\bottomrule
\end{tabular}
\end{table}

Table~\ref{tab:quantds} gives the full architecture comparison for the downstream accuracy results reported in \secref{sec:exp-quant}.

The publicly available pretrained GPT-2 models show a similar dependence on scale granularity (Table~\ref{tab:offtheshelf}). Across the four sizes, $\kappa_{\mathrm{depth}}$ ranges from \otskappamin{} to \otskappamax{}. Dynamic per-tensor int8 residual storage increases loss by $5.79$ to $9.03$ nats, while per-token scaling limits the increase to $0.046$ nats. This audit extends the observed quantization pattern to independently trained models. The depth statistic is descriptive and differs from the within-tensor spread used in the rounding bound.

\begin{table}[t!]
\centering\small
\caption{External-validity audit on off-the-shelf GPT-2. The OpenAI GPT-2 family uses Pre-LN and was trained by a third party on WebText. We evaluate the family on the same validation bins and residual-stream fake-quantization hooks as our trained models, using dynamic scales, with loss increases reported in nats relative to each unquantized model. The quantity $\kappa_{\mathrm{depth}}$ is the ratio between the maximum and minimum mean residual-stream RMS across block inputs, not the within-tensor spread of Prop.~\ref{prop:quant}. Every size has a large per-tensor loss increase and a near-lossless per-token result. For comparison, our 124M \gct{} model has $\kappa_{\mathrm{depth}}\approx1$ and a $+$\otsowngctdyn{}-nat dynamic-int8 loss increase.}
\label{tab:offtheshelf}
\begin{tabular}{lcccc}
\toprule
& & \multicolumn{2}{c}{int8} & int4 \\
\cmidrule(lr){3-4}\cmidrule(lr){5-5}
model & $\kappa_{\mathrm{depth}}$ & tensor & token & token \\
\midrule
GPT-2 (124M) & $48$ & $+5.79$ & $+0.008$ & $+4.25$ \\
GPT-2 (355M) & $236$ & $+7.91$ & $+0.046$ & $+8.02$ \\
GPT-2 (774M) & $223$ & $+9.03$ & $+0.034$ & $+6.02$ \\
GPT-2 (1.5B) & $596$ & $+7.30$ & $+0.035$ & $+8.05$ \\
\bottomrule
\end{tabular}
\end{table}

\begin{table}[t!]
\centering\small
\caption{Calibrated per-site, per-tensor int8 residual-stream quantization along the 124M, 350M, and 1B ladder. Columns report one trained model per architecture, the primary seed at 124M and 1B and the second seed at 350M. nGPT and \gct{} remain near the int8 floor. The largest baseline loss increase grows with parameter count, although individual trajectories need not be monotone.}
\label{tab:quantscale}
\begin{tabular}{lccc}
\toprule
& \multicolumn{3}{c}{model size}\\
\cmidrule(lr){2-4}
architecture & 124M & 350M & 1.0B \\
\midrule
Pre-LN & $+0.085$ & $+1.009$ & $+3.018$ \\
LLaMA-style & $+1.058$ & $+2.453$ & $+5.513$ \\
QK-Norm & $+1.099$ & $+0.325$ & $+5.426$ \\
\midrule
nGPT & $+0.005$ & $+0.010$ & $+0.032$ \\
\gct{} (ours) & $+0.001$ & $+0.003$ & $+0.009$ \\
\bottomrule
\end{tabular}
\end{table}

\subsection{Additional optimizer and width evaluations}
\label{sec:exp-breadth}

Two checks test whether the comparison generalizes beyond one optimizer and one width. Table~\ref{tab:widthmup} gives the width results.

\paragraph{Optimizer pairing.} The optimizer rule in \secref{sec:rules} requires the full displacement to transform from dual to primal units. Muon supplies a competitive shape-aware orthogonalized direction with a fixed scalar step. The experiment tests compatibility rather than the covariance law. \muonclaim{}. Relative to the shared-recipe AdamW point at matched tokens, the changes are $\muongctdelta{}$ nats for \gct{}, $\muonprelndelta{}$ for Pre-LN, and $\muonqknormdelta{}$ for QK-Norm. The result shows that the typed architecture and this shape-aware optimizer compose without changing the architecture.

\paragraph{Width transfer and $\mu$P composition.} Two experiments probe the parameterization axis in Table~\ref{tab:widthmup}. The learning-rate robustness protocol of Table~\ref{tab:sweep} is repeated at $d\in\{256,1024\}$ with $d_h=64$ and the rest of the recipe unchanged. \widthplateauclaim{}. We then compose both architectures with the learning-rate-only $\mu$P rule at the respective $d=512$-tuned learning rates. This rule scales hidden-matrix learning rates by $512/d$ and leaves initialization and multipliers unchanged. \mupwidthclaim{}. These results support combining the tested width-dependent learning-rate rule with canonical activation scaling.

\begin{table}[t!]
\centering\small
\caption{Width transfer and learning-rate-only $\mu$P composition for the 51M-centered family, with $L=8$, $d_h=64$, and the primary seed under the Table~\ref{tab:sweep} token budget. The standard-parameterization (SP) columns use the listed learning rate. The $\mu$P columns scale hidden-matrix learning rates by $512/d$ from the architecture's $d=512$-tuned rate, namely $4.8\times10^{-3}$ for \gct{} and $1.2\times10^{-3}$ for Pre-LN. At the tuning width, the rule reduces to SP.}
\label{tab:widthmup}
\begin{tabular}{lccccccc}
\toprule
 & \multicolumn{3}{c}{\gct{}} & \multicolumn{4}{c}{Pre-LN}\\
\cmidrule(lr){2-4}\cmidrule(lr){5-8}
width & $6\times10^{-4}$ & $4.8\times10^{-3}$ & $\mu$P & $6\times10^{-4}$ & $1.2\times10^{-3}$ & $4.8\times10^{-3}$ & $\mu$P\\
\midrule
$d=256$ & 4.366 & 3.773 & 3.768 & 4.419 & 4.166 & 4.307 & 4.186 \\
$d=512$ & 3.885 & 3.517 & 3.517 & 3.943 & 3.750 & 3.881 & 3.750 \\
$d=1024$ & 3.590 & 3.345 & 3.351 & 3.613 & 3.617 & 5.652 & 3.549 \\
\bottomrule
\end{tabular}
\end{table}

\subsection{Vision: robustness to photometric and input-unit shifts}
\label{sec:exp-vit}

\begin{table}[t!]
\centering\small
\caption{CIFAR-100 from scratch with the DeiT-Ti recipe for 300 epochs. Entries report top-1 accuracy in percent. Robustness columns average over shift grids with noise $\sigma\in\{0.05,0.1\}$, contrast $\{0.5,2.0\}$, brightness $\{0.7,1.3\}$, and input scale $\{0.25,0.5,2,4\}$.}
\label{tab:vit}
\begin{tabular}{lccccc}
\toprule
 & & \multicolumn{4}{c}{robustness}\\
\cmidrule(lr){3-6}
architecture & clean & noise & contrast & bright & in-scale\\
\midrule
Pre-LN & 69.81 & 37.49 & 63.27 & 68.45 & 51.58 \\
QK-Norm & 70.68 & 36.92 & 64.80 & 69.58 & 56.00 \\
\midrule
nGPT & 73.98 & 34.34 & 68.92 & 73.00 & 63.54 \\
Corrective UCT (ours) & 73.28 & 33.66 & 68.00 & 72.25 & 62.25 \\
\gct{} (ours) & 74.17 & 34.36 & 69.23 & 72.93 & 64.13 \\
\bottomrule
\end{tabular}
\end{table}

\begin{figure}[t!]
\centering
\includegraphics[width=0.70\linewidth]{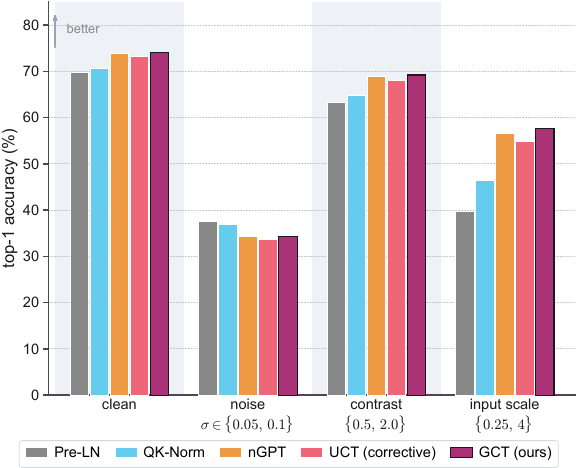}
\caption{ViT robustness grid on CIFAR-100 accompanying Table~\ref{tab:vit}. The figure reports top-1 accuracy under the photometric and input-scale shift grid.}
\label{fig:vitrobust}
\end{figure}

\gct{} \vitclaim{} (Table~\ref{tab:vit}). All models are trained to convergence with the standard DeiT-Ti recipe. Robustness is then compared on contrast, brightness, and noise shifts, with the full accuracy grid in Figure~\ref{fig:vitrobust}. The sharpest comparison is multiplicative rescaling of the normalized input, which is the vision analogue of an input-unit shift.

\paragraph{Additive-noise robustness.} \gct{} scores \vitnoisegct{} on the averaged Gaussian-noise shift, against \vitnoisebest{} for the best baseline in the noise column of Table~\ref{tab:vit}. One possible explanation is that canonicalization removes magnitude information that can help distinguish signal from additive noise. The CIFAR-100 results are consistent with this explanation, although the causal role of magnitude removal remains to be tested.

\begin{table}[t!]
\centering\small
\caption{Tiny-ImageNet-200 from scratch with ViT-S, patch size 8, the DeiT recipe, and 200 epochs. Accuracy is reported as in Table~\ref{tab:vit}, and the noise, contrast, and input-scale columns average over the corresponding shift grids.}
\label{tab:vittin}
\begin{tabular}{lcccc}
\toprule
 & & \multicolumn{3}{c}{robustness}\\
\cmidrule(lr){3-5}
architecture & clean & noise & contrast & in-scale\\
\midrule
Pre-LN & 54.94 & 44.67 & 48.53 & 37.32 \\
LLaMA-style & 55.14 & 45.11 & 49.66 & 40.91 \\
QK-Norm & 56.30 & 46.24 & 51.30 & 43.25 \\
NormFormer & 43.18 & 34.49 & 35.52 & 21.27 \\
LayerScale & 56.38 & 45.48 & 50.87 & 39.73 \\
\midrule
nGPT & 57.81 & 48.18 & 53.31 & 47.15 \\
Corrective UCT (ours) & 57.74 & 47.05 & 52.22 & 45.64 \\
\gct{} (ours) & 59.25 & 48.42 & 54.18 & 49.47 \\
\bottomrule
\end{tabular}
\end{table}

\paragraph{A second dataset at larger scale.} We repeat the from-scratch comparison on Tiny-ImageNet-200 with a ViT-S, images with $4\times$ the CIFAR pixel area, and $200$ classes (Table~\ref{tab:vittin}). Patch size 8 preserves the same $8\times8$ patch-token grid as the CIFAR model with patch size 4 instead of increasing the token count. \gct{} \vittinclaim{}. The contrast and input-scale results follow the pattern in Table~\ref{tab:vit}. The noise comparison differs. \gct{} scores $48.4\%$, compared with $48.2\%$ for nGPT and $46.2\%$ for QK-Norm. The noise ordering changes under the combined change in dataset, resolution, and backbone.

NormFormer reaches $43.2\%$ clean top-1 accuracy, about $12$ percentage points below most other baselines under this shared recipe. We report this result under the shared recipe, without architecture-specific retuning. The cause of the gap remains unresolved. \gct{} has the highest measured accuracy in each column of Table~\ref{tab:vittin}.

Across both vision datasets, \gct{} combines the highest measured clean accuracy with strong photometric and input-scale robustness. These observations complement the internal-rescaling tests in language modeling, while the dataset-dependent noise results distinguish scale robustness from robustness to arbitrary input perturbations.

\subsection{Multi-loss objectives: unit-system invariance without adaptivity}
\label{sec:exp-pinn}

\begin{figure}[t!]
\centering
\includegraphics[width=\textwidth]{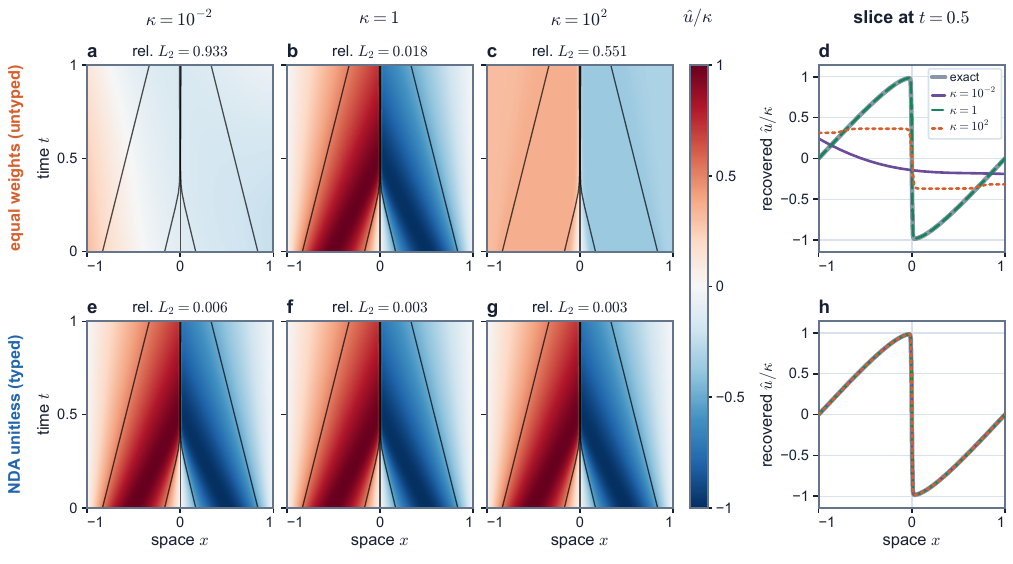}
\caption{Same Burgers flow, different numerical units: qualitative field recovery at seed 0 after 20,000 steps. Columns use $\kappa\in\{10^{-2},1,10^2\}$ and are converted back to the same physical field $\hat u/\kappa$. (a, b, c, e, f, g) Recovered field over space and time, with the recovered field's relative $L_2$ error above each panel, and black contours show the Cole--Hopf solution. (d, h) The same fields sliced at $t=0.5$ against the exact solution (grey). Equal weighting in the top row changes the recovered field off-reference, whereas the NDA-typed head and objective in the bottom row recover consistent fields. Table~\ref{tab:pinn} gives five-seed results.}
\label{fig:qual-pinn-fields}
\end{figure}

\begin{table}[t!]
\centering\small
\caption{PINN relative $L_2$ error of the recovered physical field, averaged over five seeds, when the data's unit system is rescaled by $\kappa$. The problem is physically identical across columns, and only the units change. The last column gives the worst case over $\kappa$.}
\label{tab:pinn}
\begin{tabular}{llcccc}
\toprule
problem & method & $\kappa=10^{-2}$ & $\kappa=1$ & $\kappa=10^{2}$ & worst\\
\midrule
\multirow{7}{*}{Burgers} & equal & 0.848 & 0.014 & 0.337 & 0.848 \\
 & manual $\lambda_{\mathrm{sup}}=100$ & 0.491 & 0.081 & 0.431 & 0.491 \\
 & uncertainty & 0.520 & 0.006 & 0.618 & 0.618 \\
 & LRA & 0.812 & 0.006 & 0.230 & 0.812 \\
 & NTK balance & 0.560 & 0.214 & 0.550 & 0.560 \\
\cmidrule(lr){2-6}
 & NDA-typed (ours) & 0.008 & 0.009 & 0.008 & 0.009 \\
 & NDA-typed $+$ LRA & 0.005 & 0.004 & 0.004 & 0.005 \\
\midrule
\multirow{7}{*}{Helmholtz} & equal & 0.111 & 0.110 & 2.973 & 2.973 \\
 & manual $\lambda_{\mathrm{sup}}=100$ & 0.035 & 0.005 & 2.002 & 2.002 \\
 & uncertainty & 0.054 & 0.002 & 2.175 & 2.175 \\
 & LRA & 0.048 & 0.006 & 1.620 & 1.620 \\
 & NTK balance & 0.036 & 0.390 & 0.971 & 0.971 \\
\cmidrule(lr){2-6}
 & NDA-typed (ours) & 0.010 & 0.015 & 0.014 & 0.015 \\
 & NDA-typed $+$ LRA & 0.004 & 0.003 & 0.004 & 0.004 \\
\bottomrule
\end{tabular}
\end{table}

Table~\ref{tab:pinn} reports recovered-field error as the data's unit system is rescaled by $\kappa\in\{10^{-2},1,10^{2}\}$. The physical problem is identical across columns. The NDA-typed objective and the NDA-typed $+$ LRA composition are the only evaluated methods that remain in the same low-error range across this $10^{4}$ unit sweep, whereas the other adaptive methods degrade for at least one rescaling.

Equal weighting is the unit-agnostic reference. The manual $\lambda_{\mathrm{sup}}=100$ rule is tuned at $\kappa=1$ and transferred. Uncertainty weighting, gradient-norm balancing, and NTK-trace balancing adapt from scale-sensitive training statistics. None of these baselines explicitly performs unit conversion, and each baseline degrades for at least one off-reference unit system. Equal weighting reaches a worst-case relative $L_2$ error of \pinnbaseworst{} across Burgers and Helmholtz because equal loss weights conflate unit conversion with task preference.

The NDA-typed objective types the output head and divides each term by an observable characteristic scale from App.~\ref{app:details}. The NDA-typed objective is invariant by construction and remains in the low-error range across $\kappa$, with worst case \pinnunitworst{}. The same objective also performs comparably at the reference scale without tuning, adaptivity, or access to the solution. This benchmark applies the loss-aggregation rule of \secref{sec:rules} end-to-end, typing both the output head and every loss term.

Practitioner guides for PINNs already recommend nondimensionalizing a problem before training \citep{wang2023expert}. The typed objective derives this nondimensionalization step directly from the loss-aggregation rule. Thm.~\ref{thm:lossscalar} shows that every invariant linear alternative whose weights depend only on these characteristic scales must use the same inverse-unit factors. Any remaining scale dependence is restricted to dimensionless groups. Cor.~\ref{cor:lossunique} shows that the fixed-$\alpha_i$ rule is unique in this class when the declared loss units vary independently.

Typing and adaptivity also compose. Applying LRA on top of the nondimensionalized terms, denoted NDA-typed $+$ LRA in Table~\ref{tab:pinn}, retains unit-system invariance with worst case \pinncomboworst{} across $\kappa$. This combination lets adaptation adjust dimensionless task preferences after unit conversion, demonstrating compatibility between typing and adaptive weighting.

\end{document}